%% file: core_arxiv.tex
\documentclass[twoside, 11pt]{article} 
\usepackage{amsmath,amssymb,natbib,graphicx,enumerate,color,caption,wrapfig,xspace,placeins,url, soul}
\usepackage{tikz,enumerate}
\usepackage[framemethod=TikZ]{mdframed}
\usepackage{pgf}
\usepackage{tikz}
\usetikzlibrary{arrows,automata}
\usepackage[latin1]{inputenc}
\usepackage{jmlr2e}
\usetikzlibrary{arrows,%
                shapes,positioning}

\usepackage[position=top]{subfig}
\usepackage{verbatim}
\usepackage{multirow}
\usepackage{diagbox}
\usepackage{rotating}
\usepackage[normalem]{ulem}

\usetikzlibrary{arrows}
\newdimen\arrowsize
\pgfarrowsdeclare{arcsq}{arcsq}
{
  \arrowsize=0.2pt
  \advance\arrowsize by .5\pgflinewidth
  \pgfarrowsleftextend{-4\arrowsize-.5\pgflinewidth}
  \pgfarrowsrightextend{.5\pgflinewidth}
}
{
  \arrowsize=1.5pt
  \advance\arrowsize by .5\pgflinewidth
  \pgfsetdash{}{0pt} 
  \pgfsetroundjoin   
  \pgfsetroundcap    
  \pgfpathmoveto{\pgfpoint{0\arrowsize}{0\arrowsize}}
  \pgfpatharc{-90}{-140}{4\arrowsize}
  \pgfusepathqstroke
  \pgfpathmoveto{\pgfpointorigin}
  \pgfpatharc{90}{140}{4\arrowsize}
  \pgfusepathqstroke
}

\def\abovestrut#1{\rule[0in]{0in}{#1}\ignorespaces}
\def\belowstrut#1{\rule[-#1]{0in}{#1}\ignorespaces}
\def\abovespace{\abovestrut{0.15in}}

\def\belowspace{\belowstrut{0.075in}}

\definecolor{pastelgray}{rgb}{0.81, 0.81, 0.77}

\input{definitions}

\title{Conditional Variance Penalties and Domain Shift Robustness}
\editor{}
\author{\name Christina Heinze-Deml \& Nicolai Meinshausen
 \\
\addr Seminar for Statistics\\
ETH Zurich\\
Zurich, Switzerland \\
 \email \texttt{\{heinzedeml,meinshausen\}@stat.math.ethz.ch}
}

\begin{document}

\maketitle
\begin{abstract}
When training a deep neural network for image classification,
one can broadly distinguish between two types of latent features of images
that will drive the classification. We can divide latent features
into  (i) `core' or  `conditionally invariant' features $\corefeat$ whose
distribution $\corefeat\vert Y$, conditional on the class $Y$, does not change substantially across
domains and  (ii) `style' features
$\style$ whose distribution $\style\vert Y$ can change
substantially across domains. Examples for style features include position, rotation, image quality or brightness but
also more complex ones like hair color, image quality or posture for images of
persons. Our goal is to minimize a loss that is robust under changes in the
distribution of these style features.
In contrast to previous work, we assume that
the domain itself is not observed and hence a latent variable.

 We do
assume that we can sometimes observe  a typically discrete identifier
or ``$\I$
variable''. In some applications we know, for example, that two images show the
same person, and  $\I$ then refers to the identity of the person.
The proposed method requires only a small fraction of images to have $\I$ information.
We group observations if they share the same class and identifier
$(Y,\I)=(y,\i)$ and penalize the conditional variance of the
prediction or the loss if we condition on $(Y,\I)$.
Using a causal framework, this conditional variance regularization (\core) is shown to
protect asymptotically against shifts in the distribution of the style variables. Empirically, we show that the \core penalty improves predictive accuracy substantially in settings where domain
changes occur in terms of image quality, brightness and color while we
also look at more complex changes such as changes in movement and
posture.

\end{abstract}

\begin{keywords}
 Domain shift; Dataset shift; Causal models; Distributional robustness; Anti-causal
 prediction; Image classification
\end{keywords}

\section{Introduction}\label{sec:intro}
Deep neural networks (DNNs) have achieved outstanding performance on
prediction tasks like visual object and speech recognition
\citep{Krizhevsky2012, Szegedy2015, He2015}. Issues can arise when the learned representations rely on dependencies
that vanish in test distributions (see for example \citet{Quionero-Candela2009, Torralba2011, Csurka2017} and references therein). Such domain shifts can
be caused by changing conditions such as color, background or location
changes. Predictive performance is then likely to degrade.
For example, consider the analysis presented in \citet{Kuehlkamp2017} which is concerned with the problem of predicting a person's gender based on images of their iris. The results indicate that this problem is more difficult than previous studies have suggested due to the remaining effect of cosmetics after segmenting the iris from the whole image.\footnote{Segmenting eyelashes from the iris is not entirely accurate which implies that the iris images can still contain parts of eyelashes, occluding the iris. As mascara causes the eyelashes to be thicker and darker, it is difficult to entirely remove the presence of cosmetics from the iris images.} Previous analyses obtained good predictive performance on certain datasets but when testing on a dataset only including images without cosmetics accuracy dropped. In other words, the high predictive performance previously reported relied to a significant extent on exploiting the confounding effect of mascara on the iris segmentation which is highly predictive for gender. Rather than the desired ability of discriminating based on the iris' texture the systems would mostly learn to detect the presence of cosmetics.

More generally, existing biases in
datasets used for training machine learning algorithms tend to be
replicated in the estimated models \citep{Bolukbasi2016}.
For an example involving Google's photo app, see
\citet{Crawford2016} and \citet{Emspak2016}. In \S\ref{sec:experiments} we show many examples where unwanted biases in the training data are
picked up by the trained model. As any bias in the training data is
in general used to discriminate between classes, these biases will
persist in future classifications, raising also
considerations of fairness and discrimination \citep{Barocas2016}.

Addressing the issues outlined above, we propose {\sc Co}nditional variance
{\sc Re}gularization (\core) to give differential weight to different latent
features.  Conceptually, we take a causal view of
the data generating process and categorize the latent data generating
factors into  `conditionally invariant' (\emph{core}) and
`orthogonal' (\emph{style}) features, as in \citet{Gong2016}. The core and style features are unobserved and can in general be highly nonlinear transformations of the observed input data.
It
is desirable that a classifier uses only the {core} features as they pertain to the
target of interest in a stable and coherent fashion. Basing a
prediction on the {core} features alone yields stable
predictive accuracy even if the {style} features are altered.
\core yields an estimator which is approximately invariant under
changes in the conditional distribution of the style features
(conditional on
the class labels) and it is
asymptotically robust with respect to domain shifts, arising
through interventions on the style features.  \core
relies on the fact that for certain datasets we can observe \emph{grouped
observations} in the sense that we observe the same object under
different conditions. Rather than pooling over all examples, \core
exploits knowledge about this grouping, i.e.,\ that a number of
instances relate to the same object. By penalizing between-object
variation of the prediction less than variation of the prediction for
the same object, we can steer the prediction to be based more on the
latent {core} features and less on the latent {style} features. While the proposed methodology can be motivated from the desire the achieve representational invariance with respect to the style features, the causal framework we use throughout this work allows to precisely formulate the distribution shifts we aim to protect against.

The remainder of this manuscript is structured as follows:
\S\ref{sec:motivating} starts with a few motivating examples, showing simple settings where the style features change in the test distribution such that standard empirical risk minimization approaches would fail.
In
\S\ref{sec:related_work} we review related work, introduce notation in \S\ref{sec:setting}
and in \S\ref{sec:counter_reg} we formally introduce conditional
variance regularization \core. In \S\ref{sec:theory},  \core is shown
to be asymptotically equivalent
to minimizing the risk under a suitable class of strong interventions in
a partially linear classification setting, provided one chooses
sufficiently strong \core penalties. We also show that the population \core
penalty induces domain shift robustness for general loss
 functions to first order in the intervention strength. The
size of the conditional variance penalty can be shown to determine the size of the distribution class over
which we can expect distributional robustness.
In \S\ref{sec:experiments} we evaluate the performance of \core in a variety of experiments.

\begin{figure}
\begin{center}
\begin{minipage}[t]{1\hsize}
\centering
\subfloat[Example 1, training set.]{
     \includegraphics[width=.48\textwidth, keepaspectratio=true, trim={4mm 5mm 0 0}, clip]{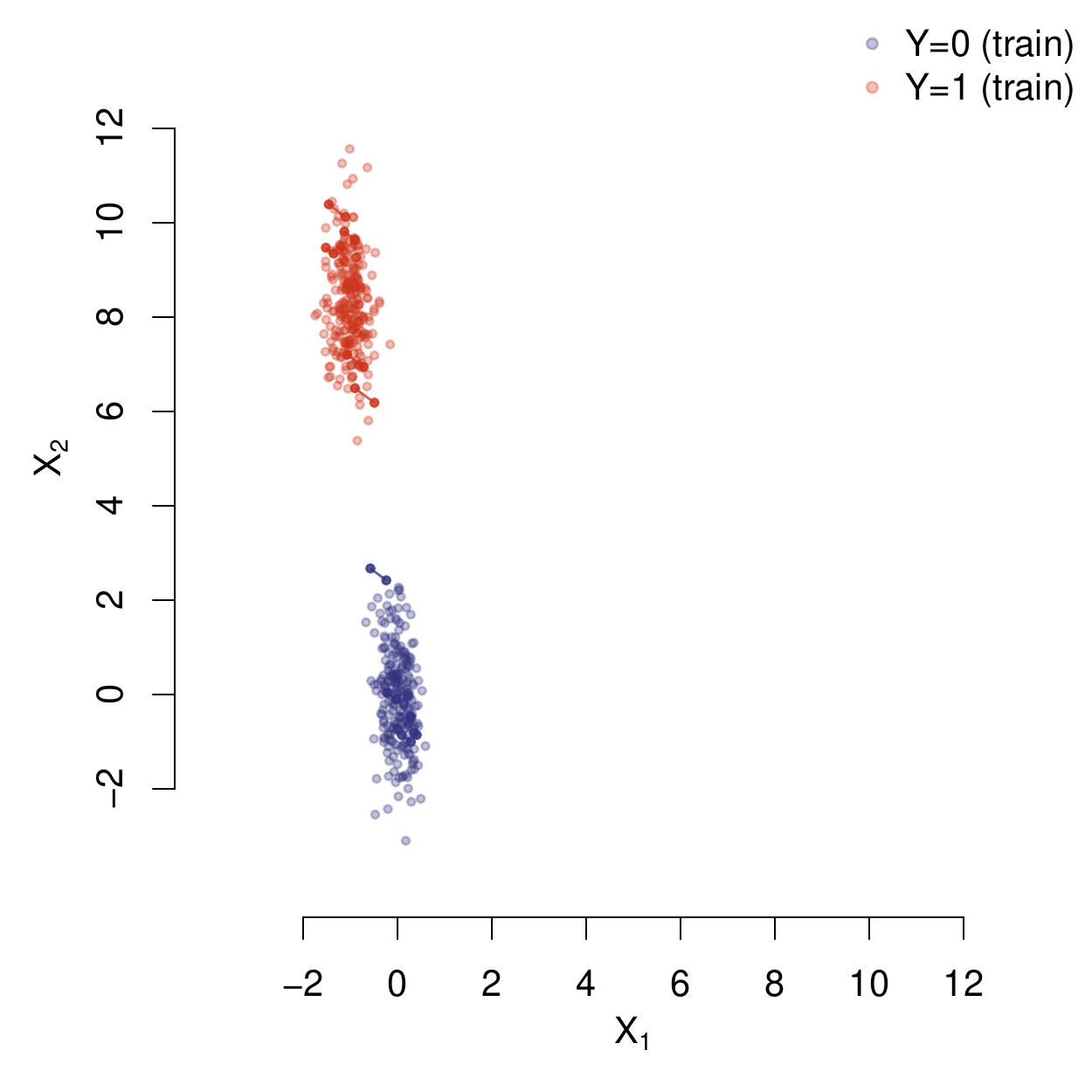}\label{ex1_train}
}
\subfloat[Example 1, test set.]{
     \includegraphics[width=.48\textwidth, keepaspectratio=true, trim={4mm 5mm 0 0}, clip]{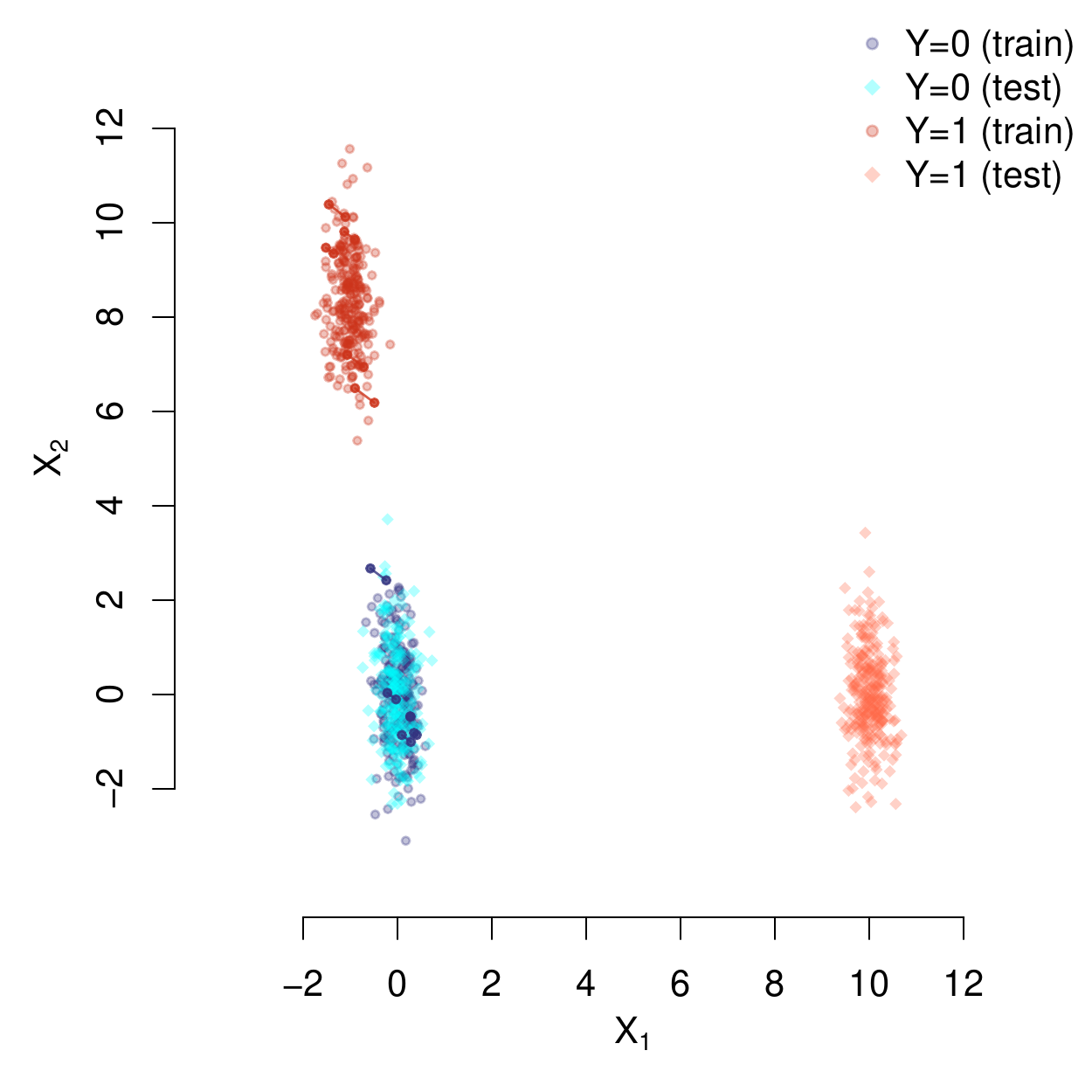}\label{ex1_test}
}

\subfloat[Example 2, training set.]{
     \includegraphics[width=.48\textwidth, keepaspectratio=true, trim={4mm 5mm 0 0}, clip]{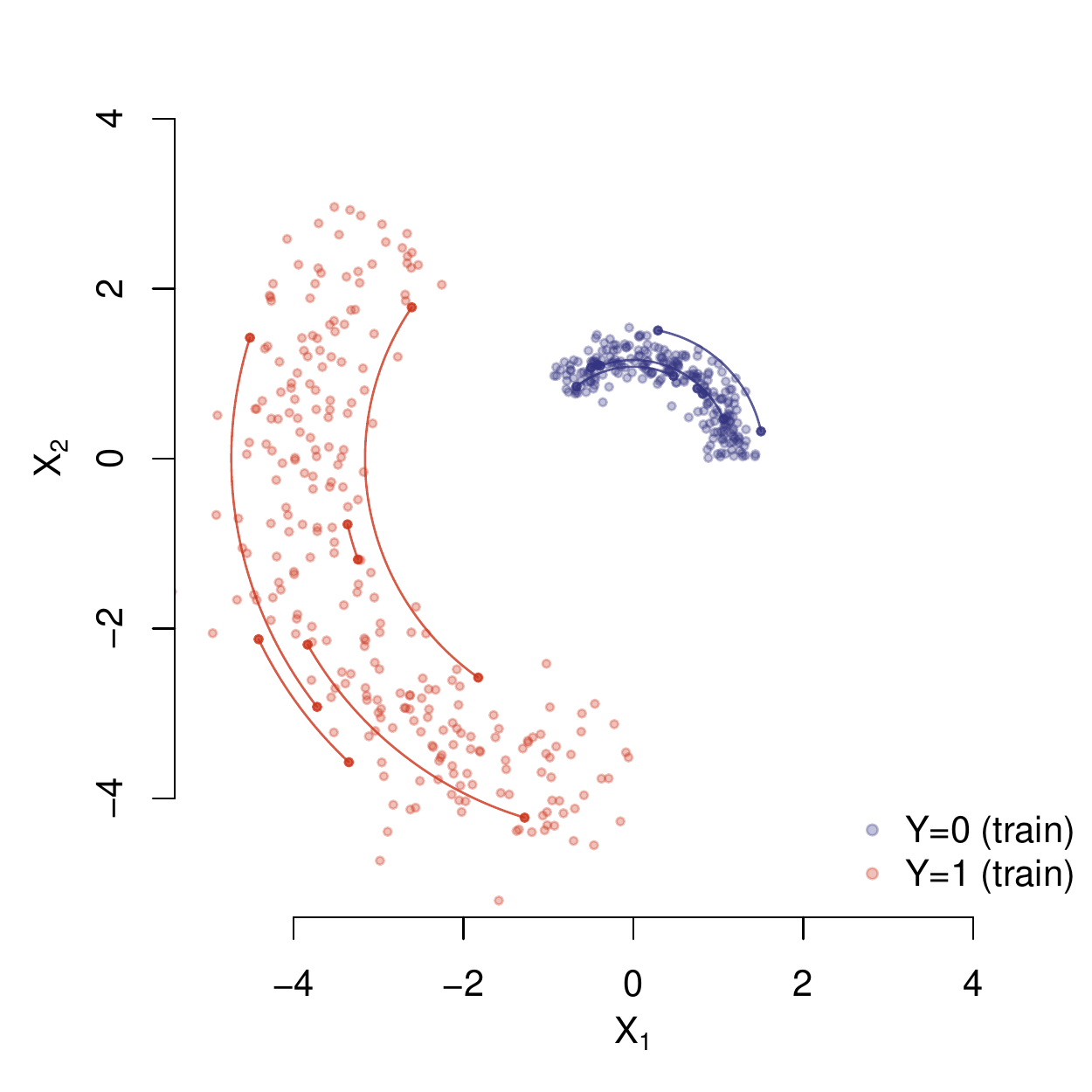}\label{ex2_train}
}
\subfloat[Example 2, test set.]{
     \includegraphics[width=.48\textwidth, keepaspectratio=true, trim={4mm 5mm 0 0}, clip]{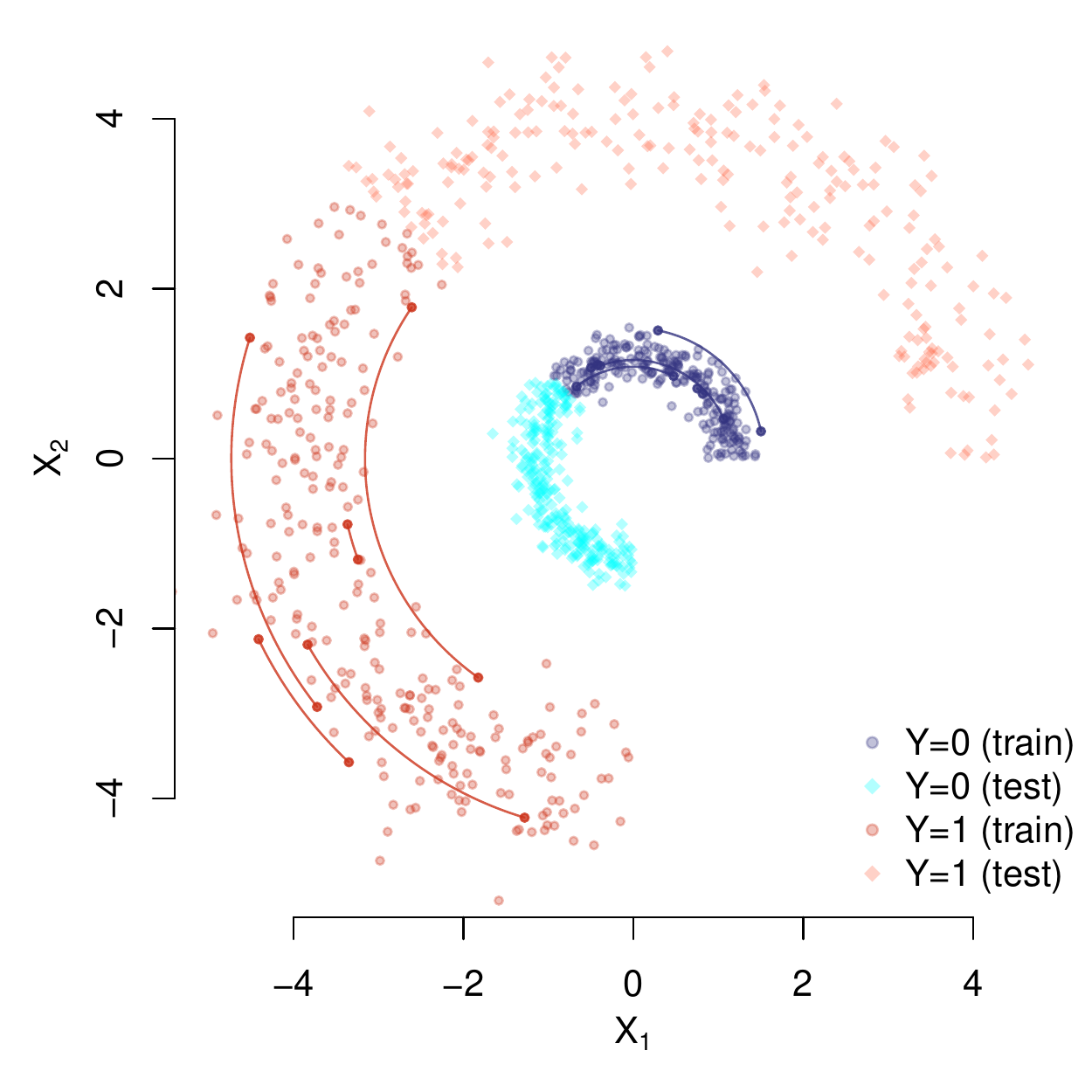}\label{ex2_test}
}

\end{minipage}

\captionof{figure}{{\small Motivating examples~1 and~2: a linear
    example in (a) and (b) and a nonlinear example in (c) and (d). The distributions are shifted in
    test data by style interventions where style in example (a/b) is the
  linear direction $(1,-0.75)$ and the polar angle in example (c/d). Standard estimators achieve error rates of $0\%$ on the training data and test data drawn from the same distribution as the training data (panels (a) and (c), respectively). On the shown test set where the distribution of the style conditional on $Y$ has changed the error rates are $> 50\%$ (panels (b) and (d), respectively).
  }}\label{fig:simple}
\end{center}
\end{figure}

\begin{figure}
\begin{center}
\begin{minipage}[t]{1\hsize}
\subfloat[Example 3, training set.]{\label{ex3_train}
     \includegraphics[width=.07\textwidth, keepaspectratio=true, trim={0mm 0mm 0 0}, clip]{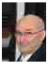}
          \includegraphics[width=.07\textwidth, keepaspectratio=true, trim={0mm 0mm 0 0}, clip]{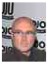}
          \includegraphics[width=.07\textwidth, keepaspectratio=true, trim={0mm 0mm 0 0}, clip]{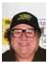}
          \includegraphics[width=.07\textwidth, keepaspectratio=true, trim={0mm 0mm 0 0}, clip]{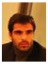}
          \includegraphics[width=.07\textwidth, keepaspectratio=true, trim={0mm 0mm 0 0}, clip]{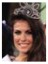}
          \includegraphics[width=.07\textwidth, keepaspectratio=true, trim={0mm 0mm 0 0}, clip]{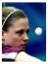}

}
\hspace{1cm}
\subfloat[Example 3, test set.]{\label{ex3_test}
     \includegraphics[width=.07\textwidth, keepaspectratio=true, trim={0mm 0mm 0 0}, clip]{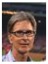}
          \includegraphics[width=.07\textwidth, keepaspectratio=true, trim={0mm 0mm 0 0}, clip]{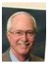}
          \includegraphics[width=.07\textwidth, keepaspectratio=true, trim={0mm 0mm 0 0}, clip]{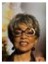}
          \includegraphics[width=.07\textwidth, keepaspectratio=true, trim={0mm 0mm 0 0}, clip]{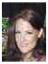}
          \includegraphics[width=.07\textwidth, keepaspectratio=true, trim={0mm 0mm 0 0}, clip]{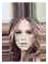}
          \includegraphics[width=.07\textwidth, keepaspectratio=true, trim={0mm 0mm 0 0}, clip]{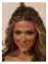}

}
\end{minipage}

\captionof{figure}{{\small Motivating example 3: The goal is to predict whether a person
    is wearing glasses. The distributions are shifted in
    test data by style interventions where style is
  the image quality.
 A   5-layer CNN achieves 0\% training error and 2\% test
  error for images that are sampled from the same distribution
  as the training images (a), but a 65\% error rate
  on images where the confounding between image quality and glasses is
  changed (b). See \S\ref{subsec:celeba_conf}  for more details.
}}\label{fig:simple_img_qual}
\end{center}
\end{figure}

To summarize, our contributions are the following:
\begin{enumerate}[(i)]
\item {\bf Causal framework and distributional robustness.}
We provide a causal framework to define distributional shifts for
style variables. Our framework allows that the domain variable itself is latent.

\item {\bf Conditional variance penalties.}
We introduce conditional variance penalties and show two robustness properties in Theorems~\ref{thm:logistic} and~\ref{th:2}. 

\item {\bf Software.}
We illustrate our ideas using synthetic and real-data experiments.
A TensorFlow implementation of \core as well as code to reproduce some of the experimental results are
available at \url{https://github.com/christinaheinze/core}.
\end{enumerate}

\subsection{Motivating examples}\label{sec:motivating}
To motivate the methodology we propose, consider the examples shown in
Figures~\ref{fig:simple} and~\ref{fig:simple_img_qual}. Example 1 shows a setting where a linear
decision boundary is suitable. Panel~\subref{ex1_train} in Figure~\ref{fig:simple} shows a subsample of the
training data where class 1 is associated with red points, dark blue
points correspond to class 0. If we were asked to draw a decision
boundary based on the training data, we would probably choose one that
is approximately horizontal. The style feature here corresponds to a linear
direction $(1, -0.75)^t$.
Panel~\subref{ex1_test} shows a subsample of the test set where the style feature is intervened upon for class~1 observations: class 1 is associated with orange squares, cyan squares correspond to class 0. Clearly, a horizontal decision boundary would have misclassified all test points of class~1.

Example 2 shows a setting where a nonlinear decision boundary is
required. Here, the {core} feature corresponds to the distance
from the origin while the style feature corresponds to the angle
between the $x_1$-axis and the vector from the origin to $(x_1,
x_2)$. Panel~\subref{ex2_train} shows a subsample of the training data and
panel~\subref{ex2_test} additionally shows a subsample of the test data where the
style---i.e. the distribution of the angle---is intervened upon. Clearly, a circular decision boundary yields optimal performance on both training and test set but is unlikely to be found by a standard classification algorithm when only using the training set for the estimation. We will return to these examples in \S\ref{sec:ex_class}.

Lastly, we introduce a strong dependence between the class label and the style feature ``image quality'' in the third example by
manipulating the face images from the CelebA dataset
\citep{Liu2015}: in the training set images of class
``wearing glasses'' are associated with a lower image quality than
images of class ``not wearing glasses''. Examples are shown in Figure~\ref{fig:simple_img_qual}\subref{ex3_train}. In the test set, this
relation is reversed, i.e.\ images showing persons wearing glasses are
of higher quality than images of persons without glasses, with examples
in Figure~\ref{fig:simple_img_qual}\subref{ex3_test}.
 We will
return to this example in \S\ref{subsec:celeba_conf} and show that
training a convolutional neural network to distinguish between people
wearing glasses or not works well on test data that are drawn from the
same distribution (with error rates below~2\%) but fails entirely on
the shown test data, with error rates worse than~65\%.

\subsection{Related work}\label{sec:related_work}
For general distributional robustness, the aim is to learn
\begin{equation}\label{eq:dr} \text{argmin}_\theta\; \sup_{F\in \mathcal{F}} E_F(\ell(Y,f_\theta(X)))\end{equation}
for a given set $\mathcal{F}$ of distributions, twice differentiable
and convex loss $\ell$, and
prediction $f_\theta(x)$.  The set $\mathcal{F}$ is the set of
distributions on which  one would like the estimator to achieve a
guaranteed performance bound.

Causal inference can be seen to be a specific instance of distributional robustness, where we take
$\mathcal{F}$ to be the class of all distributions generated under do-interventions on $X$ \citep{Meinshausen18distrib, Rothenhaeusler2018}.
Causal models thus have the defining advantage that the
predictions will be valid even under arbitrarily large interventions
on all predictor variables \citep{Haavelmo1944,Aldrich1989, Pearl2009,
  Schoelkopf2012, Peters2016, Zhang2013b, Zhang2015, Yu2017,
  Rojas-Carulla2017, Magliacane2017}. There are two difficulties in
transferring these results to the setting of  domain
shifts in image classification. The first  hurdle is that the
classification task is typically anti-causal since the image we use as
a predictor is a descendant of the true class of the object we are
interested in rather than the other way around. The second challenge
is that we do not want (or could) guard against arbitrary interventions on any
or all variables but only would like to guard against a shift of the
style features. It is hence not immediately obvious how standard
causal  inference can be used to guard against large domain shifts.

Another line of work uses a class of distributions
 of the form $\mathcal{F}= \mathcal{F}_\epsilon(F_0)$ with
\begin{equation}\label{eq:ball}  \mathcal{F}_\epsilon(F_0):= \{ \text{distributions } F \text{ such that }
D(F,F_0) \le \epsilon \},\end{equation}
with $\epsilon>0$  a small constant and $D(F,F_0)$ being, for example,  a
$\phi$-divergence \citep{Namkoong2017, Ben-Tal2013, Bagnell2005, volpi2018generalizing} or a
Wasserstein distance \citep{Shafieezadeh-Abadeh2017, Sinha2017, Gao2017}.
The distribution $F_0$ can be the true (but generally unknown)
 population distribution $P$ from which the data were drawn or its empirical counterpart $P_n$.
The distributionally robust targets in Eq.~\eqref{eq:ball} can often be expressed in
penalized form
\citep{Gao2017,Sinha2017, Xu2009}. A Wasserstein-ball is a suitable
class of distributions for example in the context of adversarial examples \citep{Sinha2017, Szegedy2014,Goodfellow2015}.

In this work, we do not try to achieve robustness with
respect to a set of distributions that are  pre-defined by a
Kullback-Leibler divergence or a Wasserstein metric as
in Eq.~\eqref{eq:ball}.
We try to achieve
robustness against a set of distributions that are generated
by interventions on latent style variables. We will formulate the class of distributions over
which we try to achieve robustness as in Eq.~\eqref{eq:dr} but with the
class of distributions in Eq.~\eqref{eq:ball} now replaced with
\begin{equation}\label{eq:classFintro} \mathcal{F}_\xi =\{F: D_{\styleo} (F,F_0) \le
  \xi \}, \end{equation}
where $F_0$ is again the distribution the training data are drawn from.
The difference to standard distributional robustness approaches listed
below Eq.~\eqref{eq:ball}  is now that the metric $D_{\styleo}$
measures the shift of the orthogonal style features. We do not know a priori which features are prone to
distributional shifts and which features have a stable (conditional)
distribution. The metric is hence not
known a priori and needs to be inferred in a suitable sense
from the data.

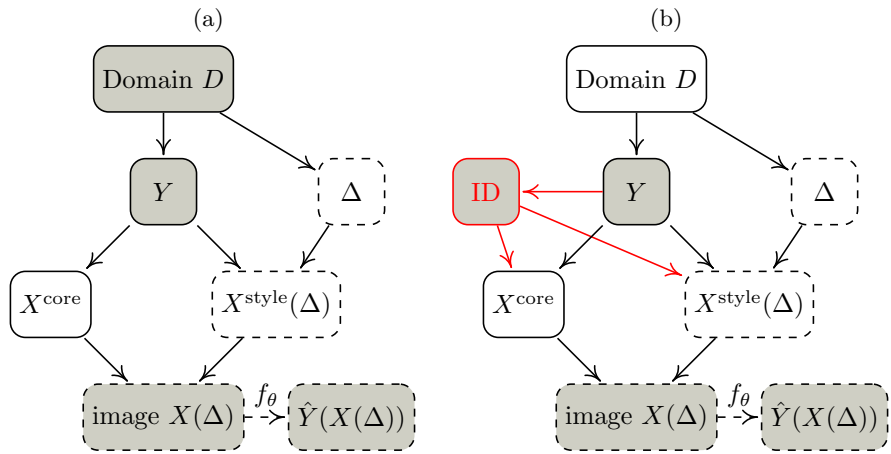
\begin{figure}
\begin{center}
\subfloat[\xspace]{
\begin{small}
\begin{tikzpicture}[>=stealth',shorten >=1pt,auto,node distance=3cm,
                    semithick]
  \tikzstyle{every state}=[fill=none,shape=rectangle,rounded corners=2mm,text=black]

  \draw (-1.5,4.5) node[state, fill=pastelgray] (H) {Domain $\domain$};
  \draw (-1.5,3) node[state, fill=pastelgray] (Y) {$Y$};
  \draw (0,1.5) node[state, dashed] (S) {$\orth(\Delta)$ };
  \draw (1,3) node[state, dashed] (D) {$\Delta$};
\draw (-3,1.5) node[state] (W) {$\corefeat$};
  \draw (-1.5,0) node[state,fill=pastelgray, dashed] (X) {image $X(\Delta)$};
  \draw (1,0) node[state,fill=pastelgray, dashed] (Yhat) {$\hat{Y}(X(\Delta))$};

 \draw [-arcsq] (H) -- (Y);
 \draw [-arcsq] (H) -- (D);
 \draw [-arcsq] (Y) -- (W);
 \draw [-arcsq] (Y) -- (S);
 \draw [-arcsq] (D) -- (S);
 \draw [-arcsq] (W) -- (X);
 \draw [-arcsq] (S) -- (X);
 \draw [-arcsq, dashed] (X) -- node {$f_\theta$} (Yhat);
\end{tikzpicture}
\end{small}
}
\subfloat[\xspace]{
\begin{small}
\begin{tikzpicture}[>=stealth',shorten >=1pt,auto,node distance=3cm,
                    semithick]
  \tikzstyle{every state}=[fill=none,shape=rectangle,rounded corners=2mm,text=black]

  \draw (-1.5,4.5) node[state] (H) {Domain $\domain$};
  \draw (-1.5,3) node[state, fill=pastelgray] (Y) {$Y$};
  \draw (0,1.5) node[state, dashed] (S) {$\orth(\Delta)$ };
  \draw (1,3) node[state, dashed] (D) {$\Delta$};
  \draw (-3.5,3) node[red, state, fill=pastelgray] (I) {\color{red}  $\I$};
  \draw (-3,1.5) node[state] (W) {$\corefeat$};
  \draw (-1.5,0) node[state, fill=pastelgray, dashed] (X) {image $X(\Delta)$};
  \draw (1,0) node[state, fill=pastelgray, dashed] (Yhat) {$\hat{Y}(X(\Delta))$};
 \draw [-arcsq] (H) -- (Y);
 \draw [-arcsq, red]  (Y) -- (I);
 \draw [-arcsq, red]  (I) -- (S);
 \draw [-arcsq] (H) -- (D);
 \draw [-arcsq] (Y) -- (W);
 \draw [-arcsq,red ] (I) -- (W);
 \draw [-arcsq] (Y) -- (S);
 \draw [-arcsq] (D) -- (S);
 \draw [-arcsq] (W) -- (X);
 \draw [-arcsq] (S) -- (X);
 \draw [-arcsq, dashed] (X) -- node {$f_\theta$} (Yhat);
\end{tikzpicture}
\end{small}
}



\captionof{figure}{ \small Observed quantities are shown as shaded nodes; nodes of latent quantities are transparent. Left: data generating process for the considered
  model as in \citet{Gong2016}, where the effect of the domain on the
  orthogonal features $\orth$ is  mediated via unobserved noise
  $\Delta$.  The style interventions and all its
  descendants are shown as nodes with dashed borders to highlight variables that are affected by
  style interventions. Right: our setting. The domain itself is unobserved but we
  can now observe the (typically discrete) $\I$ variable  we use for
  grouping. The arrow between $\I$ and $Y$ can be reversed, depending
  on the sampling scheme.}
\label{fig:DAG}
\end{center}
\end{figure}

Similar to this work in terms of their goals are  the work of \cite{Gong2016} and Domain-Adversarial Neural Networks (DANN) proposed in \citet{Ganin2016}, an approach motivated by the work of \cite{Ben-David2007}.
The main idea of~\citet{Ganin2016} is to learn a representation that contains no discriminative information about the origin of the input (source or target domain). This is achieved by an adversarial training procedure: the loss on domain classification is maximized while the loss of the target prediction task is minimized simultaneously.
The data generating process assumed in \cite{Gong2016} is similar to our model, introduced in \S\ref{subsec:non_anc_interv}, where we detail the similarities and differences between the models (cf.\ Figure~\ref{fig:DAG}). \citet{Gong2016} identify the conditionally independent features by adjusting a  transformation of the variables to  minimize the squared MMD distance between distributions in different domains\footnote{The distinction between `conditionally independent' features  and `conditionally transferable' (which is the former modulo location and scale transformations) is for our purposes not relevant as we do not make a linearity assumption in general.}.
The  fundamental difference between these very promising methods and
our approach is that we use a different data basis. The domain
identifier is explicitly observable in \cite{Gong2016} and
\citet{Ganin2016},  while it is latent in our approach. In contrast,
we exploit the presence of an identifier variable~$\I$ that relates to
the identity of an object (for example identifying a person). In other
words, we do {\it not} assume that we have data from different domains but
just different realizations of the same object under different
interventions.
This also differentiates this work from \emph{latent} domain adaptation papers from the computer vision literature \citep{Hoffman2012,Gong2013}.
Further related work is discussed in \S\ref{sec:furtherwork}.

\section{Setting}\label{sec:setting}
We introduce the assumed underlying causal graph and some
notation before discussing notions of domain shift robustness.

\subsection{Causal graph}\label{subsec:non_anc_interv}
Let $Y\in \mathcal{Y}$ be a target of interest.  Typically
$\mathcal{Y}=\mathbb{R}$ for regression or
$\mathcal{Y}=\{1,\ldots,K\}$ in classification with $K$ classes. Let
$X \in \mathbb{R}^p $ be predictor variables, for example the $p$ pixels of an
image.
The  causal structural model
for  all variables  is shown in the panel~(b) of
Figure~\ref{fig:DAG}. The domain
variable $D$ is latent, in contrast to \citet{Gong2016} whose model is shown in panel~(a) of
Figure~\ref{fig:DAG}.  We add the
$\I$ variable whose
distribution can change conditional on $Y$. In Figure~\ref{fig:DAG},  $Y \rightarrow \I$ but in some settings it might be more plausible to consider $\I \rightarrow Y$. For the proposed method both options are possible. Together with $Y$, the $\I$
variable is used to group observations. It is typically
discrete and relates to the identity of the underlying object (identity of a person, for example). The variable can be assumed to be latent in the setting of \citet{Gong2016}.

The rest of the graph is in analogy to \citet{Gong2016}.
The prediction is anti-causal, that is
the predictor variables $X$  that we  use for $\hat{Y}$  are
non-ancestral to $Y$. In other words, the class label is here seen to
be causal for the image and not the other way around\footnote{
If an existing image is classified by a human, then the image
is certainly ancestral for the attached label. If the label $Y$
refers, however, to the underlying true object (say if you generate
images by asking people to take pictures of objects), then the more
fitting model is the one where $Y$ is ancestral for
$X$.}.
The causal effect from the class label $Y$ on the image $X$ is
mediated  via two types of latent variables: the so-called \emph{core} or `conditionally
 invariant' features
$\corefeat$  and the orthogonal or \emph{style} features $\style$. The distinguishing
factor between the two is that external interventions $\Delta$ are possible on the {style}
features but not on the {core} features. If the interventions $\Delta$
have different distributions in different domains, then the
conditional
distributions $\corefeat|Y=y,\I=\i$ are invariant for all $(y,\i)$ while
$\orth|Y=y,\I=\i$ can change.
The {style} variable can include point of view,
 image quality, resolution, rotations, color changes, body posture,
 movement etc.\ and will in general be context-dependent\footnote{The type
of features we regard as style and which ones we regard as core
features can conceivably change depending on the circumstances---for instance, is the
color ``gray'' an integral part of the object ``elephant'' or can it be
changed so that a colored elephant is still considered to be an elephant?}.
The style intervention variable $\Delta$ influences  both the latent
style $\style$,
and hence also the image $X$. In potential outcome notation, we  let  $\style(\Delta=\delta)$ be the
style under intervention
$\Delta=\delta$ and $X(Y,\I,\Delta=\delta)$ the image for class $Y$,
identity $\I$ and style
intervention $\Delta$. The latter is sometimes abbreviated as
$X(\Delta=\delta)$ for notational simplicity. Finally,  $f_\theta(X(\Delta=\delta))$ is  the
prediction under the style  intervention $\Delta=\delta$. For a formal justification of
using a causal graph and potential outcome notation simultaneously see
\citet{Richardson2013}.

To be specific, if not mentioned otherwise we will assume a causal graph as follows. For
 independent $\varepsilon_Y, \varepsilon_\I,\varepsilon_{\styleo}$ in
$\mathbb{R},\mathbb{R},\mathbb{R}^q$ respectively with
positive density on their support and continuously differentiable
functions $k_y, k_\i$, and $k_{\styleo},k_{\corefeato},k_x$,
\begin{align}
 &Y \leftarrow k_y(\domain,\varepsilon_Y) \nonumber \\
\text{identifier } & \I \leftarrow k_\i(Y,\varepsilon_\I) \nonumber \\
\text{core or conditionally invariant features }& \corefeat  \leftarrow k_{\corefeato}(Y,\I)  \nonumber \\
\text{style or orthogonal features }& \style  \leftarrow k_{\styleo}(Y,\I,\varepsilon_{\styleo}) + \Delta \nonumber \\
\text{image }& X  \leftarrow  k_x(\corefeat, \style) . \label{eq:SEM}
\end{align}
Hence, the core features are assumed to be a deterministic function of $Y$ and $\I$.
The prediction $\hat{y}$ for $y$, given
$X=x$, is of the form $f_\theta(x)$ for a suitable function $f_\theta$ with
parameters $\theta\in \mathbb{R}^d$, where the parameters $\theta$
correspond to the weights in a DNN, for example.

\subsection{Data}
We assume we have $\ntot$ data points $(x_i,y_i,\i_i)$ for
$i=1,\ldots,\ntot$, where the observations $\i_i$ with $i=1,\ldots,\ntot$ of variable
$\I$ can also contain unobserved values. Let  $\nid \le \ntot$ be the number
of unique realizations of $(Y,\I)$ and let $S_1,\ldots,S_\nid$ be a
partition of $\{1,\ldots,\ntot\}$ such that, for each $j\in
\{1,\ldots,\nid\}$, the realizations   $(y_i,\i_i)$ are identical\footnote{Observations where the $\I$ variable is unobserved
  are not grouped, that is each such observation is counted as a
  unique observation of $(Y,\I)$.} for
all $i\in S_j$. While our prime application is classification,
regression settings with continuous $Y$ can be approximated in this
framework by slicing  the range of the response variable into distinct bins in
analogy to the approach in sliced inverse regression \citep{li1991sliced}.
   The cardinality of  $S_j$ is denoted by
$\ntot_j:=|S_j| \ge 1$.  Then $\ntot = \sum_i \ntot_i$ is again  the total number of samples and $\ncf = \ntot-\nid $ 
 is the total number of grouped observations.
Typically $\ntot_i=1$ for most samples and  occasionally
$\ntot_i \ge 2$ but one can also envisage scenarios with  larger groups
of the same identifier $(y,\i)$.

\subsection{Domain shift robustness}\label{subsec:counterfactual_reg}

In this section, we clarify against which classes of distributions we hope to achieve robustness.
Let $\ell$ be a suitable loss that maps $y$ and
$\hat{y}=f_\theta(x)$ to $\mathbb{R}^+$.
The  risk  under distribution $F$ and parameter $\theta$ is given by
\[   E_F\Big[  \ell( Y, f_\theta(X))  \Big] .\]
Let $F_0$ be the joint distribution of $(\I,Y,\style)$ in the
 training distribution.
A new domain and explicit interventions on the style features can now
shift the distribution of $(\I,Y,\tstyle) $ to $F$.
We can measure the distance between distributions $F_0$ and $F$ in
different ways. Below we will define the distance considered in this work and denote it by
$ D_{\styleo}(F,F_0)$. Once defined, we get a class of distributions
\begin{equation}\label{eq:classF} \mathcal{F}_\xi =\{F: D_{\styleo}(F_0,F) \le
  \xi \} \end{equation} and the goal will be to optimize a worst-case loss over this distribution
class in the sense of Eq.~\eqref{eq:dr}, where larger values of $\xi$ afford  protection against larger
distributional changes. The relevant loss for
distribution class $\mathcal{F}_\xi$ is then
\begin{equation}\label{eq:adv_loss_finite}
	L_\xi(\theta) =  \sup_{F\in \mathcal{F}_\xi} E_{F}\Big[
\ell\big(Y, f_\theta\big(X\big)\big)\Big] .
\end{equation}
In the limit of arbitrarily strong interventions on the
  style features $\style$, the loss  is given by
\begin{equation}\label{eq:adv_loss}
	L_\infty(\theta) =  \lim_{\xi\to\infty } \sup_{F\in \mathcal{F}_\xi} E_{F}\Big[
\ell\big(Y, f_\theta\big(X\big)\big)\Big] .
\end{equation}
Minimizing the loss $L_\infty(\theta)$ with respect to
$\theta$ guarantees an accuracy in prediction which will work well
across arbitrarily large shifts in the conditional distribution of the
style features.

A natural choice to define $D_\styleo$ is to use a Wasserstein-type distance
\citep[see e.g.][]{villani2003topics}.
We will first define a distance $D_{y,\i}$   for the
conditional distributions
\[ \style |Y=y,\I=\i\quad \text{   and    } \quad \tstyle |
Y=y,\I=\i,\]
and then set $D(F_0,F) = E(D_{Y,\I})$, where the expectation is with
respect to random $\I$ and labels $Y$.
The distance $D_{y,\i}$
between the two conditional distributions of $\style$ will be
defined as a  Wasserstein $W_2^2(F_0,F)$-distance for a suitable cost function
$c(x,\tilde{x})$. Specifically, let $\Pi_{y,\i}$
be the couplings between the conditional distributions of $\style$ and
$\tstyle$, meaning measures supported on $\mathbb{R}^q \times
\mathbb{R}^q$ such that the marginal distribution over the first $q$
components is equal to the distribution of $\style$ and the marginal
distribution over the remaining $q$ components equal to the
distribution of $\tstyle$. Then the distance between the conditional
distributions is defined as
\[ D_{y,\i} = \min_{M\in \Pi_{y,\i}} \; E\big[ c(x,\tilde{x})\big],\]
where $c :\mathbb{R}^q\times \mathbb{R}^q \mapsto \mathbb{R}^+$ is a
nonnegative, lower semi-continuous cost function.
 Here, we focus on a Mahalanobis distance as cost
\[ c^2(x,\tilde{x} ) = (x-\tilde{x})^t  \Sigma_{y,\i}^{-1}
(x-\tilde{x}) .\]
The cost of a shift is hence measured against the  variability under the distribution $F_0$,
$\Sigma_{y,\i} =\text{Cov}(\style|Y,\I)$\footnote{
As an example, if the change in distribution for $\style$ is caused by
random shift-interventions $\Delta$, then
$ \tstyle  \leftarrow \style +\Delta$,
and the distance $D_{\styleo}$ induced in the distributions is
\[ D_{\styleo} (F_0,F) \le E\big[  E( \Delta^t \Sigma^{-1}_{y,\i} \Delta
|Y=y,\I=\i)  \big],\]
ensuring that the strength of the shifts is measured against the
natural variability $\Sigma_{y,\i}$ of the style features.}.

\section{Conditional variance regularization}\label{sec:counter_reg}

\subsection{Pooled estimator}
Let $(x_i,y_i)$ for $i=1,\ldots,\ntot$ be the observations that
constitute the training data and $\hat{y_i}=f_\theta(x_i)$ the prediction
for $y_i$.
The standard approach is to simply pool over all available
observations, ignoring any grouping information that might be
available.
 The pooled
estimator thus treats all examples identically by summing over the
empirical loss as
\begin{equation}\label{eq:pool} \hat{\theta}^{pool} = \mbox{ argmin}_\theta\;
   \hat{E}\Big[\ell(Y, f_\theta(X))\Big] + \gamma\cdot \text{pen}(\theta),\end{equation}
where the first part is simply the empirical loss over the  training
data,
\[ \hat{E}\Big[\ell(Y, f_\theta(X))\Big]  = \frac{1}{\ntot} \sum_{i=1}^\ntot   \ell\big( y_i ,
f_\theta(x_{i})\big) .\]
In the second part, $\text{pen}(\theta)$ is a complexity penalty, for example a
squared $\ell_2$-norm of the weights $\theta$ in a convolutional neural
network as a ridge penalty. All examples that compare to the pooled
estimator will include a ridge penalty as default.

\subsection{\core estimator}\label{sec:core}
The \core estimator is defined in Lagrangian form for penalty
$\lambda\ge 0$ as
\begin{equation}\label{eq:hat}  \hat{\theta}^{core}(\lambda)
  \;=\;\text{argmin}_\theta\;   \hat{E}\Big[\ell(Y, f_\theta(X))\Big]
  \;    + \lambda\cdot \hat{C}_{\theta}.\end{equation}
The penalty $\hat{C}_{\theta} $ is a conditional variance penalty of the
form
\begin{align}
\text{conditional-variance-of-prediction:}\qquad
  \hat{C}_{f,\nu,\theta}&:=\hat{E}\big[ \widehat{\text{Var}}(
                  f_\theta(X)|Y,\I)^\nu\big] \label{eq:condvar} \\
\text{conditional-variance-of-loss:}\qquad \hat{C}_{\ell.\nu,\theta}&:=\hat{E}\big[ \widehat{\text{Var}}( \ell(Y, f_\theta(X))|Y,\I) ^\nu\big] \label{eq:condvarloss},
\end{align}
where typically $\nu\in\{1/2,1\}$. For $\nu=1/2$, we also refer to the respective penalties as ``conditional-standard-deviation'' penalties. In the equivalent constrained form, the estimator can be viewed as an instance of a restricted maximum likelihood estimator \citep{harville1974bayesian,verbeke2009linear}.
In practice in the context of classification and DNNs, we apply the penalty~\eqref{eq:condvar} to the predicted logits.
The conditional-variance-of-loss penalty~\eqref{eq:condvarloss} takes a similar form to
\citet{Namkoong2017}. The crucial difference of our approach to \citet{Namkoong2017} is that we penalize with the expected
\emph{conditional} variance or standard deviation. The fact that we take a conditional variance is here
important as we try to achieve distributional robustness with respect
to interventions on the style variables. Conditioning on $\I$
allows to guard specifically against these interventions. An unconditional variance
penalty, in contrast, can achieve robustness against a pre-defined
class of distributions such as a ball of distributions defined in a Kullback-Leibler or Wasserstein
metric.
The population \core estimator
 is defined as in Eq.~\eqref{eq:hat} where empirical estimates are replaced by their
 respective population quantities.

Before showing numerical examples, we discuss the estimation of the
expected conditional variance in \S\ref{subsec:convar} and
return to the simple
examples of \S\ref{sec:motivating} in
\S\ref{sec:ex_class}. Domain shift robustness
in a classification setting
for a partially linear version of the structural equation model~\eqref{eq:SEM} is shown in
\S\ref{sec:log_reg}. Furthermore, we discuss the
population limit of $\hat{\theta}^{core}(\lambda)$
 in \S\ref{sec:first_order}, where we show that the regularization
parameter  $\lambda\ge 0$ is  proportional to the size of the future style
interventions 
that we want to guard against for future test data.

\subsection{Estimating the expected conditional variance}\label{subsec:convar}
Recall that $S_j\subseteq \{1,\ldots,\ntot\}$ contains samples with
identical realizations of $(Y,\I)$ for $j\in\{1,\ldots,\nid\}$.  For
each $j\in \{1,\ldots,\nid\}$, define $\hat{\mu}_{\theta,j}$ as the
arithmetic mean across all $f_\theta(x_{i}), i\in S_j$.
The canonical estimator of the conditional
variance $\hat{C}_{f,1,\theta}$ is then
\begin{align*} \hat{C}_{f,1,\theta} &:= \frac{1}{\nid} \sum_{j =1}^\nid
\frac{1}{|S_j|} \sum_{i\in S_j}
                                      (f_\theta(x_{i})-\hat{\mu}_{\theta,j})^2,\quad
                                      \mbox{where}\;\;
                                      \hat{\mu}_{\theta,j} =
\frac{1}{|S_j|} \sum_{i\in S_j} f_\theta(x_{i})  \end{align*}
and analogously for the conditional-variance-of-loss, defined in Eq.~\eqref{eq:condvarloss}\footnote{The right hand side can also be interpreted as the graph Laplacian
\citep{Belkin2006} of an appropriately weighted graph that fully connects all
observations $i \in S_j$ for each $j\in\{1,\ldots,\nid\}$.}.
If there are no groups of samples that share the same identifier
$(y,\i)$, we  define $\hat{C}_{f,1,\theta}$ to vanish. The \core estimator is then identical to pooled estimation in this
special case.

\subsection{Motivating examples (continued)}\label{sec:ex_class}
\begin{figure}
\centering
\subfloat[Example 1, training set.]{
     \includegraphics[width=.48\textwidth, keepaspectratio=true, trim={4mm 5mm 0 0}, clip]{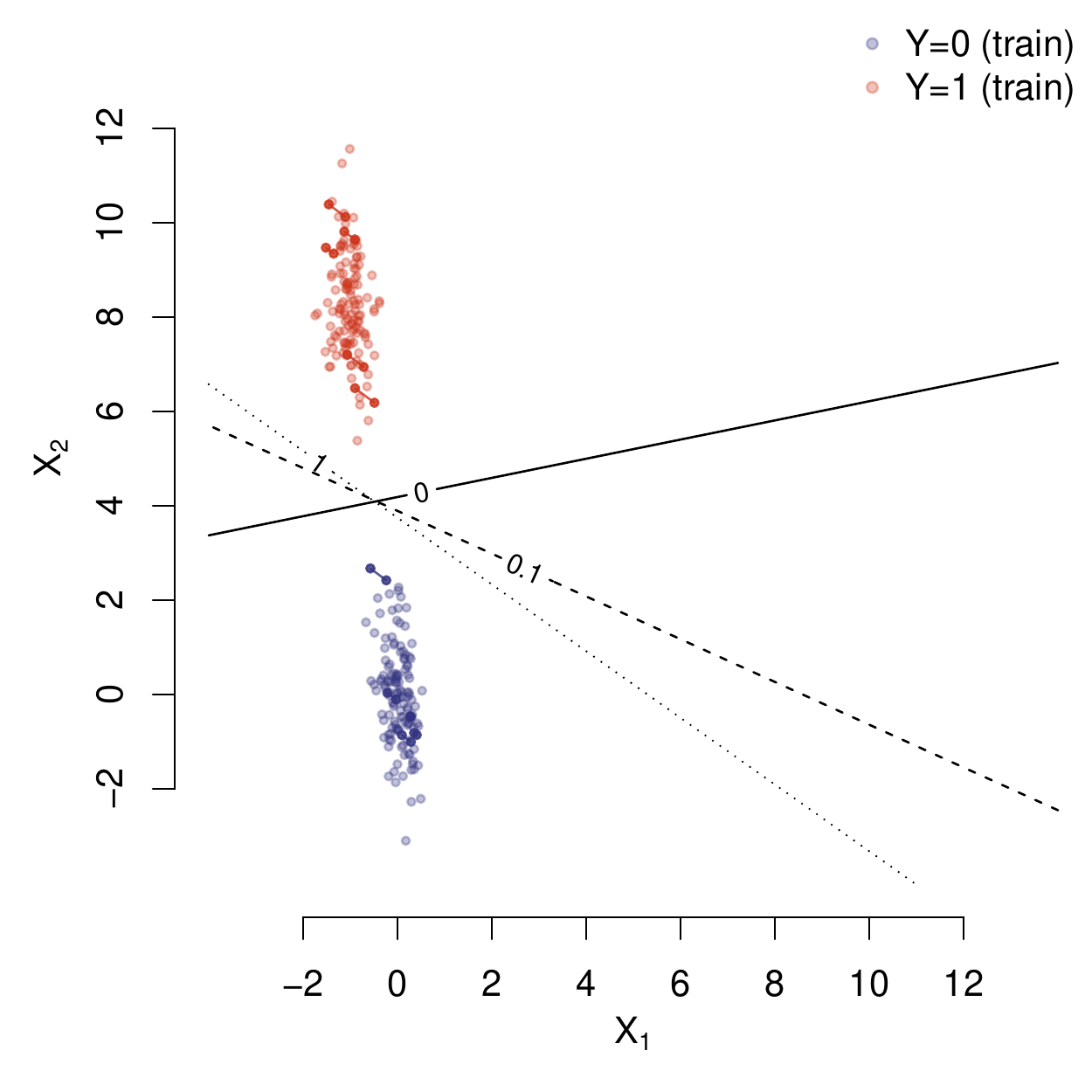}\label{ex1_train_outcome}
}
\subfloat[Example 1, test set.]{
     \includegraphics[width=.48\textwidth, keepaspectratio=true, trim={4mm 5mm 0 0}, clip]{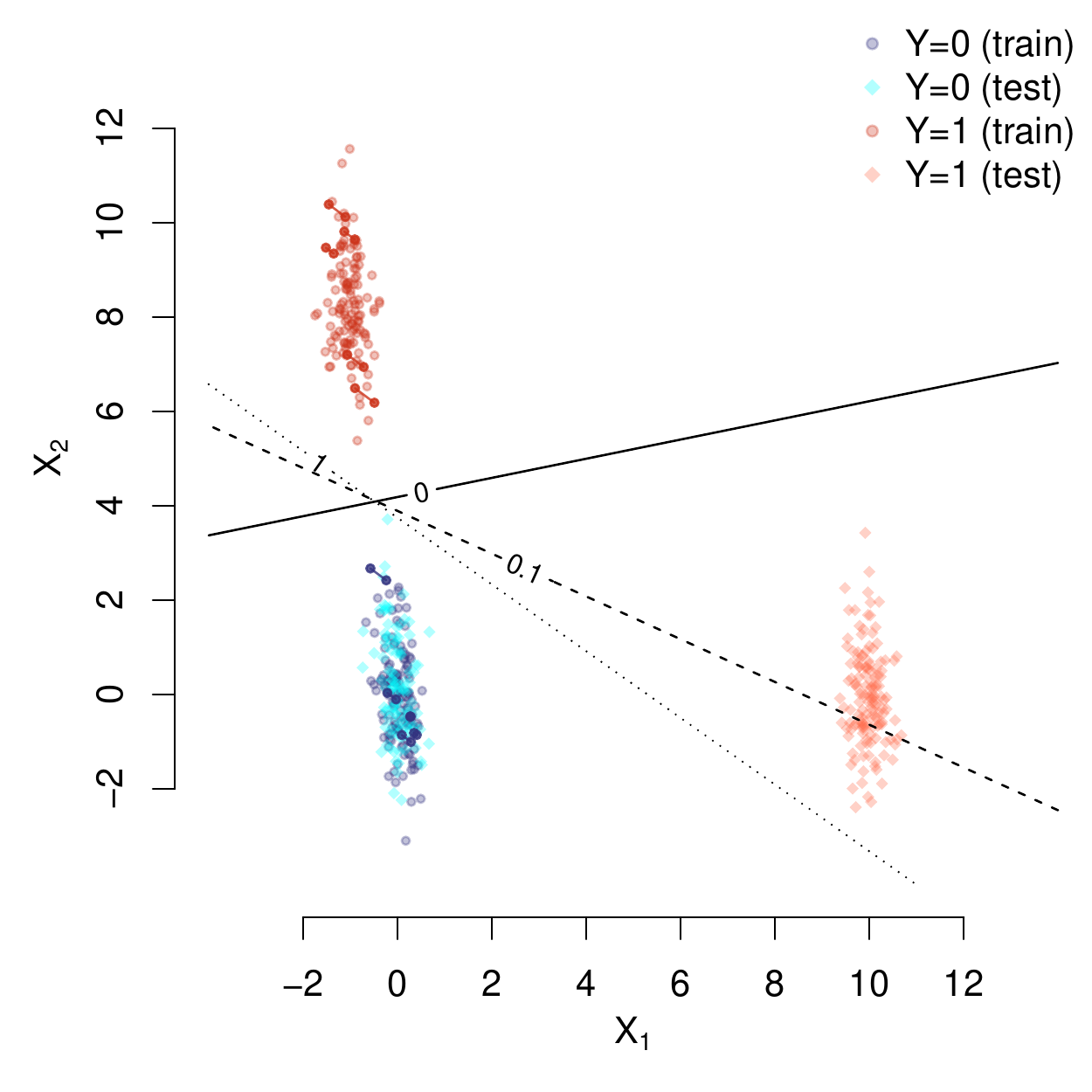}\label{ex1_test_outcome}
}

\subfloat[Example 2, training set.]{
     \includegraphics[width=.48\textwidth, keepaspectratio=true, trim={4mm 5mm 0 0}, clip]{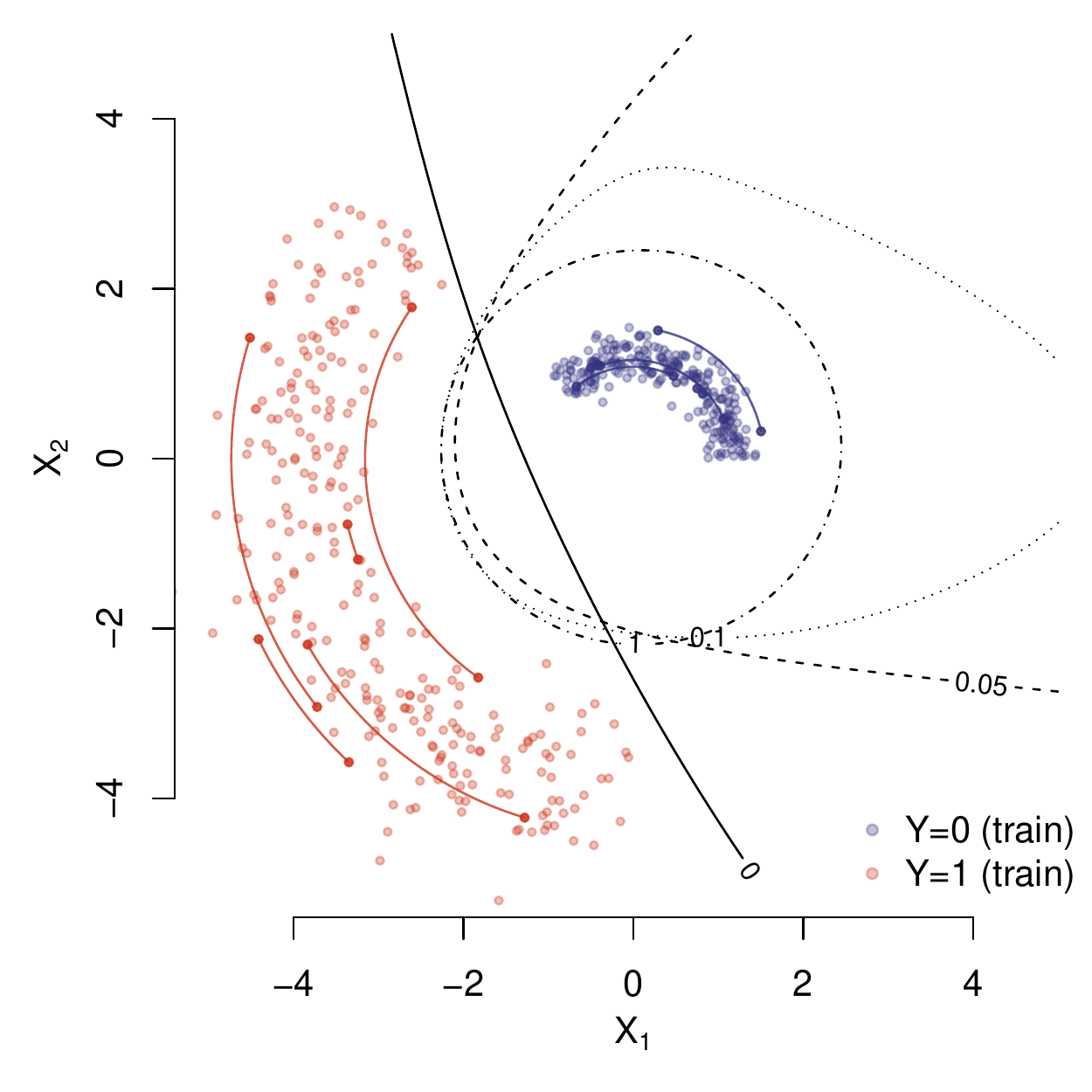}\label{ex2_train_outcome}
}
\subfloat[Example 2, test set.]{
     \includegraphics[width=.48\textwidth, keepaspectratio=true, trim={4mm 5mm 0 0}, clip]{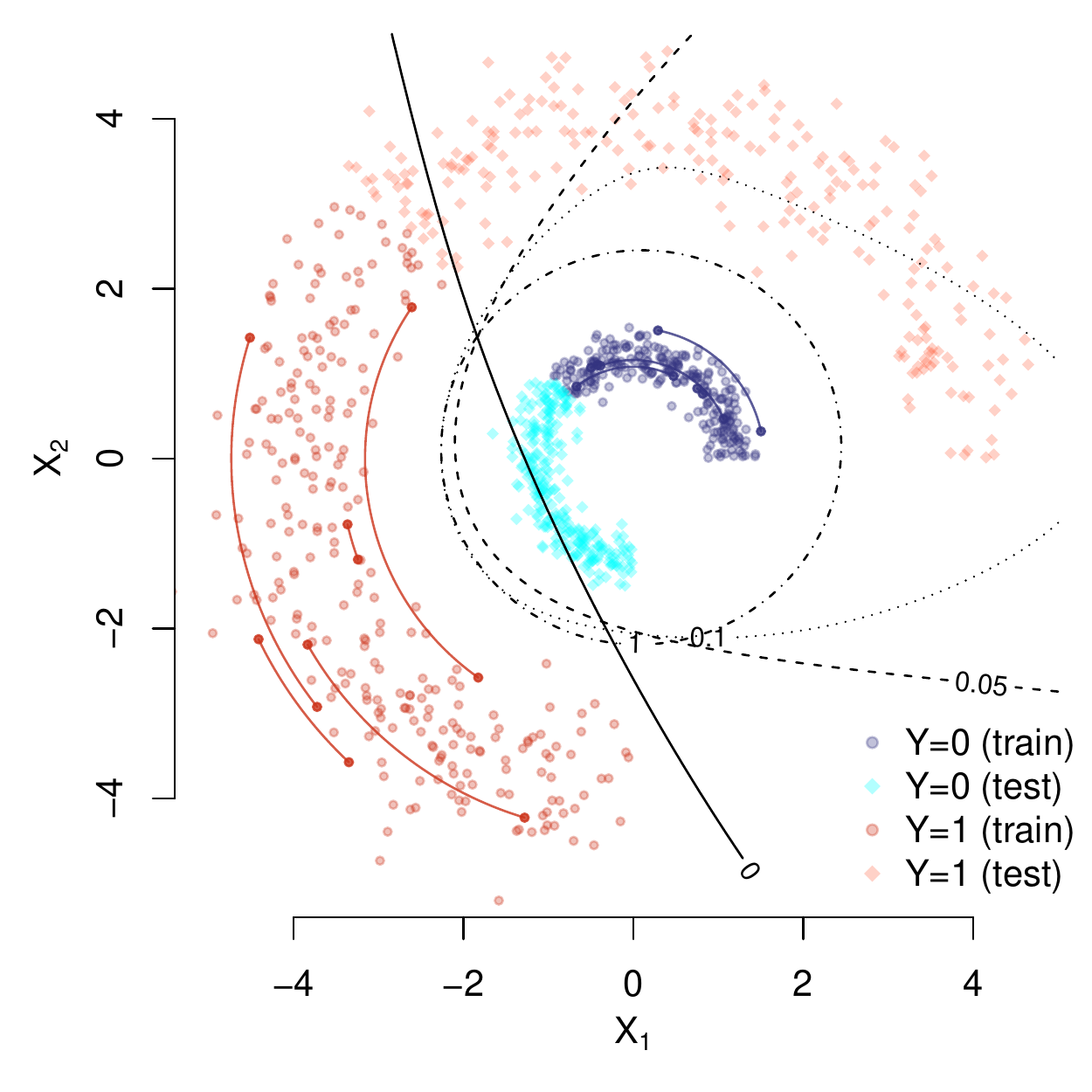}\label{ex2_test_outcome}
}
\captionof{figure}{{\small  The decision boundary as function of the
    penalty parameters $\lambda$ for the examples~1 and~2 from
    Figure~\ref{fig:simple}. There are ten pairs of samples visualized
    that share the
    same identifier $(y,\i)$ and these are connected by a line resp.\ a curve in
    the figures (better visible in panels~(c) and~(d)). The decision boundary associated with a solid line corresponds to
    $\lambda=0$, the standard pooled estimator that ignores the
    groupings. The broken lines are decision boundaries
    for increasingly strong penalties, taking into account the
    groupings in the data. Here, we only show a subsample of the data to avoid overplotting.}}\label{fig:simple_outcome}
\end{figure}

We revisit the first and the second example from
\S\ref{sec:motivating}. Figure~\ref{fig:simple_outcome} shows subsamples of the
respective training and test sets with the estimated decision
boundaries for different values of the penalty parameter
$\lambda$; in both examples, $n=20000$ and $\ncf=500$. Additionally, grouped examples that share the same $(y,\i)$
are visualized: two grouped observations are connected by a
line or curve, respectively. In each example, there are ten such groups visualized (better visible in the nonlinear example).

Panel~\subref{ex1_train_outcome} shows the
linear decision boundaries for $\lambda=0$, equivalent to the
pooled estimator, and for \core with $\lambda\in \{.1, 1\}$. The pooled
estimator misclassifies all test points of class 1 as can be seen in
panel~\subref{ex1_test_outcome}, suffering from a test error of $\approx 51\%$. In contrast, the decision boundary of
the \core estimator with $\lambda=1$ aligns with the direction along which the
grouped observations vary, classifying the test set with almost
perfect accuracy (test error is $\approx 0\%$).

Panels~\subref{ex2_train_outcome}
and~\subref{ex2_test_outcome} show the corresponding plots for the
second example for penalty values $\lambda \in \{0, 0.05, 0.1, 1\}$. While all of them yield good performance on the training set, only a value of $\lambda = 1$, which is associated with a circular decision boundary, achieves almost perfect accuracy on the test set (test error is $\approx 0\%$). The pooled estimator suffers from a test error of $\approx 58\%$.

\section{Domain shift robustness for the \core estimator}\label{sec:theory}
We show two properties of the \core estimator. First, consistency is shown under the risk definition~\eqref{eq:adv_loss} for an infinitely large conditional variance penalty and the logistic loss in a partially linear structural equation model. Second, the population \core estimator is shown to achieve distributional robustness against shift interventions in a first order expansion.

\subsection{Asymptotic domain shift robustness under strong interventions}\label{sec:log_reg}
We analyze the  loss under strong domain shifts, as given in Eq.~\eqref{eq:adv_loss}, for the pooled and the \core estimator in a
one-layer network for binary classification (logistic
regression) in an asymptotic setting of large sample size and strong interventions. 

Assume the structural equation for the image $X\in \mathbb{R}^p$ is
linear in the style features $\style\in \mathbb{R}^q$ (with generally
$p\gg q$) and
we use logistic regression to predict the class label $Y\in \{-1,1\}$. Let
 the interventions $\Delta \in \mathbb{R}^q$ act additively on the style
 features $\style$ (this is only for notational  convenience) and let the
 style features $\style$ act in a linear way on the image $X$ via a matrix
 $\Wmat\in \mathbb{R}^{p\times q}$ (this is an important assumption
 without which results are more involved). The core  or `conditionally
 invariant' features are $\corefeat \in \mathbb{R}^r$, where in
 general  $r\le p$ but this is not important for the following. For
 independent $\varepsilon_Y, \varepsilon_\I,\varepsilon_{\styleo}$ in
$\mathbb{R},\mathbb{R},\mathbb{R}^q$ respectively with
positive density on their support and continuously differentiable
functions $k_y, k_\i,k_{\styleo},k_{\corefeato},k_x$,
\begin{align}
\text{class } &Y \leftarrow k_y(\domain,\varepsilon_Y) \nonumber \\
\text{identifier } & \I \leftarrow k_\i(Y,\varepsilon_\I) \nonumber \\
\text{core or conditionally invariant features }& \corefeat
                                                  \leftarrow
                                                  k_{\corefeato}(Y,\I)
                                                  \nonumber \\
\text{style or orthogonal features }& \style  \leftarrow k_{\styleo}(Y,\I,\varepsilon_{\styleo}) + \Delta \nonumber\\
\text{image }& X  \leftarrow  k_x(\corefeat) +  \Wmat
               \style. \label{eq:modelI}
\end{align}

We assume a logistic regression as a prediction of $Y$ from the image
data $X$:
\[   f_\theta(x) := \frac{ \exp( x^t
  \theta)}{1+\exp(x^t \theta)} .\]
Given training data with $\ntot$ samples, we
estimate $\theta$ with $\hat{\theta}$ and use here a  logistic
loss $\ell_{\theta}(y_i,x_i) = \log(1+ \exp( - y_i ( x_i^t \theta)))$.

The formulation of Theorem~\ref{thm:logistic} relies on the following assumptions.

\begin{assn}\label{assump:logistic} We require the following conditions:
\begin{enumerate}
\item[(A1)] Assume  the conditional distribution $\style|Y=y,\I=\i$
  under the training distribution $F_0$  has positive density (with respect to the Lebesgue measure) in an $\epsilon$-ball in $\ell_2$-norm around the
origin for some $\epsilon>0$ for all $y\in \mathcal{Y}$ and $\i\in \mathcal{I}$.
\item[(A2)] Assume the matrix $\Wmat$ has full rank $q$.
\item[(A3)] Let $M\le n$ be the number of  unique realizations
 among $n$ iid samples of  $(Y,\I)$ and let $p_n:=P(M\le n-q)$. Assume
  that $p_n\to 1$ for $n\to\infty$.
\end{enumerate}
\end{assn}
Assumption (A3) guarantees that  the number $c=\ntot-\nid$ of
  grouped examples  is at least as large as the
  dimension of the style variables. If we have too few or no grouped
  examples (small $c$), we
  cannot estimate the conditional variance accurately.
Under these assumptions  we can prove domain shift robustness.
\begin{theorem}[{\bf Asymptotic domain shift robustness under
     strong interventions}]\label{thm:logistic}
Under model~\eqref{eq:modelI} and Assumption~\ref{assump:logistic}, with probability 1, the
pooled estimator~\eqref{eq:pool} has infinite loss~\eqref{eq:adv_loss}
under arbitrarily large shifts in the distribution of the style features,
\[ L_\infty(\hat{\theta}^{pool}) \;=\; \infty  .\]
The \core estimator~\eqref{eq:hat} $\hat{\theta}^{core}$ with $\lambda \to \infty$
is domain shift robust under strong interventions in the sense that for $\ntot\to\infty$,
\[ L_\infty(\hat{\theta}^{core})
\; \to_p \; \inf_\theta L_\infty(\theta) .\]
\end{theorem}
A proof is given in \S\ref{sec:log_reg_supp}.
The respective ridge penalties in both estimators~\eqref{eq:pool}
and~\eqref{eq:hat} are assumed to be zero for the proof,
but the proof can easily be generalized to include ridge penalties that
vanish sufficiently fast for large sample sizes. The
Lagrangian regularizer $\lambda$ is assumed to be infinite for the
\core estimator to achieve domain shift robustness under these strong interventions. The next section considers the population \core estimator in a setting with weak interventions and finite values of the penalty parameter.

\subsection{Population domain shift robustness under weak interventions}\label{sec:first_order}
The previous theorem states that the \core estimator can achieve
domain shift robustness under strong interventions for an infinitely
strong penalty in an asymptotic setting. An open question is how the loss~\eqref{eq:adv_loss_finite},
\[ L_\xi(\theta) = \sup_{F\in \mathcal{F}_\xi}  E_{F}\Big[
\ell\big(Y,f_\theta(X)\big)\Big] \]
behaves under interventions of small to medium size and
correspondingly smaller values of the penalty. Here, we aim to minimize this loss for a given value of $\xi$ and show that
domain shift robustness can be achieved to first order with the population \core estimator using the
conditional-standard-deviation-of-loss penalty, i.e., Eq.~\eqref{eq:condvarloss} with $\nu = 1/2$,
by choosing an appropriate value of the penalty $\lambda$. Below we will show this appropriate choice of the penalty weight is $\lambda=\sqrt \xi$.

\begin{assn}
\begin{enumerate}
\item[(B1)] Define the loss under a deterministic shift $\delta$ as
  \[ h_\theta(\delta) := E_{F_\theta}[\ell(Y,f_\theta(X))],\]
where the expectation is with respect to random $(\I,Y,\tstyle)\sim
F_\theta$, with $F_\theta$ defined by the deterministic shift intervention
$\tstyle =\style +\delta$ and $(\I,Y,\tstyle)\sim F_0$.
  Assume that for
  all $\theta\in \Theta$, $h_\theta(\delta)$ is twice
  continuously differentiable with bounded   second derivative for a
  deterministic shift $\delta \in \mathbb{R}^q$.
\item[(B2)]
The spectral norm of the conditional variance $\Sigma_{y,\i}$ of $\style|Y,\I$ under
  $F_0$ is assumed to be smaller or equal to some $\zeta \in \mathbb{R}$  for all $y\in
  \mathcal{Y}$ and $\i\in \mathcal{I}$.
\end{enumerate}
\end{assn}
The first assumption (B1) ensures
that the loss is well behaved under interventions on the style
variables. The second assumption (B2) allows to take the limit of
small conditional variances in the style variables.

If setting
$\lambda=\sqrt{\xi}$ and using the conditional-standard-deviation-of-loss penalty, the \core estimator optimizes according to
\[ \hat{\theta}^{core}(\sqrt{\xi})
  \;=\;\text{argmin}_\theta\;   \hat{E}_{F_0}\big[\ell(Y, f_\theta(X))\big]
  \;    + \sqrt{\xi} \cdot \hat{C}_{\ell,1/2,\theta}.\]
The next theorem shows that this is to first order equivalent to
minimizing the worst-case loss over the distribution class $\mathcal{F}_\xi$. The following result holds for the population \core estimator, see below for a discussion about consistency.
\begin{theorem}\label{th:2}
The supremum of the loss over the class of distribution
$F_\xi$ is to first-order given by the expected loss under
distribution $F_0$ with an additional conditional-standard-deviation-of-loss penalty ${C}_{\ell,1/2,\theta}$
\begin{equation}\label{eq:th2} \sup_{F\in \mathcal{F}_\xi}  E_{F}\big[
\ell\big(Y,f_\theta(X)\big)\big] =E_{F_0}\big[  \ell\big( Y ,f_\theta(X)\big)\big]
 + \sqrt{\xi}\cdot {C}_{\ell,1/2,\theta}+O(\max\{\xi,\zeta\})
 .\end{equation}

\end{theorem}
A proof is given in Appendix \S \ref{sec:proof2}.
The objective of the population \core estimator matches thus to first order
the loss under domain shifts if we set the penalty weight
$\lambda=\sqrt{\xi}$. Larger anticipated domain shifts thus require naturally a
larger penalty $\lambda$ in the \core estimation. The result is
possible as we have chosen the Mahalanobis distance to measure shifts
in the style variable and define $\mathcal{F}_\xi$, ensuring that
the strength of shifts on style variables are measured against the
natural variance on the training distribution $F_0$.

In practice, the choice of $\lambda$ involves a somewhat subjective choice about the strength of the distributional robustness guarantee. A stronger distributional robustness property is traded off against a loss in predictive accuracy if the distribution is not changing in the future. One option for choosing $\lambda$ is to choose the largest penalty weight before the validation loss increases considerably. This approach would provide the best distributional robustness guarantee that keeps the loss of predictive accuracy in the training distribution within a pre-specified bound.

As a caveat, the result takes the limit of small conditional variance
of $\style$ in the training distribution and small additional
interventions. Under larger interventions higher-order terms
could start to dominate, depending on the geometry of the loss
function and $f_\theta$.
A further caveat is that the result looks at
the population \core estimator. For finite sample sizes, we would
optimize a noisy version on the rhs of~\eqref{eq:th2}.
To show domain shift robustness in an asymptotic sense, we
would need additional uniform convergence (in $\theta$) of both the empirical
loss and the conditional variance in that for $n\to \infty$,
\begin{align*}
\sup_\theta | \hat{E}_{F_0}\big[\ell(Y, f_\theta(X))\big]  -
E_{F_0}\big[  \ell\big( Y ,f_\theta(X)\big)\big] | & \rightarrow_p 0, \quad \text{and} \\
\sup_\theta | \hat{C}_{\ell,1/2,\theta} -C_{\ell,1/2,\theta} |& \rightarrow_p 0.
\end{align*}
While this is in general a reasonable assumption to make, the validity of the
assumption will depend on the specific function class and on the
chosen estimator of the conditional variance.

\section{Experiments}\label{sec:experiments}
We perform an array of different experiments, showing the
applicability and advantage of the conditional variance penalty for
two broad settings:
\begin{enumerate}
\item Settings where we {\bf do not know} what the style variables correspond to but still
  want to protect against a change in their distribution in the
  future.  In the examples we show cases where the style variable ranges from fashion (\S\ref{subsec:gender}),
  image quality (\S\ref{subsec:celeba_conf}), movement (\S\ref{subsec:stick}) and brightness (\S\ref{subsec:celeb_brightness}), which are all not known
  explicitly to the method. We also include genuinely unknown style
 variables in \S\ref{subsubsec:small_n}  (in the sense that they are
 unknown not only to the methods but also to us as we did
 not explicitly create the style interventions).
\item Settings where we {\bf do know} what type of style interventions we would like to
  protect against. This is usually dealt with by data augmentation
  (adding images which are, say, rotated or shifted compared to the training data if
  we want to protect against rotations or translations in the test
  data; see for example \citet{Schoelkopf1996}).
  The conditional variance penalty is here exploiting that some
  augmented samples were generated from the same original sample and we
  use as $\I$ variable the index of the original image. We show that
  this approach      generalizes better than simply pooling the
  augmented data, in the sense that we need fewer
  augmented samples to achieve the same test error.
This setting is shown in \S\ref{subsec:data_aug}.
\end{enumerate}
Details of the network architectures can be found in Appendix \S\ref{subsec:architec}. All reported error rates are averaged over five runs of the respective method. A TensorFlow \citep{Abadi2015} implementation of \core can be found at \url{https://github.com/christinaheinze/core}.

\begin{figure*}
\begin{minipage}[t]{0.49\textwidth}
\begin{center}
\fcolorbox{red}{white}{
\hspace{-2mm}
\includegraphics[width=0.15\textwidth]{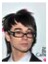}
\includegraphics[width=0.15\textwidth]{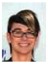}
\includegraphics[width=0.15\textwidth]{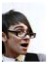}
\includegraphics[width=0.15\textwidth]{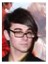}
\includegraphics[width=0.15\textwidth]{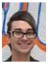}
\includegraphics[width=0.15\textwidth]{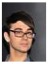}}

\fcolorbox{red}{white}{
\hspace{-2mm}
\includegraphics[width=0.15\textwidth]{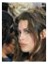}
\includegraphics[width=0.15\textwidth]{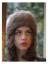}
\includegraphics[width=0.15\textwidth]{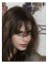}
\includegraphics[width=0.15\textwidth]{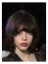}
\includegraphics[width=0.15\textwidth]{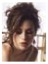}
\includegraphics[width=0.15\textwidth]{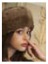}}

\fcolorbox{red}{white}{
\hspace{-2mm}
\includegraphics[width=0.15\textwidth]{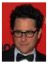}
\includegraphics[width=0.15\textwidth]{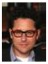}
\includegraphics[width=0.15\textwidth]{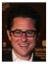}
\includegraphics[width=0.15\textwidth]{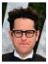}
 \includegraphics[width=0.15\textwidth]{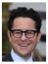}
\includegraphics[width=0.15\textwidth]{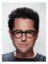}}
\end{center}

\end{minipage}
\hspace{2mm}
\begin{minipage}[t]{0.49\textwidth}
\begin{center}
\includegraphics[width=0.15\textwidth]{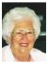}
\includegraphics[width=0.15\textwidth]{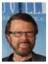}
\includegraphics[width=0.15\textwidth]{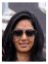}
\includegraphics[width=0.15\textwidth]{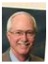}
\includegraphics[width=0.15\textwidth]{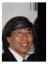}
\includegraphics[width=0.15\textwidth]{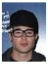}

\includegraphics[width=0.15\textwidth]{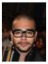}
\includegraphics[width=0.15\textwidth]{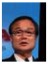}
\includegraphics[width=0.15\textwidth]{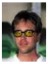}
\includegraphics[width=0.15\textwidth]{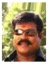}
\includegraphics[width=0.15\textwidth]{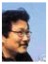}
\includegraphics[width=0.15\textwidth]{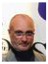}

\includegraphics[width=0.15\textwidth]{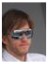}
\includegraphics[width=0.15\textwidth]{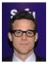}
\includegraphics[width=0.15\textwidth]{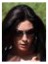}
\includegraphics[width=0.15\textwidth]{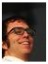}
\includegraphics[width=0.15\textwidth]{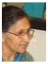}
\includegraphics[width=0.15\textwidth]{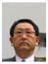}
\end{center}

\end{minipage}
\captionof{figure}{\small Eyeglass detection for CelebA
  dataset with small sample size. The goal is to predict whether a person wears glasses or
  not. Random samples from training and test data are shown. Groups
of observations in the training data that have common $(Y,\I)$ here
correspond to pictures of the same person with either glasses on or
off. These are labelled by red boxes in the training data and the
conditional variance penalty is calculated across these groups of
pictures.}\label{fig:res_celeb1_orig}
\end{figure*}

\subsection{Eyeglasses detection with small sample size}\label{subsubsec:small_n}
In this example, we explore a setting where training and test data are
drawn from the same distribution, so we might not expect a distributional shift
between the two. However, we consider a small training sample size which gives rise to statistical
fluctuations between training and test data. We assess to which
extent the conditional variance penalty can help to improve test
accuracies in this setting.

Specifically, we use a subsample of the CelebA dataset \citep{Liu2015} and
try to classify images according to whether or not  the person in the image
wears glasses. For construction of the
$\I$ variable, we exploit the fact that several photos of the same person are available and set $\I$ to be the
identifier of the person in the dataset.
Figure~\ref{fig:res_celeb1_orig} shows examples from both the
training and the test data set
The conditional variance penalty is estimated
across groups of observations that share a common $(Y,\I)$. Here, this
corresponds to pictures of the same person where all pictures show
the person either with glasses (if $Y=1$) or all pictures show the
person without glasses ($Y=0$).
Statistical fluctuations between training and test set could for instance arise if
by chance the background of eyeglass wearers is darker in the training sample
than in test samples, the eyeglass wearers happen to be outdoors more often or might
be more often female than male etc.

Below, we present the following analyses. First, we look at five different
datasets and analyze the effect of adding the \core penalty (using
conditional-variance-of-prediction) to the cross-entropy loss. Second, we focus on
one dataset and compare the four different variants of the \core penalty in Eqs.~\eqref{eq:condvar} and~\eqref{eq:condvarloss} with $\nu \in \{ 1/2, 1 \}.$

\subsubsection{\core penalty using the conditional variance of the predicted logits}
We consider five different training sets which are created as
follows. For each person in the standard CelebA training data we count
the number of available images and select the 50 identities for which most images
are available individually. We partition these 50 identities into 5
disjoint subsets of size 10 and consider the resulting 5 datasets,
containing the images of 10 unique identities each. The resulting 5 datasets have sizes $\{289, 296, 292, 287, 287\}$. For the validation and the test set, we consider the usual CelebA validation and test split but balance these with respect to the target variable ``Eyeglasses''. The balanced validation set consists of 2766 observations; the balanced test set contains 2578 images. The identities in the validation and test sets are disjoint from the identities in the training sets.

Given a training dataset, the standard approach would be to pool all examples. The only additional information we exploit is that some observations can be grouped.
If using a 5-layer convolutional neural network  with a standard
ridge penalty (details can be found in
Table~\ref{tab:architecture}) and pooling all data, the test error on unseen images ranges from 18.08\% to
25.97\%. Exploiting the group structure with the \core penalty (in
addition to a ridge penalty) results in test errors ranging from
14.79\% to 21.49\%, see Table~\ref{tab:five_datasets_smalln}. The relative improvements when using the \core
penalty range from 9\% to 28.6\%.

\begin{figure}
\centering
\subfloat[]{
     \includegraphics[width=.48\textwidth, keepaspectratio=true]{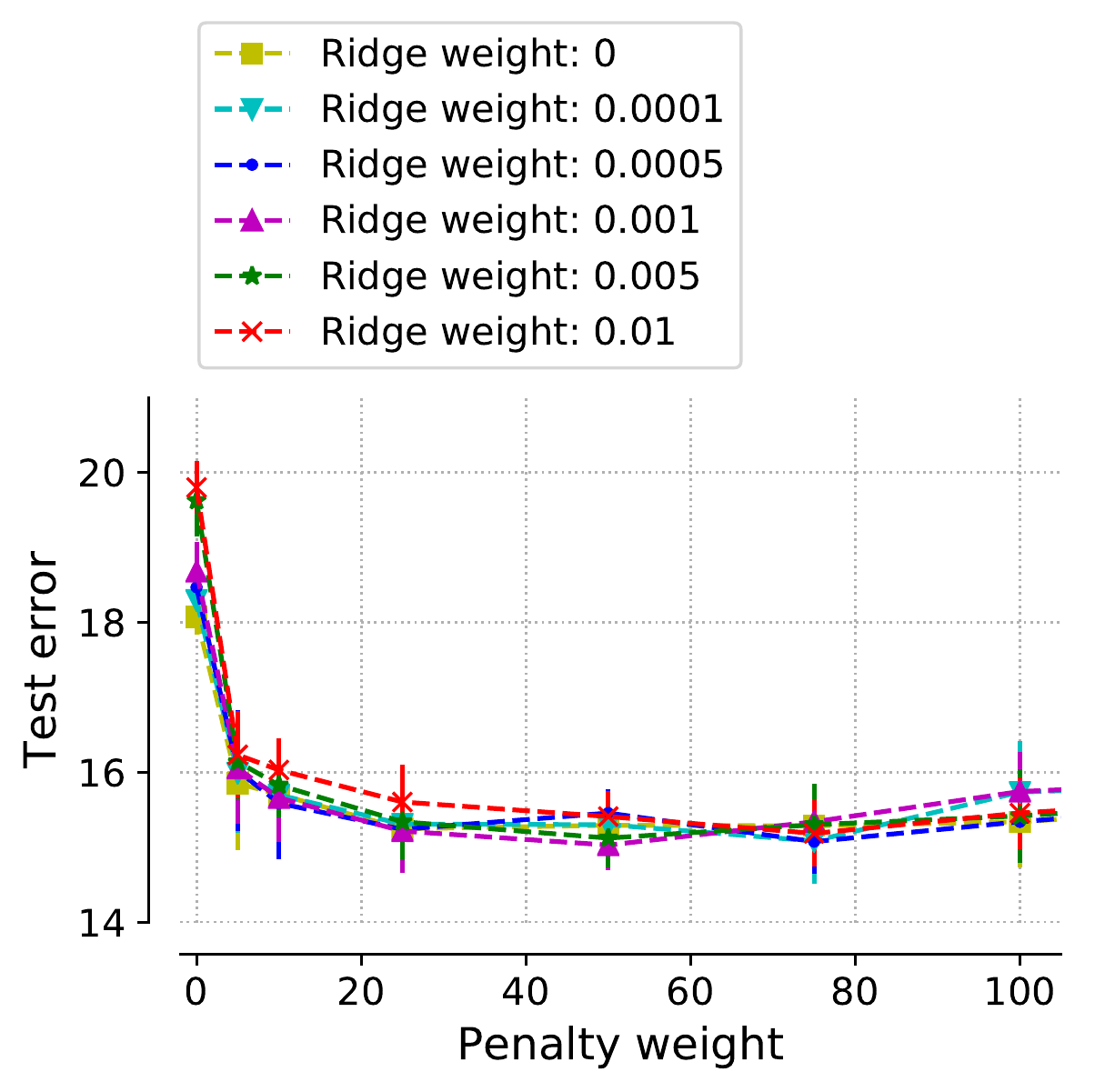}\label{fig:smalln_lambda_error}
}
\subfloat[]{
     \includegraphics[width=.48\textwidth, keepaspectratio=true]{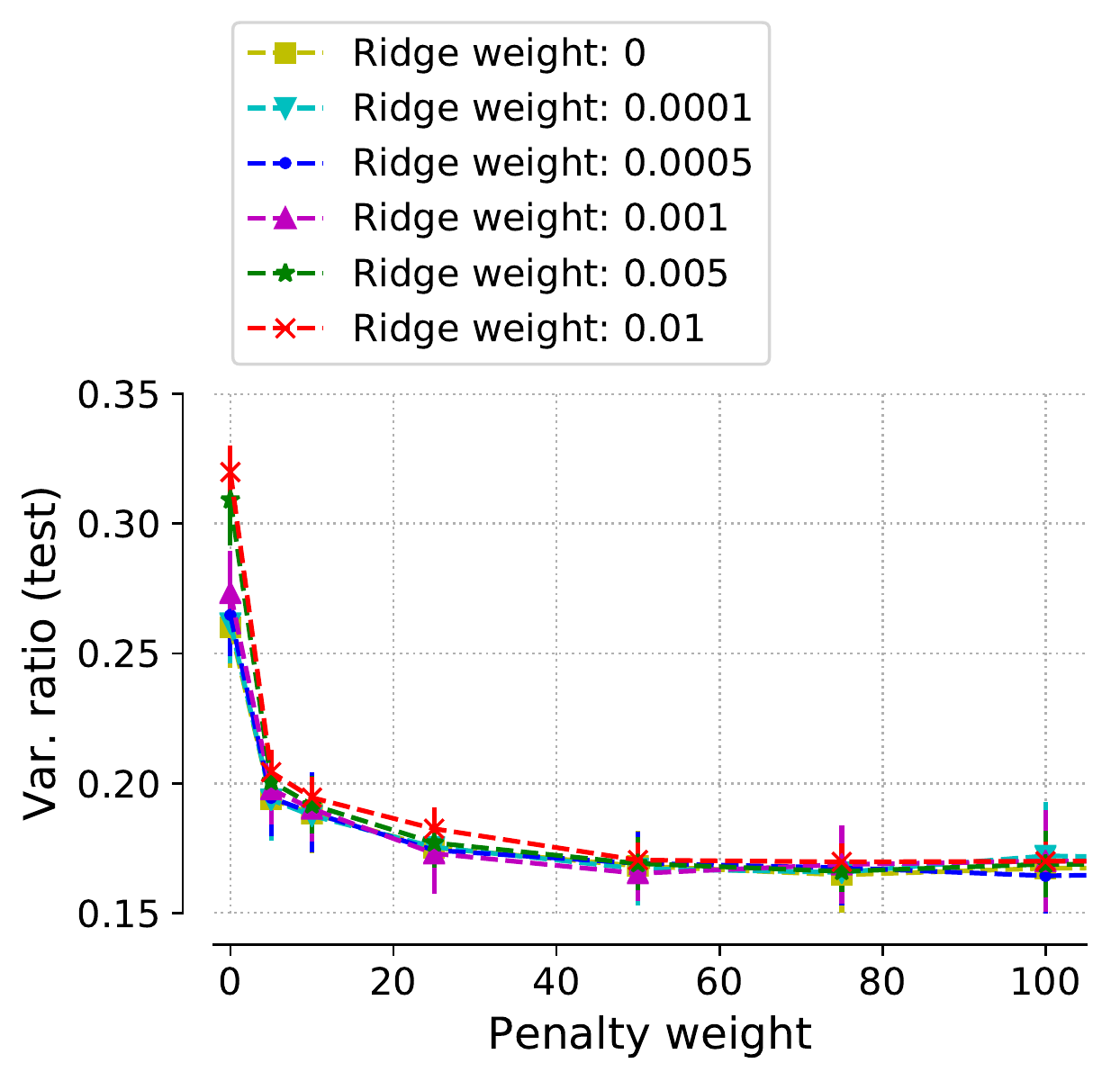}\label{fig:smalln_lambda_var_ratio}
}
\captionof{figure}{{\small Eyeglass detection, trained on a small subset (DS1) of the CelebA dataset with disjoint identities. (a) Average test error as a function of both the \core penalty on
      $x$-axis and various levels of the ridge penalty. The results can
      be seen to be   fairly insensitive to the ridge penalty. (b) The
    variance ratio~\eqref{eq:varratio} on test data as a function of both the \core
  and ridge penalty weights. The \core penalty can be seen to penalize the
  within-group variance selectively, whereas a strong ridge penalty decreases
  both the within- and between-group variance.}}\label{fig:smalln_lambda}
\end{figure}

The test error is not very
sensitive to the weight of the \core penalty as shown in
Figure~\ref{fig:smalln_lambda}\subref{fig:smalln_lambda_error}: for a
large range of penalty weights, adding the \core penalty decreases the
test error compared to the pooled estimator (identical to a \core penalty weight of 0). This holds true for various ridge penalty weights.

While test error rates shown in Figure~\ref{fig:smalln_lambda}
suggests already that the \core penalty differentiates itself clearly
from a standard ridge penalty, we examine next the differential effect
of the \core penalty on the between- and within-group
variances. Concretely, the variance of the predictions can be
decomposed as
\[ \text{Var}(f_\theta(X)) = {E}\big[ {\text{Var}}(f_\theta(X)|Y,\I)
\big]  + {\text{Var}}\big[ {E}(f_\theta(X)|Y,\I) \big] ,\]
where the first term on the rhs is the within-group variance that
\core penalizes, while a ridge penalty would penalize both the within-
and also the between-group variance (the second term on the rhs above).
In Figure~\ref{fig:smalln_lambda}\subref{fig:smalln_lambda_var_ratio} we show the ratio between the \core penalty and the between-group variance where groups are defined by conditioning on $(Y,\I)$. Specifically, the ratio is computed as
\begin{equation}\label{eq:varratio}
\hat{E}\big[ \widehat{\text{Var}}(f_\theta(X)|Y,\I) \big] / \widehat{\text{Var}}\big[ \hat{E}(f_\theta(X)|Y,\I) \big].
\end{equation}
The results shown in Figure~\ref{fig:smalln_lambda}\subref{fig:smalln_lambda_var_ratio} are computed on dataset 1 (DS 1). While increasing ridge penalty weights do
lead to a smaller value of the \core penalty, the between-group
variance is also reduced such that the ratio between the two terms
does not decrease with larger weights of the ridge penalty\footnote{In
  Figure~\ref{fig:smalln_lambda_supp} in the Appendix, the numerator
  and the denominator are plotted separately as a function of the
  \core penalty weight.}. With increasing weight of the \core penalty, the variance ratio decreases, showing that the \core penalty indeed penalizes the within-group variance more than the between-group variance.

\begin{table*}
\begin{center}
{\renewcommand{\arraystretch}{1.4}
\begin{tabular}{cl|cc|cc}
& & \multicolumn{2}{c|}{Error} &  \multicolumn{2}{c}{Penalty value} \\
& Method & Training & Test & Training & Test \\ \hline
\multirow{2}{*}{\rotatebox[origin=c]{90}{DS 1}}
& 5-layer CNN & 0.0\% (0.00\%) & 18.08\% (0.24\%) & 19.14 (1.70) & 18.86 (1.87)  \\
& 5-layer CNN + \core & 0.0\% (0.00\%) & {\bf  15.08\% (0.43\%)} & 0.01 (0.01) & 0.70 (0.05)  \\ \hline
\multirow{2}{*}{\rotatebox[origin=c]{90}{DS 2}}
& 5-layer CNN & 0.0\% (0.00\%) & 23.81\% (0.51\%) & 6.20 (0.35) & 6.97 (0.46)  \\
& 5-layer CNN + \core & 0.0\% (0.00\%) & {\bf 17.00\% (0.75\%)} & 0.00 (0.00) & 0.41 (0.04) \\ \hline
\multirow{2}{*}{\rotatebox[origin=c]{90}{DS 3}}
& 5-layer CNN & 0.0\% (0.00\%) & 18.61\% (0.52\%) & 7.33 (1.40) & 7.91 (1.13) \\
& 5-layer CNN + \core & 0.0\% (0.00\%) & {\bf 14.79\% (0.89\%)} & 0.00 (0.00) &  0.26 (0.03) \\ \hline
\multirow{2}{*}{\rotatebox[origin=c]{90}{DS 4}}
& 5-layer CNN & 0.0\% (0.00\%) & 25.97\% (0.24\%) & 6.19 (0.43) & 7.13 (0.54) \\
& 5-layer CNN + \core & 0.0\% (0.00\%) & {\bf 21.12\% (0.40\%)} & 0.00 (0.00) & 0.63 (0.04) \\ \hline
\multirow{2}{*}{\rotatebox[origin=c]{90}{DS 5}}
& 5-layer CNN & 0.0\% (0.00\%)  &  23.64\% (0.64\%) & 20.20 (2.46) & 24.85 (3.56)  \\
& 5-layer CNN + \core &  0.0\% (0.00\%) &  {\bf 21.49\% (1.27\%)} &  0.00 (0.00) & 0.59 (0.10) \\ \hline
\end{tabular}}
\end{center}
\caption{Eyeglass detection, trained on small subsets (DS1---DS5) of the CelebA dataset with disjoint identities. We report training and test error as well as the value of the \core penalty $\hat{C}_{f,1,\theta}$ on the training and the test set after training, evaluated for both the pooled estimator and the \core estimator. The weights of the ridge and the \core penalty were chosen based on their performance on the validation set.
}\label{tab:five_datasets_smalln}
\end{table*}

Table~\ref{tab:five_datasets_smalln} also reports the value of the \core penalty after training when evaluated for the pooled and the \core estimator on the training and the test set.
As a qualitative measure to assess the presence of sample bias in the data (provided the model assumptions hold), we can compare the value the \core penalty takes after training when evaluated for the pooled estimator and the \core estimator. The difference yields a measure for the extent the respective estimators are functions of $\Delta$. If the respective hold-out values are both small, this would indicate that the style features are not very predictive for the target variable. If, on the other hand, the \core penalty evaluated for the pooled estimator takes a much larger value than for the \core estimator (as in this case), this would indicate the presence of sample bias.

\subsubsection{Other \core penalty types}
We now compare all \core penalty types, i.e., penalizing with (i)
the conditional variance of the predicted logits
$\hat{C}_{f,1,\theta}$, (ii) the conditional standard deviation of the
predicted logits $\hat{C}_{f,1/2,\theta}$, (iii) the conditional
variance of the loss $\hat{C}_{l,1,\theta}$ and (iv) the conditional
standard deviation of the loss $\hat{C}_{l,1/2,\theta}$. For this
comparison, we use the training dataset 1 (DS 1) from
above. Table~\ref{tab:smalln1_all_penalties} contains the test error
(training error was $0\%$ for all methods) as well as the value the
respective \core penalty took after training on the training set and
the test set. The four \core penalty variants' performance differences
are not statistically significant. Hence, we mostly focus on the conditional variance of the predicted logits $\hat{C}_{f,1,\theta}$ in the other experiments.

\begin{table*}
\begin{center}
{\renewcommand{\arraystretch}{1.4}
\begin{tabular}{l|c|cc}
& \multicolumn{1}{c|}{Error} &  \multicolumn{2}{c}{Penalty value} \\
Method & Test & Training & Test \\ \hline
5-layer CNN & 18.08\% (0.24\%) & 19.14 (1.70) & 18.86 (1.87)  \\
5-layer CNN + \core w/ $\hat{C}_{f,1,\theta}$ & {15.08\% (0.43\%)} & 0.01 (0.01) & 0.70 (0.05)  \\
5-layer CNN + \core w/ $\hat{C}_{f,1/2,\theta}$ &  {15.34\% (0.83\%)} & 0.03 (0.01) & 0.89 (0.03)  \\
5-layer CNN + \core w/ $\hat{C}_{l,1,\theta}$ & {15.12\% (0.27\%)} & 0.00 (0.00) & 0.38 (0.03)  \\
5-layer CNN + \core w/ $\hat{C}_{l,1/2,\theta}$ & {15.59\% (0.36\%)} & 0.00 (0.00) & 0.35 (0.02)  \\
\end{tabular}}
\end{center}
\caption{\small Eyeglass detection, trained on a small subset (DS1) of the CelebA dataset with disjoint identities. We report training and test error as well as the value of the \core penalties $\hat{C}_{f,1,\theta}$, $\hat{C}_{f,1/2,\theta}$, $\hat{C}_{l,1,\theta}$ and $\hat{C}_{l,1/2,\theta}$ on the training and the test set after training, evaluated for both the pooled estimator and the \core estimator. The weights of the ridge and the \core penalty were chosen based on their performance on the validation set. The four \core penalty variants' performance differences
are not statistically significant.
}\label{tab:smalln1_all_penalties}
\end{table*}

\subsubsection{Discussion}
While the distributional shift in this example arises due to statistical fluctuations
which will diminish as the sample size grows,
the following examples are more concerned with biases that
will persist even if the number of training and test samples is very
large. A second difference to the subsequent examples is the grouping
structure---in this example, we consider only a few identities, namely
$\nid=10$, with a relatively large number $n_i$ of associated observations
(about thirty observations per individual). In the following examples, $\nid$ is much larger while $\ntot_i$ is typically smaller than five.

\begin{figure*}
\begin{minipage}[t]{0.32\hsize}

Training data ($\ntot=16982$):
\flushright
   \begin{center}
\includegraphics[width=0.17\textwidth]{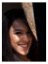}
\includegraphics[width=0.17\textwidth]{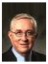}
\includegraphics[width=0.17\textwidth]{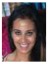}
\includegraphics[width=0.17\textwidth]{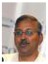}
\includegraphics[width=0.17\textwidth]{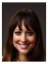}

\fcolorbox{red}{white}{
\hspace{-2mm}  \includegraphics[width=0.17\textwidth]{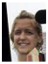}
\includegraphics[width=0.17\textwidth]{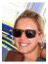}}
\includegraphics[width=0.17\textwidth]{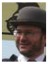}
\includegraphics[width=0.17\textwidth]{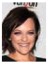}
\includegraphics[width=0.17\textwidth]{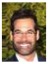}

\includegraphics[width=0.17\textwidth]{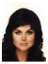}
\includegraphics[width=0.17\textwidth]{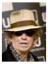}
\includegraphics[width=0.17\textwidth]{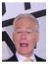}
\fcolorbox{red}{white}{
\hspace{-2mm}  \includegraphics[width=0.17\textwidth]{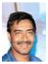}
\includegraphics[width=0.17\textwidth]{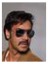}}
\end{center}

\end{minipage}
\begin{minipage}[t]{0.32\hsize}

Test data 1 ($\ntot=4224$):
\flushright

 \begin{center}
\includegraphics[width=0.17\textwidth]{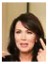}
\includegraphics[width=0.17\textwidth]{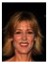}
\includegraphics[width=0.17\textwidth]{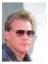}
\includegraphics[width=0.17\textwidth]{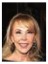}
\includegraphics[width=0.17\textwidth]{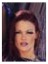}

\includegraphics[width=0.17\textwidth]{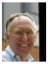}
\includegraphics[width=0.17\textwidth]{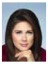}
\includegraphics[width=0.17\textwidth]{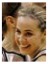}
\includegraphics[width=0.17\textwidth]{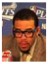}
\includegraphics[width=0.17\textwidth]{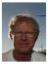}

\includegraphics[width=0.17\textwidth]{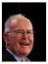}
\includegraphics[width=0.17\textwidth]{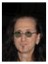}
\includegraphics[width=0.17\textwidth]{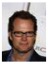}
\includegraphics[width=0.17\textwidth]{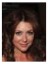}
\includegraphics[width=0.17\textwidth]{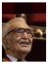}
\end{center}

\end{minipage}
\begin{minipage}[t]{0.32\hsize}
Test data 2 ($\ntot=1120$):
\flushright

\begin{center}
\includegraphics[width=0.17\textwidth]{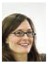}
\includegraphics[width=0.17\textwidth]{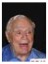}
\includegraphics[width=0.17\textwidth]{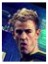}
\includegraphics[width=0.17\textwidth]{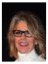}
\includegraphics[width=0.17\textwidth]{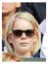}

\includegraphics[width=0.17\textwidth]{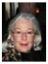}
\includegraphics[width=0.17\textwidth]{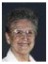}
\includegraphics[width=0.17\textwidth]{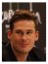}
\includegraphics[width=0.17\textwidth]{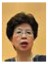}
\includegraphics[width=0.17\textwidth]{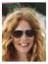}

\includegraphics[width=0.17\textwidth]{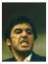}
\includegraphics[width=0.17\textwidth]{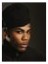}
\includegraphics[width=0.17\textwidth]{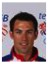}
\includegraphics[width=0.17\textwidth]{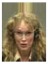}
\includegraphics[width=0.17\textwidth]{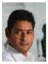}
\end{center}
\end{minipage}
\captionof{figure}{\small  Classification for
  $Y \in \{\text{woman},\text{man}\}$.  There is an unknown confounding here
as men are very likely to wear glasses in training and test set~1
data, while it is women that are likely to wear
glasses in test set~2. Estimators that pool all observations are making use of this confounding and
hence fail for test set~2. The conditional variance penalty for the \core estimator is computed over groups of
images of the same person (and consequently same class label), such as
the images in the red box on the left. The number of grouped examples $\ncf$ is $500$. We vary the proportion of males in the grouped examples between 50\% and 100\% (cf.\ \S\ref{subsec:gender_label_shift}).}\label{fig:res_gender}
\end{figure*}

\begin{figure}
\centering
\subfloat[]{
     \includegraphics[width=.48\textwidth, keepaspectratio=true]{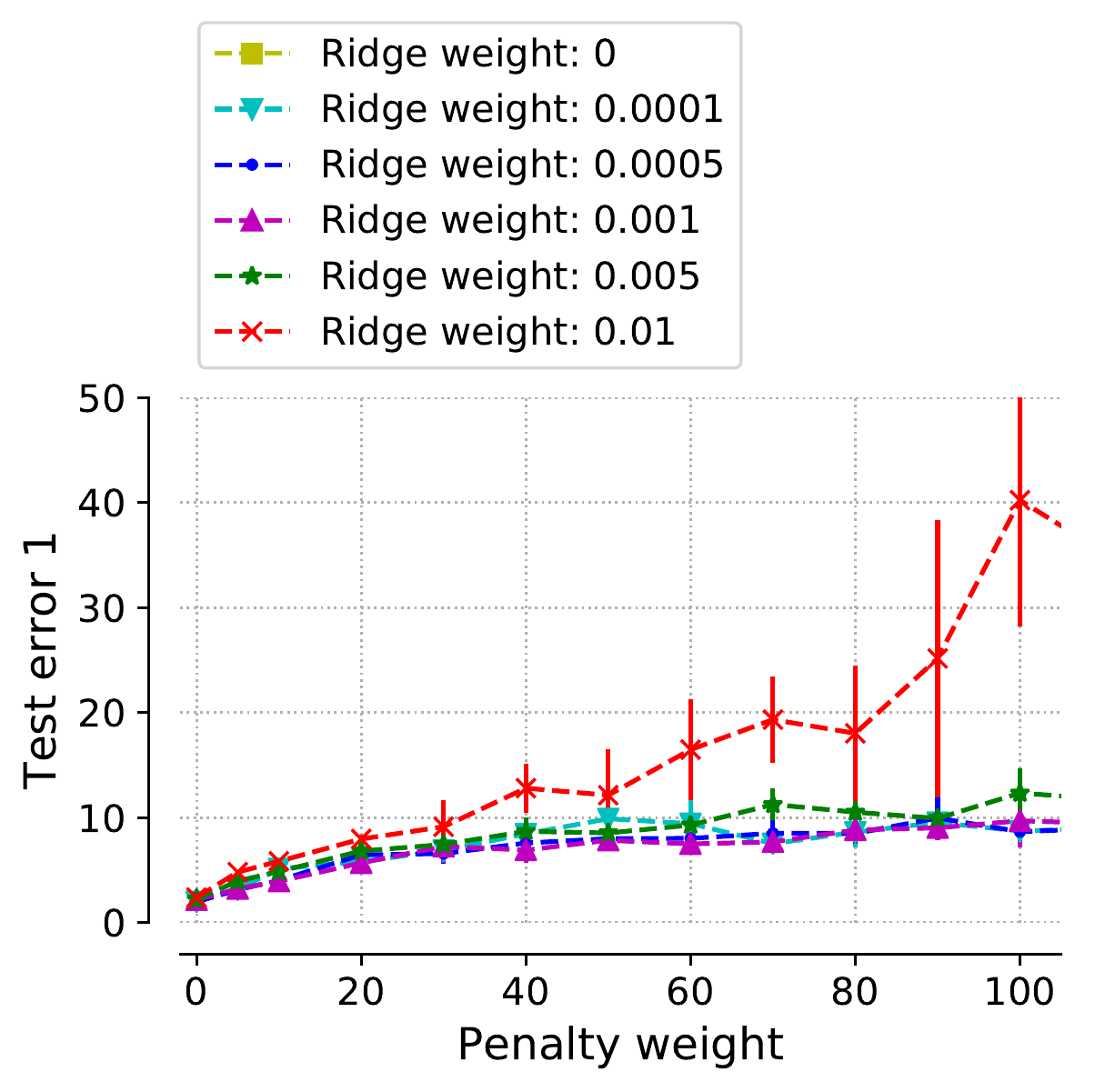}\label{fig:gender_lambda_error1}
}
\subfloat[]{
     \includegraphics[width=.48\textwidth, keepaspectratio=true]{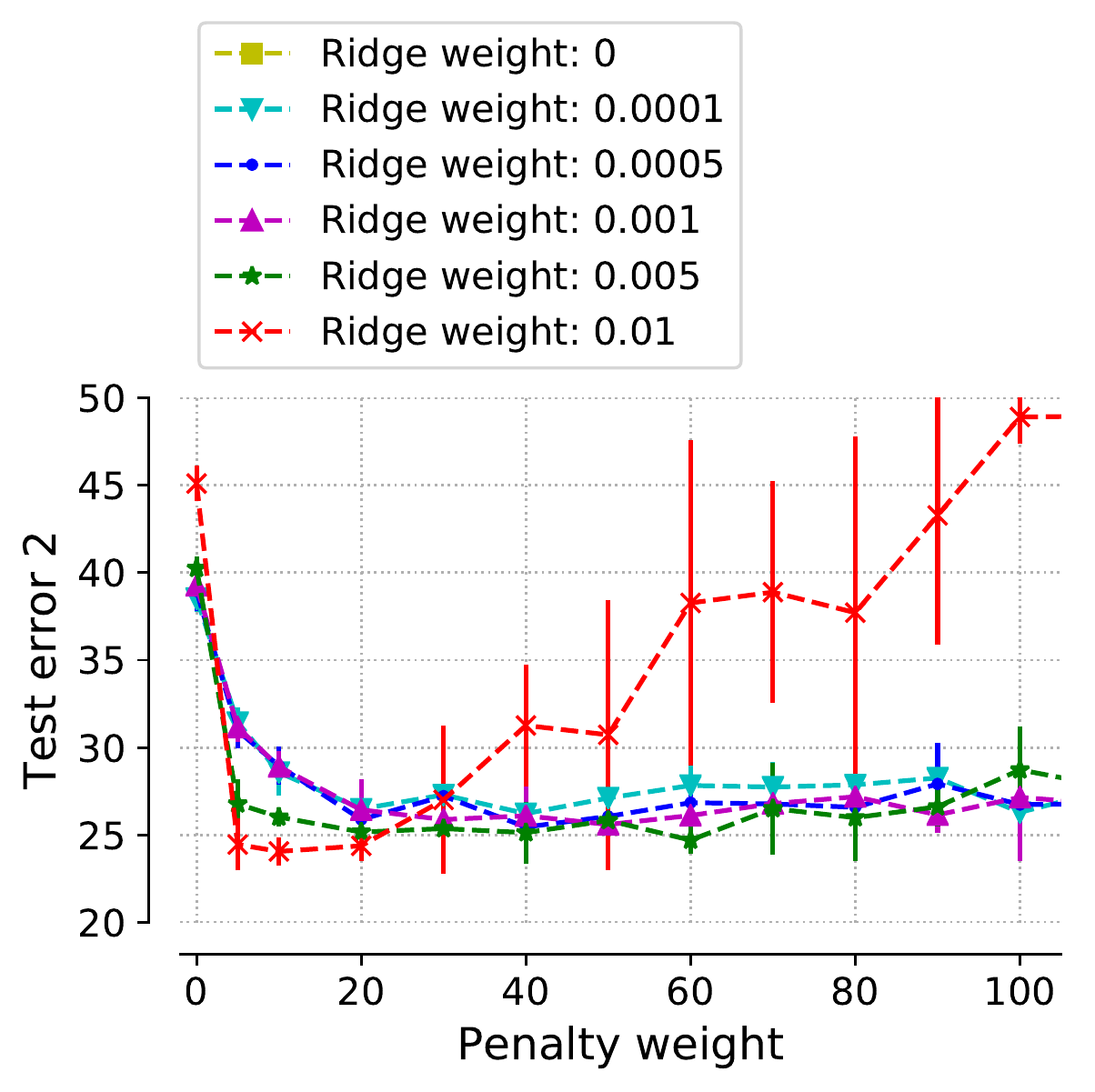}\label{fig:gender_lambda_error2}
}

\subfloat[]{
     \includegraphics[width=.48\textwidth, keepaspectratio=true, trim={0 0 0 122}, clip]{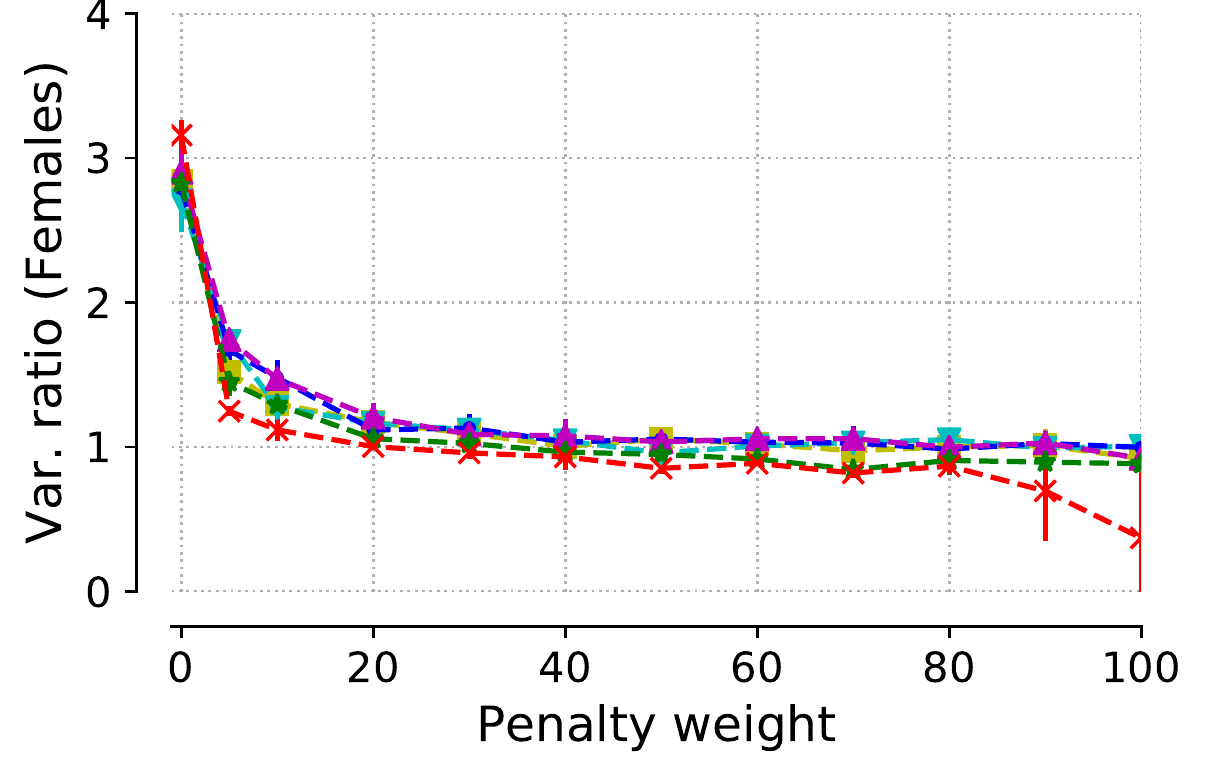}\label{fig:gender_lambda_ratio_fem}
}
\subfloat[]{
     \includegraphics[width=.48\textwidth, keepaspectratio=true, trim={0 0 0 120}, clip]{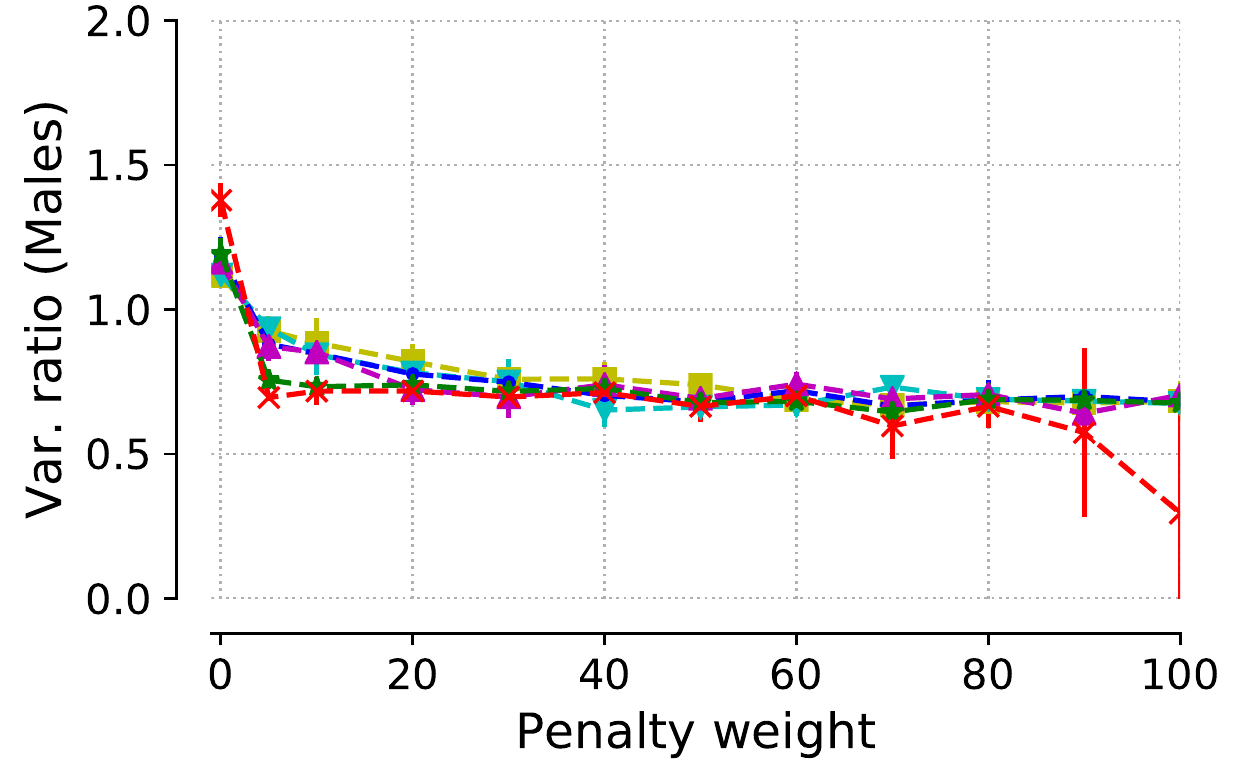}\label{fig:gender_lambda__ratio_male}
}
\captionof{figure}{{\small  Classification for
  $Y \in \{\text{woman},\text{man}\}$ with $\propmales=0.5$. Panels (a) and (b) show the test error on
    test data sets 1 and 2 respectively as a function of the \core and
  ridge penalty. Panels (c) and (d) show the variance
  ratio~\eqref{eq:varratio} (comparing within- and between- group
  variances) for females and males separately.}}\label{fig:gender_lambda_error}
\end{figure}

\subsection{Gender classification with unknown confounding}\label{subsec:gender}
In the following set of experiments, we work again with the CelebA dataset and the 5-layer convolutional neural network architecture described in Table~\ref{tab:architecture}. This time we consider the problem of classifying whether the person shown in the image is male or female. We create a confounding in training and test set~I by including mostly images of men wearing glasses and women not wearing glasses.
In test set~2 the association between gender and glasses is flipped: women always
wear glasses while men never wear glasses. Examples from the
training and test sets~1 and~2 are shown in Figure~\ref{fig:res_gender}. The training set, test set~1 and~2 are subsampled such that they are balanced with respect to $Y$, resulting in 16982, 4224 and 1120 observations, respectively.

To compute the conditional variance penalty, we use again
images of the same person. The $\I$ variable is, in other words, the
identity of the person and gender $Y$ is constant across all examples
with the same $\I$. Conditioning on $(Y,\I)$ is hence identical
to conditioning on $\I$ alone. Another difference to the other experiments is that we consider a binary style feature here.

\subsubsection{Label shift in grouped observations}\label{subsec:gender_label_shift}
We compare six different datasets that vary with respect to the
distribution of $Y$ {\it in the grouped observations}. In all training
datasets, the total number of observations is 16982 and the total number of grouped observations is 500. In the
first dataset, 50\% of the grouped observations correspond to males and 50\% correspond
to females. In the remaining 5 datasets, we increase the number of
grouped observations with $Y=\text{``man''}$, denoted by $\propmales$, to 75\%, 90\%, 95\%, 99\%
and 100\%, respectively. Table~\ref{tab:gender_males_prop} shows the
performance obtained for these datasets when using the pooled
estimator compared to the \core estimator with
$\hat{C}_{f,1,\theta}$. The results show that both the pooled
estimator as well as the \core estimator perform better if the
distribution of $Y$ in the grouped observations is more balanced. The \core estimator improves the
error rate of the pooled estimator by $\approx 28-39\%$ on a relative scale.
Figure~\ref{fig:gender_lambda_error} shows the performance for
$\propmales=50\%$ as a function of the \core penalty
weight. Significant improvements can be obtained across a large range
of values for the \core penalty and the ridge penalty.  Test errors become
more sensitive to the chosen value of the \core penalty for very large
values of the ridge penalty weight as the overall amount of regularization is already large.

\begin{table*}
\begin{center}
{\renewcommand{\arraystretch}{1.4}
\begin{tabular}{cl|ccc|ccc}
& & \multicolumn{3}{c|}{Error} &  \multicolumn{3}{c}{Penalty value} \\
& Method & Train & Test 1 & Test 2 & Train & Test: Females & Test: Males \\ \hline
\multirow{2}{*}{\rotatebox[origin=c]{90}{$\propmales = .5$}}
& 5-layer CNN & {\bf 0.00\%} &  {\bf 2.00\%}  & 38.54\%  & 22.77  & 74.05  & 30.67   \\
& 5-layer CNN + \core & 6.43\%   & 5.85\%  & {\bf 24.07\%} & 0.01   & 1.61 & 0.93   \\ \hline
\multirow{2}{*}{\rotatebox[origin=c]{90}{$\propmales = .75$}}
& 5-layer CNN & {\bf 0.00\%} & {\bf 1.98\%} & 43.41\% & 8.23  & 32.98  & 11.76   \\
& 5-layer CNN + \core & 7.61\%   & 6.99\%   & {\bf 27.05\%}   & 0.00   & 1.44  & 0.62   \\ \hline
\multirow{2}{*}{\rotatebox[origin=c]{90}{$\propmales = .9$}}
& 5-layer CNN & {\bf 0.00\%}   & {\bf 2.00\%}   & 47.64\%  & 9.47  & 40.51   & 14.37   \\
& 5-layer CNN + \core & 8.76\% & 7.74\%  & {\bf 30.63\%}  & 0.00   & 1.26   & 0.42  \\ \hline
\multirow{2}{*}{\rotatebox[origin=c]{90}{$\propmales = .95$}}
& 5-layer CNN & {\bf 0.00\%}   & {\bf 1.89\% }  & 48.96\%  & 13.62   & 61.01   & 21.26 \\
& 5-layer CNN + \core & 10.45\%   & 9.35\%  & {\bf 29.57\%}   & 0.00  & 0.42  & 0.16  \\ \hline
\multirow{2}{*}{\rotatebox[origin=c]{90}{$\propmales = .99$}}
& 5-layer CNN & {\bf 0.00\%}  & {\bf 1.70\% }  & 50.11\%  & 20.66   & 70.80  & 27.80 \\
& 5-layer CNN + \core & 11.10\%  & 10.51\%   & {\bf 32.91\%} &  0.00   &  0.00   &  0.00  \\ \hline
\multirow{2}{*}{\rotatebox[origin=c]{90}{$\propmales = 1$}}
& 5-layer CNN &  {\bf 0.00\%}  & {\bf 1.93\%}  & 49.41\% & 821.32  & 2524.77 & 1253.21  \\
& 5-layer CNN + \core & 11.12\%   & 10.11\%  &{\bf 35.68\%}   & 0.00   & 0.02  & 0.01   \\ \hline
\end{tabular}}
\end{center}
\caption{{\small  Classification for
  $Y \in \{\text{woman},\text{man}\}$. We compare six different datasets that vary with respect to the
distribution of $Y$ {\it in the grouped observations}. Specifically, we vary the proportion of images showing men between $\propmales=0.5$ and $\propmales=1$. In all training
datasets, the total number of observations is 16982 and the total number of grouped observations is 500.  Both the pooled
estimator as well as the \core estimator perform better if the
distribution of $Y$ in the grouped observations is more balanced. The \core estimator improves the
error rate of the pooled estimator by $\approx 28-39\%$ on a relative scale. Table~\ref{tab:gender_males_prop_supp} in the Appendix additionally contains the standard error of all shown results.}}\label{tab:gender_males_prop}
\end{table*}

\subsubsection{Using pre-trained Inception V3 features}
To verify that the above conclusions do not change when using more powerful features, we here compare $\ell_2$-regularized logistic regression using pre-trained Inception V3 features\footnote{Retrieved from \url{https://tfhub.dev/google/imagenet/inception_v3/feature_vector/1}.} with and without the \core penalty. Table~\ref{tab:gender_inception} shows the results for $\propmales=0.5$. While the results show that both the pooled estimator as well as the \core estimator perform better using pre-trained Inception features, the relative improvement with the \core penalty is still 28\% on test set 2.

\begin{table*}
\begin{center}
{\renewcommand{\arraystretch}{1.4}
\begin{tabular}{l|ccc}
 & \multicolumn{3}{c}{Error} 
 \\
 Method & Train & Test 1 & Test 2 
 \\ \hline
Inception V3 & {\bf 5.74\%} & {\bf 5.53\%} & 30.29\%   \\
Inception V3 + \core & 6.15\% & 5.85\% & {\bf 21.70\%} \\
\end{tabular}}
\end{center}
\caption{{\small  Classification for  $Y \in \{\text{woman},\text{man}\}$ with $\propmales=0.5$ Here, we compared $\ell_2$-regularized logistic regression based on Inception V3 features with and without the \core penalty. The \core estimator improves the performance of the pooled estimator by $\approx 28\%$ on a relative scale.}}\label{tab:gender_inception}
\end{table*}

\subsubsection{Additional baselines: Unconditional variance regularization and grouping by class label}\label{sec:baselines_gender}

\begin{figure}
\centering
\subfloat[Baseline: Grouping-by-$Y$]{
     \includegraphics[width=.48\textwidth, keepaspectratio=true]{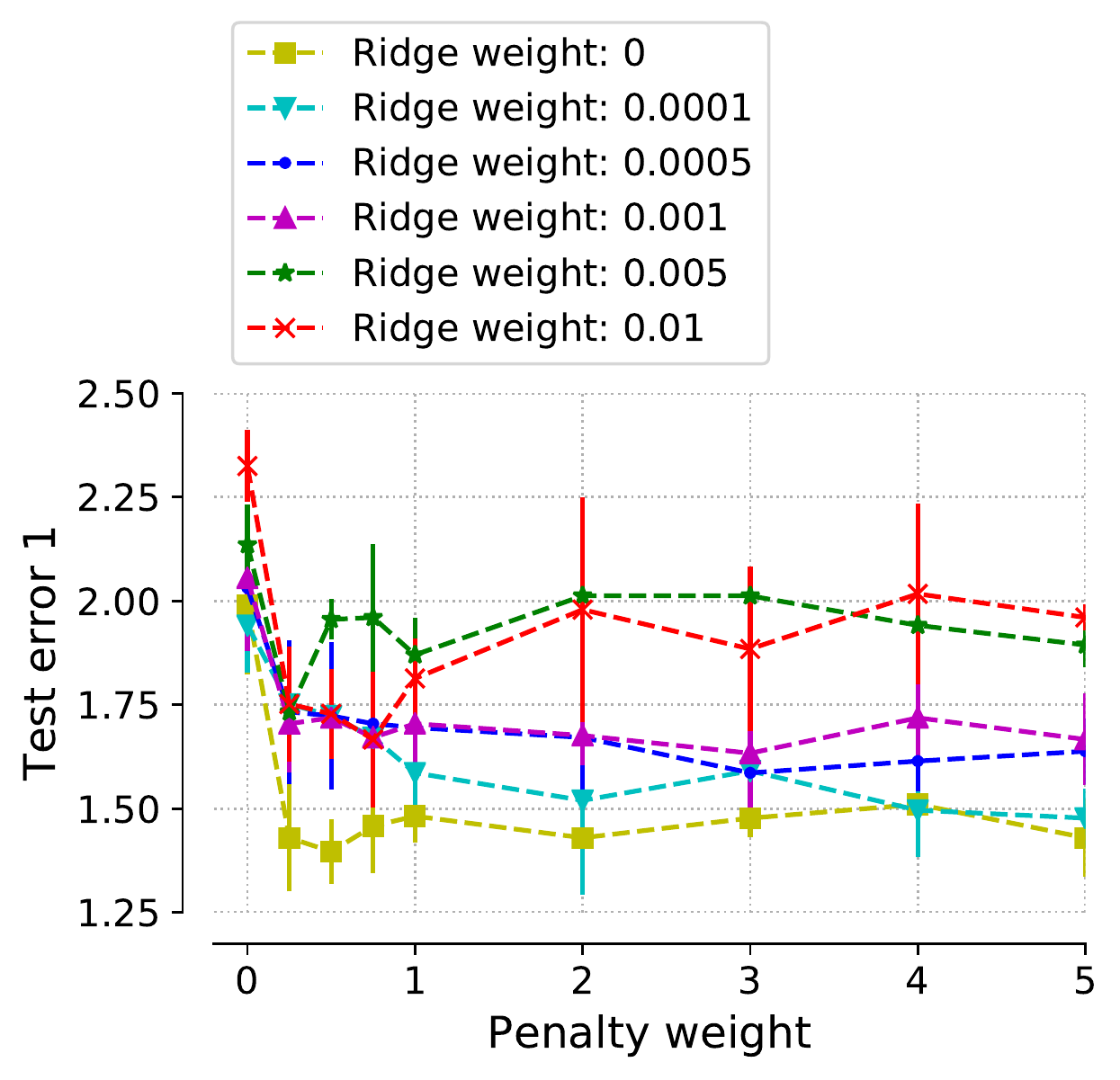}\label{fig:gender_condY_lambda_error1}
}
\subfloat[Baseline: Grouping-by-$Y$]{
     \includegraphics[width=.48\textwidth, keepaspectratio=true]{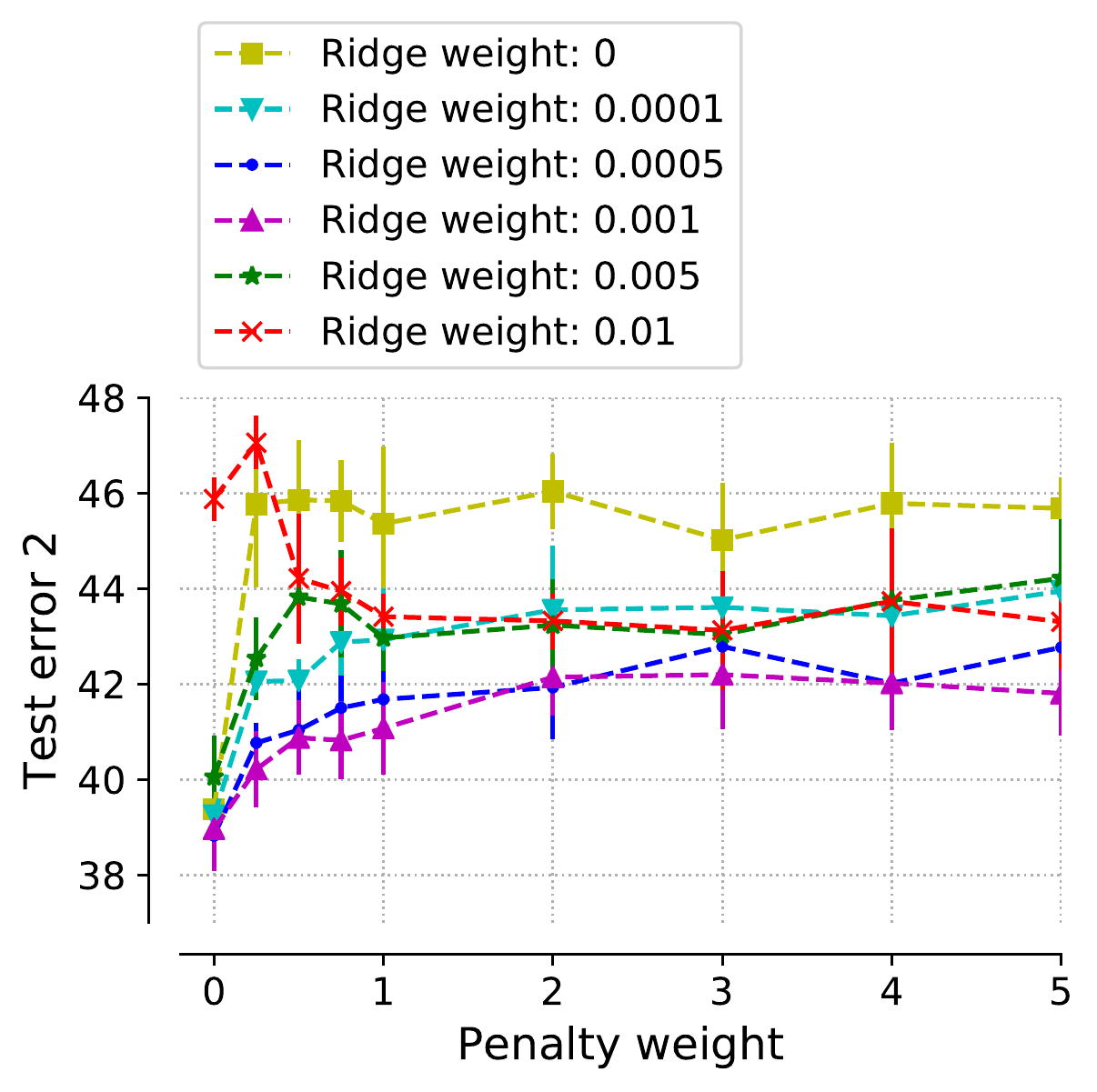}\label{fig:gender_condY_lambda_error2}
}

\subfloat[Baseline: Unconditional variance penalty]{
     \includegraphics[width=.48\textwidth, keepaspectratio=true, trim={0 0 0 120}, clip]{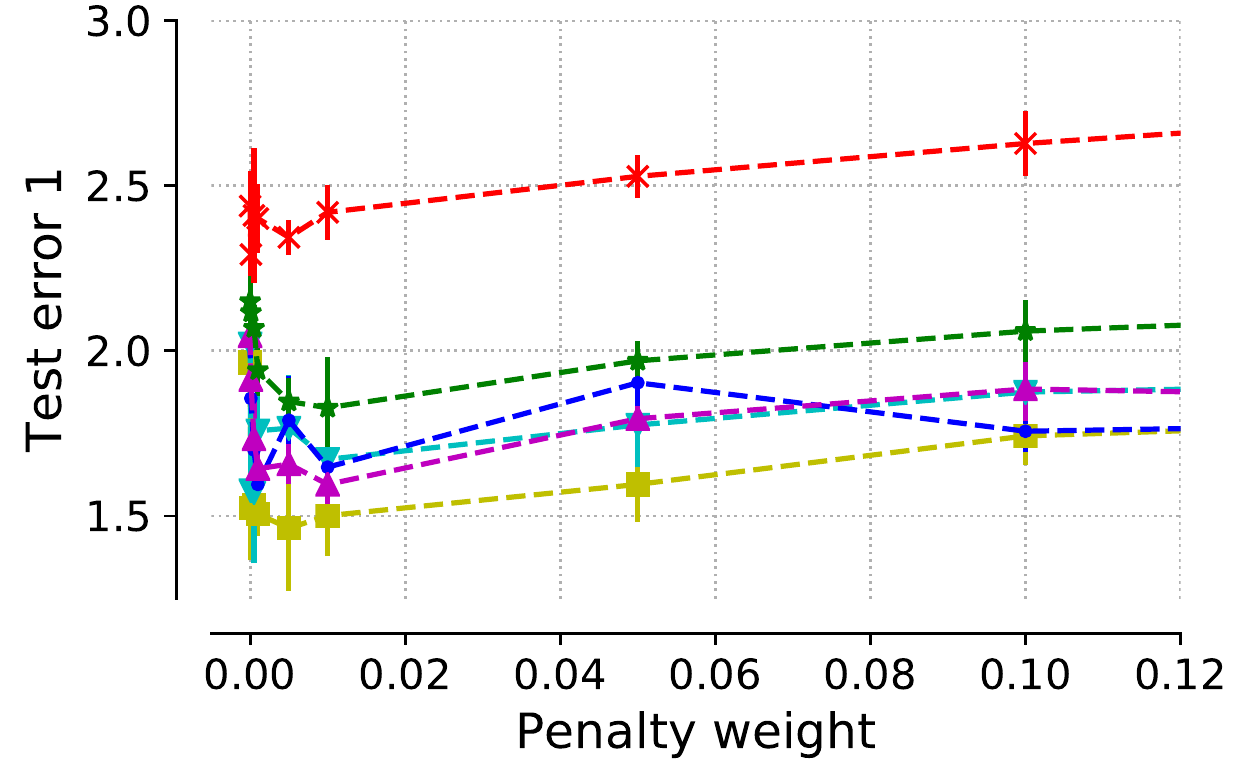}\label{fig:gender_uncond_lambda_error1}
}
\subfloat[Baseline: Unconditional variance penalty]{
     \includegraphics[width=.48\textwidth, keepaspectratio=true, trim={0 0 0 120}, clip]{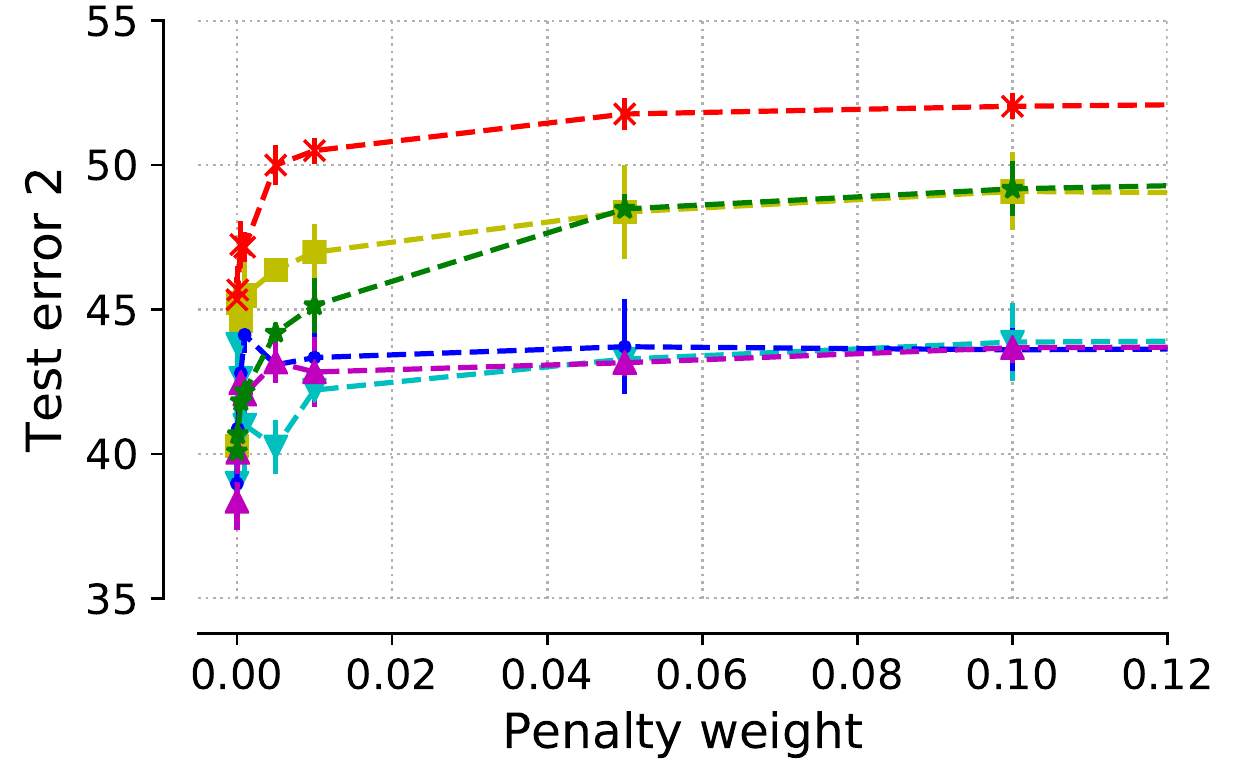}\label{fig:gender_uncond_lambda_error2}
}
\captionof{figure}{{\small Classification for
  $Y \in \{\text{woman},\text{man}\}$ with $\propmales=0.5$, using the baselines which (i) penalize the variance of the predicted logits conditional on the class label $Y$ only; and (ii) penalize the overall variance of the predicted logits (cf.\ \S\ref{sec:baselines_gender}). For baseline (i), panels \protect\subref{fig:gender_condY_lambda_error1} and \protect\subref{fig:gender_condY_lambda_error2} show the test error on
    test data sets 1 and 2 respectively as a function of the ``baseline penalty weight'' for various ridge penalty strengths.  For baseline (ii), the equivalent plots are shown in panels \protect\subref{fig:gender_uncond_lambda_error1} and \protect\subref{fig:gender_uncond_lambda_error2}. In contrast to the \core penalty, regularizing with these two baselines does not yield performance improvements on test set 2, compared to the pooled estimator (corresponding to a penalty weight of 0).}}\label{fig:gender_lambda_error_cond_Y}
\end{figure}

As additional baselines, we consider the following two schemes: (i) we  group all examples sharing the same class label and penalize with the conditional variance of the predicted logits, computed over these two groups; (ii) we penalize the {\it overall} variance of the predicted logits, i.e., a form of unconditional variance regularization.
Figure~\ref{fig:gender_lambda_error_cond_Y} shows the performance of these two approaches.
In contrast to the \core penalty, regularizing with the variance of the predicted logits conditional on $Y$ only does not yield performance improvements on test set 2, compared to the pooled estimator (corresponding to a penalty weight of 0). Interestingly, using baseline (i) without a ridge penalty does yield an improvement on test set I, compared to the pooled estimator with various strengths of the ridge penalty.

\begin{figure*}
\begin{minipage}[t]{0.32\textwidth}
Training data ($\ntot=20000$):
\flushright
{\small
5-layer CNN training error: 0\%\\
with add.\ \core  penalty: 10\% }
\begin{center}
\includegraphics[width=0.17\textwidth]{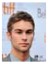}
\includegraphics[width=0.17\textwidth]{figures/imgs_presentation/quality_set2_mu30/train_10636.jpg}
\includegraphics[width=0.17\textwidth]{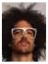}
\includegraphics[width=0.17\textwidth]{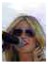}
\includegraphics[width=0.17\textwidth]{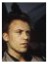}

\fcolorbox{red}{white}{
\hspace{-2mm}
\includegraphics[width=0.17\textwidth]{figures/imgs_presentation/quality_set2_mu30/train_cf_1438_1438.jpg}
\includegraphics[width=0.17\textwidth]{figures/imgs_presentation/quality_set2_mu30/train_cf_1438_16692.jpg}}
\includegraphics[width=0.17\textwidth]{figures/imgs_presentation/quality_set2_mu30/train_13288.jpg}
\includegraphics[width=0.17\textwidth]{figures/imgs_presentation/quality_set2_mu30/train_13914.jpg}
\includegraphics[width=0.17\textwidth]{figures/imgs_presentation/quality_set2_mu30/train_8161.jpg}

\includegraphics[width=0.17\textwidth]{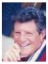}
\includegraphics[width=0.17\textwidth]{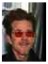}
\includegraphics[width=0.17\textwidth]{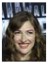}
\fcolorbox{red}{white}{
\hspace{-2mm}
\includegraphics[width=0.17\textwidth]{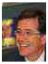}
\includegraphics[width=0.17\textwidth]{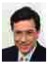}}
\end{center}

\end{minipage}
\begin{minipage}[t]{0.32\textwidth}
Test set 1 ($\ntot=5344$):
\flushright
{\small
5-layer CNN test error: 2\%\\
with add.\ \core penalty: 13\% }
\begin{center}
\includegraphics[width=0.17\textwidth]{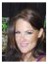}
\includegraphics[width=0.17\textwidth]{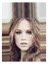}
\includegraphics[width=0.17\textwidth]{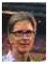}
\includegraphics[width=0.17\textwidth]{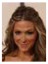}
\includegraphics[width=0.17\textwidth]{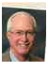}

\includegraphics[width=0.17\textwidth]{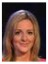}
\includegraphics[width=0.17\textwidth]{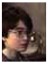}
\includegraphics[width=0.17\textwidth]{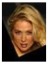}
\includegraphics[width=0.17\textwidth]{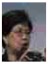}
\includegraphics[width=0.17\textwidth]{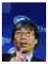}

\includegraphics[width=0.17\textwidth]{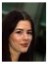}
\includegraphics[width=0.17\textwidth]{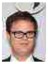}
\includegraphics[width=0.17\textwidth]{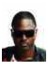}
\includegraphics[width=0.17\textwidth]{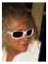}
\includegraphics[width=0.17\textwidth]{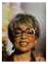}
\end{center}
\end{minipage}
\begin{minipage}[t]{0.32\textwidth}
Test set 2 ($\ntot=5344$):
\flushright
{\small
5-layer CNN test error: 65\%\\
with add.\ \core penalty: 29\% }
\begin{center}
\includegraphics[width=0.17\textwidth]{figures/imgs_presentation/quality_set2_mu30/test2_1735.jpg}
\includegraphics[width=0.17\textwidth]{figures/imgs_presentation/quality_set2_mu30/test2_1783.jpg}
\includegraphics[width=0.17\textwidth]{figures/imgs_presentation/quality_set2_mu30/test2_1813.jpg}
\includegraphics[width=0.17\textwidth]{figures/imgs_presentation/quality_set2_mu30/test2_1852.jpg}
\includegraphics[width=0.17\textwidth]{figures/imgs_presentation/quality_set2_mu30/test2_2082.jpg}

\includegraphics[width=0.17\textwidth]{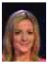}
\includegraphics[width=0.17\textwidth]{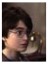}
\includegraphics[width=0.17\textwidth]{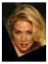}
\includegraphics[width=0.17\textwidth]{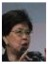}
\includegraphics[width=0.17\textwidth]{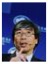}

\includegraphics[width=0.17\textwidth]{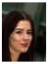}
\includegraphics[width=0.17\textwidth]{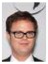}
\includegraphics[width=0.17\textwidth]{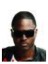}
\includegraphics[width=0.17\textwidth]{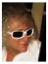}
\includegraphics[width=0.17\textwidth]{figures/imgs_presentation/quality_set2_mu30/test2_5125.jpg}
\end{center}
\end{minipage}
\captionof{figure}{\small Eyeglass detection for CelebA
  dataset  with image
  quality interventions (which are unknown to any procedure used). The
  JPEG compression level is lowered for $Y=1$ (glasses) samples on training
  data and test set~1  and lowered for $Y=0$ (no glasses) samples
  for test set~2. To the human eye, these interventions are barely
  visible but the CNN that uses pooled data without \core penalty has exploited the
  correlation between image quality and outcome $Y$ to achieve a
  (arguably spurious) low test error of 2\% on test set~1. However, if the
  correlation between image quality and $Y$ breaks down, as in test
  set~2, the CNN that uses pooled data without a \core penalty has a 65\% misclassification rate. The training data on the left show paired
  observations in two red boxes: these observations share the same label $Y$ and show the same person $\I$. They are
  used to compute the conditional variance penalty for the \core
  estimator that does not suffer from the same degradation in
  performance for test set~2. }\label{fig:res_celeb1_qual2}
\end{figure*}

\subsection{Eyeglasses detection with known and unknown image quality intervention}\label{subsec:celeba_conf}
We now revisit the third example from \S\ref{sec:motivating}. We again use the
CelebA dataset and consider the problem of classifying whether the
person in the image is wearing eyeglasses. Here, we modify the images in the following way: in
the training set and in test set 1, we  sample the image quality\footnote{We use
ImageMagick (\url{https://www.imagemagick.org}) to change the
quality of the compression through \texttt{convert
  -quality q\_ij input.jpg output.jpg} where $q_{i,j}\sim\mathcal{N}(30,100)$.} for all samples
$\{i:y_i =1\}$ (all samples that show glasses) from a Gaussian
distribution with mean $\mu=30$ and standard deviation $\sigma=10$. Samples with
$y_i=0$ (no glasses) are unmodified.  In other
words, if the image
shows a person wearing glasses, the image quality tends to be
lower.
In test set 2, the quality is reduced in the same way for  $y_i =0$
samples (no glasses),
while images with $y_i =1$ are not changed. Figure~\ref{fig:res_celeb1_qual2} shows examples from the
training set and test sets 1 and 2.
For the \core penalty, we calculate  the
conditional variance across images that share the same $\I$ if $Y=1$, that
is across images that show the same person wearing glasses on
all images. Observations with $Y=0$ (not wearing glasses) are not grouped. Two examples are
shown in the red box of Figure~\ref{fig:res_celeb1_qual2}. Here, we
have $\ncf=5000$ grouped observations among a total sample size of $\ntot=20000$.

Figure~\ref{fig:res_celeb1_qual2} shows misclassification rates for
\core and the pooled estimator on test sets~1 and~2. The pooled estimator (only penalized with an $\ell_2$ penalty) achieves low
error rates of 2\% on test set 1, but suffers from a 65\%
misclassification error on test set 2, as now the relation between $Y$
and the implicit $\style$ variable (image quality) has been flipped. The \core estimator
has a larger error of 13\% on test set 1 as image quality as a feature
is penalized by \core implicitly and the signal is less strong if
image quality has been removed as a dimension. However, in  test set 2
the performance of the \core estimator is 28\% and
improves substantially on the 65\% error of the pooled estimator. The reason is
again the same: the \core penalty ensures that image quality is not
used as a feature to the same extent as for the pooled estimator.
This increases the test error slightly if the samples are generated
from the same distribution as training data (as here for test set~1) but substantially improves
the test error if the distribution of  image quality, conditional on
the class label, is changed on test data (as here for test set~2).

\begin{figure*}
\begin{minipage}[t]{0.32\hsize}
Training data ($\ntot= 20000$):
\flushright
{\small 5-layer CNN training error: 0\%\\
with added \core  penalty:  3\%}
   \begin{center}
\includegraphics[width=0.17\textwidth]{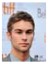}
\includegraphics[width=0.17\textwidth]{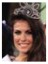}
\includegraphics[width=0.17\textwidth]{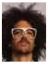}
\includegraphics[width=0.17\textwidth]{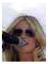}
\includegraphics[width=0.17\textwidth]{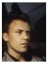}

\fcolorbox{red}{white}{
\hspace{-2mm}
 \includegraphics[width=0.17\textwidth]{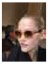}
\includegraphics[width=0.17\textwidth]{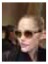}}
\includegraphics[width=0.17\textwidth]{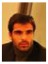}
\includegraphics[width=0.17\textwidth]{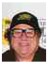}
\includegraphics[width=0.17\textwidth]{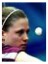}

\includegraphics[width=0.17\textwidth]{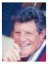}
\includegraphics[width=0.17\textwidth]{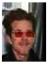}
\includegraphics[width=0.17\textwidth]{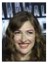}
\fcolorbox{red}{white}{
\hspace{-2mm}
\includegraphics[width=0.17\textwidth]{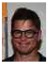}
\includegraphics[width=0.17\textwidth]{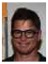}}
\end{center}

\end{minipage}
\begin{minipage}[t]{0.32\hsize}
Test set 1 ($\ntot=5344$):
\flushright
{\small 5-layer CNN test error: 2\%\\
with added \core penalty: 7\%}
\begin{center}
\includegraphics[width=0.17\textwidth]{figures/imgs_presentation/quality_set2_mu30/test1_1735.jpg}
\includegraphics[width=0.17\textwidth]{figures/imgs_presentation/quality_set2_mu30/test1_1783.jpg}
\includegraphics[width=0.17\textwidth]{figures/imgs_presentation/quality_set2_mu30/test1_1813.jpg}
\includegraphics[width=0.17\textwidth]{figures/imgs_presentation/quality_set2_mu30/test1_1852.jpg}
\includegraphics[width=0.17\textwidth]{figures/imgs_presentation/quality_set2_mu30/test1_2082.jpg}

\includegraphics[width=0.17\textwidth]{figures/imgs_presentation/quality_set2_mu30/test1_2133.jpg}
\includegraphics[width=0.17\textwidth]{figures/imgs_presentation/quality_set2_mu30/test1_2319.jpg}
\includegraphics[width=0.17\textwidth]{figures/imgs_presentation/quality_set2_mu30/test1_2544.jpg}
\includegraphics[width=0.17\textwidth]{figures/imgs_presentation/quality_set2_mu30/test1_3800.jpg}
\includegraphics[width=0.17\textwidth]{figures/imgs_presentation/quality_set2_mu30/test1_4036.jpg}

\includegraphics[width=0.17\textwidth]{figures/imgs_presentation/quality_set2_mu30/test1_4148.jpg}
\includegraphics[width=0.17\textwidth]{figures/imgs_presentation/quality_set2_mu30/test1_4486.jpg}
\includegraphics[width=0.17\textwidth]{figures/imgs_presentation/quality_set2_mu30/test1_4614.jpg}
\includegraphics[width=0.17\textwidth]{figures/imgs_presentation/quality_set2_mu30/test1_4764.jpg}
\includegraphics[width=0.17\textwidth]{figures/imgs_presentation/quality_set2_mu30/test1_5125.jpg}
\end{center}

\end{minipage}
\begin{minipage}[t]{0.32\hsize}
Test set 2 ($\ntot=5344$):
\flushright
{\small5-layer CNN test error: 65\%\\
with add.\ \core penalty: 13\%}
\begin{center}
\includegraphics[width=0.17\textwidth]{figures/imgs_presentation/quality_set2_mu30/test2_1735.jpg}
\includegraphics[width=0.17\textwidth]{figures/imgs_presentation/quality_set2_mu30/test2_1783.jpg}
\includegraphics[width=0.17\textwidth]{figures/imgs_presentation/quality_set2_mu30/test2_1813.jpg}
\includegraphics[width=0.17\textwidth]{figures/imgs_presentation/quality_set2_mu30/test2_1852.jpg}
\includegraphics[width=0.17\textwidth]{figures/imgs_presentation/quality_set2_mu30/test2_2082.jpg}

\includegraphics[width=0.17\textwidth]{figures/imgs_presentation/quality_set2_mu30/test2_2133.jpg}
\includegraphics[width=0.17\textwidth]{figures/imgs_presentation/quality_set2_mu30/test2_2319.jpg}
\includegraphics[width=0.17\textwidth]{figures/imgs_presentation/quality_set2_mu30/test2_2544.jpg}
\includegraphics[width=0.17\textwidth]{figures/imgs_presentation/quality_set2_mu30/test2_3800.jpg}
\includegraphics[width=0.17\textwidth]{figures/imgs_presentation/quality_set2_mu30/test2_4036.jpg}

\includegraphics[width=0.17\textwidth]{figures/imgs_presentation/quality_set2_mu30/test2_4148.jpg}
\includegraphics[width=0.17\textwidth]{figures/imgs_presentation/quality_set2_mu30/test2_4486.jpg}
\includegraphics[width=0.17\textwidth]{figures/imgs_presentation/quality_set2_mu30/test2_4614.jpg}
\includegraphics[width=0.17\textwidth]{figures/imgs_presentation/quality_set2_mu30/test2_4764.jpg}
\includegraphics[width=0.17\textwidth]{figures/imgs_presentation/quality_set2_mu30/test2_5125.jpg}
\end{center}
\end{minipage}
\captionof{figure}{\small Eyeglass detection for CelebA
  dataset with image quality interventions.
  The
  only difference to Figure~\ref{fig:res_celeb1_qual2} is in the training data where the paired images now use the
  same underlying image in two different JPEG compressions. The compression level is drawn from the same distribution.
  The \core penalty performs better than for the
  experiment in Figure~\ref{fig:res_celeb1_qual2} since we could explicitly control that only
$\style\equiv\textit{image quality}$ varies between grouped examples.
On the other hand, the performance of
  the pooled estimator is not changed in a noticeable way if we add
  augmented images as the (spurious) correlation  between image
  quality and outcome $Y$ still persists in the
  presence of the extra augmented images. Thus, the pooled estimator
  continues to be susceptible
  to image quality interventions.}\label{fig:res_celeb1_qual3}
\end{figure*}

 \paragraph{Eyeglasses detection with known image quality intervention}
To compare to the above results, we repeat the experiment by changing the grouped observations as follows. Above, we grouped images that had the same person $\I$ when $Y=1$. We refer to this scheme of grouping observations with the same $(Y, \I)$ as `Grouping setting 2'. Here, we use an explicit augmentation
 scheme and augment $c=5000$ images with $Y=1$ in the following way: each
 image is paired with a copy of itself and the image quality is adjusted as described above. In other words, the only difference between the two images is that image quality differs slightly, depending on the value that was drawn from the Gaussian distribution with mean $\mu=30$ and standard deviation $\sigma=10$, determining the strength of the image quality intervention. Both the original and the copy get
 the same value of identifier variable $\I$. We call this grouping scheme `Grouping setting 1'. Compare the left panels
 of Figures~\ref{fig:res_celeb1_qual2} and~\ref{fig:res_celeb1_qual3}
 for examples.

While we used explicit changes in image quality in both
above and here, we referred to grouping setting 2
as `unknown image quality interventions' as the training sample as in
the left panel of Figure~\ref{fig:res_celeb1_qual2} does
not immediately reveal that image quality is the important style
variable. In contrast, the augmented data samples (grouping setting 1) we use
here differ
only in their image quality for a constant  $(Y,\I)$.

 Figure~\ref{fig:res_celeb1_qual3} shows  examples and results. The pooled estimator performs more or less identical to the
 previous dataset. The explicit augmentation did not help as  the
 association between image quality and whether eyeglasses are worn is not changed in the pooled data after including the augmented data
 samples.      The misclassification
 error of the \core estimator is substantially better  than the error rate of the
 pooled estimator. The error rate on test set 2 of 13\% is also improving on the
 rate of 28\% of the \core estimator in
 grouping setting 2.
  We see that using grouping setting 1
works best since we could explicitly control that only
$\style\equiv\textit{image quality}$ varies between grouped examples. In
grouping setting 2, different images of the same person can vary in
many factors, making it more challenging to isolate image quality as the
factor to be invariant against.

\begin{figure*}
\begin{minipage}[t]{0.32\hsize}
Training data ($\ntot=20000$):
\flushright
{\small 5-layer CNN training error: 4\%\\
with added \core  penalty:  4\% }
   \begin{center}
\includegraphics[width=0.17\textwidth]{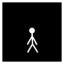}
\includegraphics[width=0.17\textwidth]{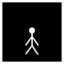}
\includegraphics[width=0.17\textwidth]{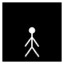}
\fcolorbox{red}{white}{
\hspace{-2mm}
\includegraphics[width=0.17\textwidth]{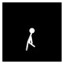}
\includegraphics[width=0.17\textwidth]{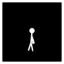}}

\includegraphics[width=0.17\textwidth]{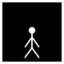}
\includegraphics[width=0.17\textwidth]{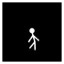}
\includegraphics[width=0.17\textwidth]{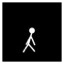}
\includegraphics[width=0.17\textwidth]{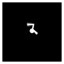}
\includegraphics[width=0.17\textwidth]{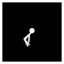}

\fcolorbox{red}{white}{
\hspace{-2mm} \includegraphics[width=0.17\textwidth]{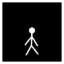}
\includegraphics[width=0.17\textwidth]{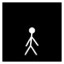}}
\includegraphics[width=0.17\textwidth]{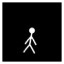}
\includegraphics[width=0.17\textwidth]{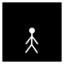}
\includegraphics[width=0.17\textwidth]{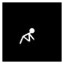}

\end{center}

\end{minipage}
\begin{minipage}[t]{0.32\hsize}
Test set 1 ($\ntot=20000$):
\flushright
{\small  5-layer CNN test error: 3\%\\
with added \core penalty: 4\% }
\begin{center}
\includegraphics[width=0.17\textwidth]{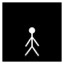}
\includegraphics[width=0.17\textwidth]{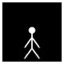}
\includegraphics[width=0.17\textwidth]{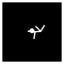}
\includegraphics[width=0.17\textwidth]{figures/imgs_presentation/stick/test1_667.jpg}
\includegraphics[width=0.17\textwidth]{figures/imgs_presentation/stick/test1_679.jpg}

\includegraphics[width=0.17\textwidth]{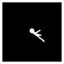}
\includegraphics[width=0.17\textwidth]{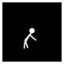}
\includegraphics[width=0.17\textwidth]{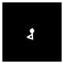}
\includegraphics[width=0.17\textwidth]{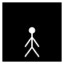}
\includegraphics[width=0.17\textwidth]{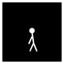}

\includegraphics[width=0.17\textwidth]{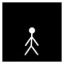}
\includegraphics[width=0.17\textwidth]{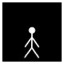}
\includegraphics[width=0.17\textwidth]{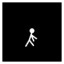}
\includegraphics[width=0.17\textwidth]{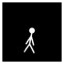}
\includegraphics[width=0.17\textwidth]{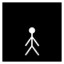}
\end{center}

\end{minipage}
\begin{minipage}[t]{0.32\hsize}
Test set 2 ($\ntot=20000$):
\flushright
{\small 5-layer CNN test error: 41\%\\
with added \core penalty: 9\% }
\begin{center}
\includegraphics[width=0.17\textwidth]{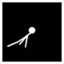}
\includegraphics[width=0.17\textwidth]{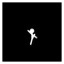}
\includegraphics[width=0.17\textwidth]{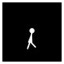}
\includegraphics[width=0.17\textwidth]{figures/imgs_presentation/stick/test2_667.jpg}
\includegraphics[width=0.17\textwidth]{figures/imgs_presentation/stick/test2_679.jpg}

\includegraphics[width=0.17\textwidth]{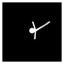}
\includegraphics[width=0.17\textwidth]{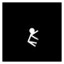}
\includegraphics[width=0.17\textwidth]{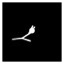}
\includegraphics[width=0.17\textwidth]{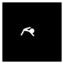}
\includegraphics[width=0.17\textwidth]{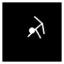}

\includegraphics[width=0.17\textwidth]{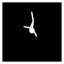}
\includegraphics[width=0.17\textwidth]{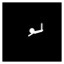}
\includegraphics[width=0.17\textwidth]{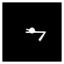}
\includegraphics[width=0.17\textwidth]{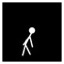}
\includegraphics[width=0.17\textwidth]{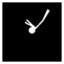}
\end{center}
\end{minipage}
\captionof{figure}{\small  Classification into
  $\{\text{adult},\text{child}\}$ based on stickmen images, where
  children tend to be smaller and adults taller. In training and test
  set 1 data, children tend to have stronger movement whereas adults
  tend to stand still. In test set 2 data, adults show stronger
  movement. The two red boxes in the panel with the training data show two out of the
$c=50$ pairs of examples over which the conditional variance is
calculated. The \core penalty leads to a network that generalizes
better for test set 2 data, where the spurious correlation between age
and movement is reversed, if compared to the training data.}\label{fig:res_stickmenA}
\end{figure*}

\subsection{Stickmen image-based age classification with unknown
  movement interventions}\label{subsec:stick}

In this example we consider synthetically generated stickmen images;
see Figure~\ref{fig:res_stickmenA} for some examples. The target of
interest is $Y \in \{\text{adult}, \text{child}\}$. The core feature
$\corefeat$ is here the height of each person.  The class $Y$ is
causal for {height} and {height} cannot be easily intervened on or
change in different domains. Height  is thus a robust predictor for differentiating between children and
adults. As style feature we have here the movement of a person
(distribution of angles between body, arms and legs). For the training data we created a dependence between {age} and the
style feature `movement', which can
be thought to arise through a
hidden common cause $\domain$, namely the place of observation. The
data generating process is illustrated in
Figure~\ref{fig:DAG_stick}. For instance, the images of children might
mostly show children playing while the images of adults typically show
them in more ``static'' postures. The left panel of Figure~\ref{fig:res_stickmenA} shows examples from the training set where
large movements are associated with children and small movements are
associated with adults. Test set~1 follows the same distribution, as
shown in the middle panel. A standard CNN will exploit this
relationship between movement and the label $Y$ of interest, whereas
this is discouraged by the conditional variance penalty of \core. The latter is pairing images of the same person in slightly different movements
as shown by the red boxes in the leftmost panel of Figure~\ref{fig:res_stickmenA}.  If
the learned model exploits this dependence between movement and age for
predicting $Y$, it will fail when presented images of, say, dancing
adults. The right panel of Figure~\ref{fig:res_stickmenA} shows such
examples (test set 2). The standard CNN suffers in this case from a 41\%
misclassification rate, as opposed to the 3\% on test set~1 data. For as few as $c=50$ paired observations,  the
network with an added \core penalty, in contrast, achieves also 4\% on
test set~1 data and succeeds in achieving an 9\% performance on test set~2, whereas the pooled estimator
fails on this dataset with a test error of 41\%.

These results suggest that the learned
representation of the pooled estimator uses {movement} as a predictor
for {age} while \core does not use this feature due to the
conditional variance regularization. Importantly, including more
grouped examples would not improve the performance of the
pooled estimator as these would  be subject to the same bias and hence
 also predominantly have examples of heavily moving children and ``static'' adults (also see Figure~\ref{fig:res_stick_supp} which shows results for $\ncf \in \{ 20, 500, 2000\}$).

\begin{figure*}
\begin{center}
\begin{minipage}[t]{0.4\hsize}
Training data ($\ntot=10200$):
\flushright
3-layer CNN training error: 0\%\\
with added \core  penalty:  1\%
   \begin{center}

\fcolorbox{red}{white}{
\hspace{-2mm}  \includegraphics[width=0.175\textwidth]{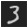}
\includegraphics[width=0.175\textwidth]{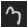}}
\includegraphics[width=0.175\textwidth]{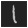}
\includegraphics[width=0.175\textwidth]{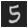}
\includegraphics[width=0.175\textwidth]{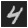}

\includegraphics[width=0.175\textwidth]{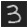}
\includegraphics[width=0.175\textwidth]{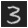}
\fcolorbox{red}{white}{
\hspace{-2mm} \includegraphics[width=0.175\textwidth]{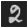}
\includegraphics[width=0.175\textwidth]{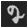}}
\includegraphics[width=0.175\textwidth]{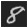}

\includegraphics[width=0.175\textwidth]{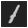}
\includegraphics[width=0.175\textwidth]{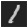}
\includegraphics[width=0.175\textwidth]{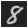}
\includegraphics[width=0.175\textwidth]{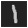}
\includegraphics[width=0.175\textwidth]{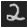}
\end{center}

\end{minipage}
\hspace{4mm}
\begin{minipage}[t]{0.4\hsize}
Test set ($\ntot=10000$):
\flushright
3-layer CNN test error: 22\%\\
with added \core penalty: 10\%
\begin{center}
\includegraphics[width=0.175\textwidth]{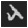}
\includegraphics[width=0.175\textwidth]{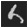}
\includegraphics[width=0.175\textwidth]{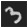}
\includegraphics[width=0.175\textwidth]{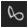}
\includegraphics[width=0.175\textwidth]{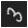}

\includegraphics[width=0.175\textwidth]{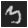}
\includegraphics[width=0.175\textwidth]{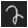}
\includegraphics[width=0.175\textwidth]{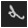}
\includegraphics[width=0.175\textwidth]{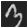}
\includegraphics[width=0.175\textwidth]{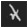}

\includegraphics[width=0.175\textwidth]{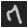}
\includegraphics[width=0.175\textwidth]{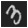}
\includegraphics[width=0.175\textwidth]{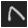}
\includegraphics[width=0.175\textwidth]{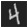}
\includegraphics[width=0.175\textwidth]{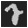}
\end{center}
\end{minipage}
\end{center}
\captionof{figure}{\small Data augmentation for MNIST images. The left 
shows training data with a few rotated images. Evaluating on only
rotated images from the test set, a standard network achieves only
22\% accuracy. We can add the \core penalty by computing the
conditional variance over images that were generated from the same
original image. The test error is then lowered to 10\% on the test
data of rotated images.   }\label{fig:res_mnist}
\end{figure*}

\subsection{MNIST: more sample efficient data
  augmentation}\label{subsec:data_aug}
The goal of using \core in this example is to make data augmentation more efficient in terms of the required samples. In data augmentation, one creates additional samples by modifying the original inputs, e.g.\ by rotating, translating, or flipping the images \citep{Schoelkopf1996}. In other words, additional samples are generated by interventions on style features.
Using this augmented data set for training results in invariance of the estimator with respect to the transformations (style features) of interest. For \core we can use the grouping information that the original and the augmented samples belong to the same object. This enforces the invariance with respect to the style features more strongly compared to normal data augmentation which just pools all samples.
We assess this for the style feature `rotation' on MNIST
\citep{LeCun1998} and only include $\ncf=200$
augmented training examples for $\nid=10000$ original samples,
resulting in a total sample size of $\ntot=10200$. The degree of the
rotations is sampled uniformly at random from
$[35,70]$. Figure~\ref{fig:res_mnist} shows examples from the
training set. By using \core the average test error on rotated
examples is reduced from 22\% to  10\%. Very few augmented sample are
thus sufficient to  lead to stronger rotational invariance. The
standard approach of creating augmented data and pooling all images
requires, in contrast, many more samples to achieve the same effect.  Additional results for $\nid\in
\{1000, 10000\}$ and $\ncf$ ranging from 100 to 5000 can be found in
Figure~\ref{fig:res_mnist_da_supp} in Appendix \S\ref{subsec:mnist_supp}.


\begin{figure*}
\begin{minipage}[t]{0.32\hsize}

Training data ($\ntot=1850$):
\flushright
{\small
5-layer CNN training error: 0\%\\
with added \core penalty: 0\% }
 \begin{center}

\includegraphics[width=0.17\textwidth]{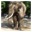}
\includegraphics[width=0.17\textwidth]{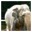}
\includegraphics[width=0.17\textwidth]{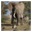}
\includegraphics[width=0.17\textwidth]{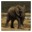}
\includegraphics[width=0.17\textwidth]{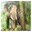}

\fcolorbox{red}{white}{
\hspace{-2mm} \includegraphics[width=0.17\textwidth]{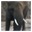}
\includegraphics[width=0.17\textwidth]{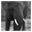}}
\includegraphics[width=0.17\textwidth]{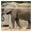}
\includegraphics[width=0.17\textwidth]{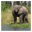}
\includegraphics[width=0.17\textwidth]{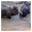}

\includegraphics[width=0.17\textwidth]{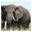}
\includegraphics[width=0.17\textwidth]{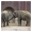}
\includegraphics[width=0.17\textwidth]{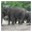}
\includegraphics[width=0.17\textwidth]{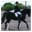}
\includegraphics[width=0.17\textwidth]{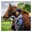}
\end{center}

\end{minipage}
\begin{minipage}[t]{0.32\hsize}

Test data 1 ($\ntot=414$):
\flushright
{\small
5-layer CNN test error: 24\%\\
with add.\ \core penalty: 30\% }

\begin{center}
\includegraphics[width=0.17\textwidth]{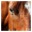}
\includegraphics[width=0.17\textwidth]{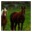}
\includegraphics[width=0.17\textwidth]{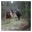}
\includegraphics[width=0.17\textwidth]{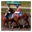}
\includegraphics[width=0.17\textwidth]{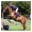}

\includegraphics[width=0.17\textwidth]{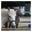}
\includegraphics[width=0.17\textwidth]{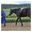}
\includegraphics[width=0.17\textwidth]{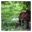}
\includegraphics[width=0.17\textwidth]{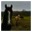}
\includegraphics[width=0.17\textwidth]{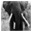}

\includegraphics[width=0.17\textwidth]{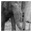}
\includegraphics[width=0.17\textwidth]{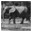}
\includegraphics[width=0.17\textwidth]{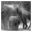}
\includegraphics[width=0.17\textwidth]{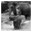}
\includegraphics[width=0.17\textwidth]{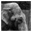}
\end{center}

\end{minipage}
\begin{minipage}[t]{0.32\hsize}
Test data 2 ($\ntot=414$):
\flushright
{\small
5-layer CNN test error: 52\%\\
with add.\ \core penalty: 30\% }

\begin{center}
\includegraphics[width=0.17\textwidth]{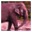}
\includegraphics[width=0.17\textwidth]{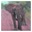}
\includegraphics[width=0.17\textwidth]{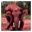}
\includegraphics[width=0.17\textwidth]{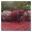}
\includegraphics[width=0.17\textwidth]{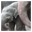}

\includegraphics[width=0.17\textwidth]{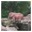}
\includegraphics[width=0.17\textwidth]{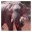}
\includegraphics[width=0.17\textwidth]{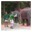}
\includegraphics[width=0.17\textwidth]{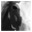}
\includegraphics[width=0.17\textwidth]{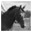}

\includegraphics[width=0.17\textwidth]{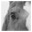}
\includegraphics[width=0.17\textwidth]{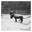}
\includegraphics[width=0.17\textwidth]{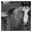}
\includegraphics[width=0.17\textwidth]{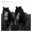}
\includegraphics[width=0.17\textwidth]{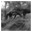}
\end{center}
\end{minipage}

\captionof{figure}{\small Elmer-the-Elephant dataset. The left panel
 shows training data with a few additional grayscale elephants.
The pooled estimator learns that color is predictive for the animal
class and achieves test error of 24\% on test set~1 where this
association is still true but suffers a misclassification error of 53\% on test set~2 where
this association breaks down.
By  adding the \core penalty,  the test error is consistently around
30\%, irrespective of the color distribution of horses and elephants.  }\label{fig:res_ani}

\end{figure*}

\subsection{Elmer the Elephant}\label{subsec:elmer}

In this example, we want to assess whether invariance with respect to
the style feature `color' can be achieved. In the children's book
`Elmer the elephant'\footnote{\url{https://en.wikipedia.org/wiki/Elmer_the_Patchwork_Elephant}} one instance of a colored elephant suffices to recognize it as being an elephant, making the color `gray' no longer an integral part of the object `elephant'.
Motivated by this process of concept formation, we would like to
assess whether \core can exclude `color' from its learned
representation by penalizing conditional variance appropriately.

We work with the `Animals with attributes 2' (AwA2) dataset
\citep{Xian2017} and consider classifying images of horses and
elephants. We include additional examples by adding
grayscale images for $\ncf=250$ images of elephants. These additional
examples do not distinguish themselves strongly from the original
training data as the elephant images are already close to grayscale images.
The total training sample size is 1850.

Figure~\ref{fig:res_ani} shows examples and misclassification rates  from the training set and test
sets for \core and the pooled estimator on different test
sets. Examples from these and more test sets can be found in
Figure~\ref{fig:data_set_ani_supp}. Test set 1 contains original,
colored images only. In test set 2 images of horses are in grayscale
and the colorspace of elephant images is modified, effectively
changing the color gray to red-brown. We observe that the pooled
estimator does not perform well on test set 2 as its learned
representation seems to exploit the fact that `gray' is predictive for
`elephant' in the training set. This association is no longer
valid for test set~2.  In contrast, the predictive performance of \core is hardly affected by the changing color distributions. More details can be found in Appendix \S\ref{subsec:elmer_supp}.

It is noteworthy that a colored elephant can be recognized as an
elephant by adding a few examples of a grayscale elephant to the
very lightly colored pictures of natural elephants. If we just pool
over these examples, there is still a strong bias that elephants are
gray. The \core estimator, in contrast, demands invariance of the prediction for instances of
the same elephant and we can learn color invariance with a few added
grayscale images.

\begin{figure*}
\begin{minipage}[t]{0.32\hsize}
Training data ($\ntot=20000$):
\flushright
{\small
5-layer CNN training error: 0\%\\
with added \core  penalty:  6\% }
   \begin{center}
\includegraphics[width=0.17\textwidth]{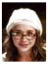}
\includegraphics[width=0.17\textwidth]{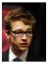}
\includegraphics[width=0.17\textwidth]{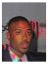}
\includegraphics[width=0.17\textwidth]{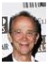}
\includegraphics[width=0.17\textwidth]{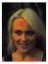}

\fcolorbox{red}{white}{
\hspace{-2mm}  \includegraphics[width=0.17\textwidth]{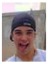}
\includegraphics[width=0.17\textwidth]{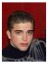}}
\includegraphics[width=0.17\textwidth]{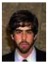}
\includegraphics[width=0.17\textwidth]{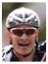}
\includegraphics[width=0.17\textwidth]{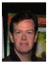}

\includegraphics[width=0.17\textwidth]{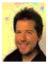}
\includegraphics[width=0.17\textwidth]{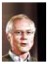}
\includegraphics[width=0.17\textwidth]{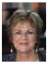}
\fcolorbox{red}{white}{
\hspace{-2mm}  \includegraphics[width=0.17\textwidth]{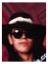}
\includegraphics[width=0.17\textwidth]{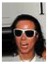}}

\end{center}

\end{minipage}
\begin{minipage}[t]{0.32\hsize}
Test set 1 ($\ntot=5344$):
\flushright
{\small
5-layer CNN test error: 4\%\\
with added \core penalty: 6\% }
\begin{center}
\includegraphics[width=0.17\textwidth]{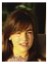}
\includegraphics[width=0.17\textwidth]{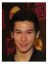}
\includegraphics[width=0.17\textwidth]{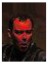}
\includegraphics[width=0.17\textwidth]{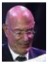}
\includegraphics[width=0.17\textwidth]{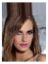}

\includegraphics[width=0.17\textwidth]{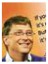}
\includegraphics[width=0.17\textwidth]{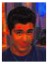}
\includegraphics[width=0.17\textwidth]{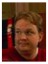}
\includegraphics[width=0.17\textwidth]{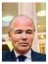}
\includegraphics[width=0.17\textwidth]{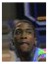}

\includegraphics[width=0.17\textwidth]{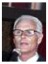}
\includegraphics[width=0.17\textwidth]{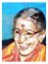}
\includegraphics[width=0.17\textwidth]{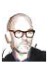}
\includegraphics[width=0.17\textwidth]{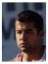}
\includegraphics[width=0.17\textwidth]{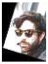}
\end{center}

\end{minipage}
\begin{minipage}[t]{0.32\hsize}
Test set 2 ($\ntot=5344$):
\flushright
{\small
5-layer CNN test error: 37\%\\
with add.\ \core penalty: 25\% }
\begin{center}
\includegraphics[width=0.17\textwidth]{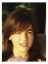}
\includegraphics[width=0.17\textwidth]{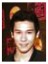}
\includegraphics[width=0.17\textwidth]{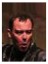}
\includegraphics[width=0.17\textwidth]{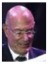}
\includegraphics[width=0.17\textwidth]{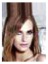}

\includegraphics[width=0.17\textwidth]{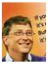}
\includegraphics[width=0.17\textwidth]{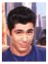}
\includegraphics[width=0.17\textwidth]{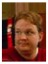}
\includegraphics[width=0.17\textwidth]{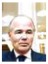}
\includegraphics[width=0.17\textwidth]{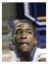}

\includegraphics[width=0.17\textwidth]{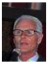}
\includegraphics[width=0.17\textwidth]{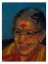}
\includegraphics[width=0.17\textwidth]{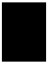}
\includegraphics[width=0.17\textwidth]{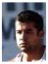}
\includegraphics[width=0.17 \textwidth]{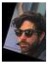}
\end{center}
\end{minipage}
\captionof{figure}{\small Eyeglass detection for CelebA
  dataset with brightness interventions (which are unknown to any procedure
  used). On training data and test set 1 data, images where people
  wear glasses tend to be brighter whereas on test set 2 images where
people do not wear glasses tend to be brighter.}\label{fig:res_celeb1_bright2}

\end{figure*}

\subsection{Eyeglasses detection: unknown brightness intervention}\label{subsec:celeb_brightness}
As in \S\ref{subsec:celeba_conf} we work with the CelebA dataset and
try to  classify whether the person in the image is wearing eyeglasses.
Here we analyze a confounded setting that could arise as follows. Say
the hidden common cause $\domain$ of $Y$ and $\style$ is a binary
variable and indicates
whether the image was taken outdoors or indoors. If it was taken
outdoors, then the person tends to wear (sun-)glasses more often and the image tends to be
brighter. If the image was taken indoors, then the person tends not to
wear (sun-)glasses and the image tends to be darker. In other words, the
style variable $\style$ is here equivalent to brightness and the structure of the data generating process is equivalent to the one shown in Figure~\ref{fig:DAG_stick}.
Figure~\ref{fig:res_celeb1_bright2} shows examples from the training
set and test sets. As previously, we compute the conditional variance
over images of the same person, sharing the same class label (and the \core estimator is hence not
using the knowledge that brightness is important). Two alternatives for constructing grouped observations in this setting are discussed in \S\ref{subsec:celeb_brightness_supp}. We use $\ncf=2000$ and $\ntot=20000$.
For the brightness intervention, we sample the value for the magnitude of the brightness increase resp.\ decrease from an exponential distribution with mean $\meanexp = 20$.
In the training set and test set 1, we sample the brightness value as
$b_{i,j} = [100 + y_i e_{i,j}]_+ $ where $e_{i,j} \sim
\text{Exp}(\meanexp^{-1})$ and $y_i \in \{-1, 1\}$, where  $y_i = 1$
indicates presence of glasses and $y_i=-1$ indicates absence.\footnote{Specifically, we use
ImageMagick (\url{https://www.imagemagick.org}) and modify the
brightness of each image by applying the command
\texttt{convert -modulate b\_ij,100,100 input.jpg output.jpg} to the
image.} For test set~2, we use instead $b_{i,j} = [100 - y_i e_{i,j}]_+
$, so that the relation between brightness and glasses is flipped.

Figure~\ref{fig:res_celeb1_bright2} shows misclassification rates for \core and the pooled estimator on different test sets. Examples from all test sets can be found in Figure~\ref{fig:data_set_celeb_bri_supp}.
First, we notice that the pooled estimator performs better than \core on test set~1. This can be explained by the fact that it can exploit the predictive information contained in the brightness of an image while \core is restricted not to do so.
Second, we observe that the pooled estimator does not perform well on test set~2 as its learned representation seems to use the image's brightness as a predictor for the response which fails when the brightness distribution in the test set differs significantly from the training set. In contrast, the predictive performance of \core is hardly affected by the changing brightness distributions.
Results for $\meanexp \in \{ 5, 10, 20\}$ and $\ncf \in \{200, 5000\}$
can be found in Figure~\ref{fig:res_celeb1_supp} in Appendix \S\ref{subsec:celeb_brightness_supp}.

\section{Further related work}\label{sec:furtherwork}
Encoding certain invariances in estimators is a well-studied area in computer vision and machine learning with an extensive body of literature. While a large part of this work assumes the desired invariance to be known, fewer approaches aim to learn the required invariances from data and the focus often lies on geometric transformations of the input data or explicitly creating augmented observations \citep{Sohn2012, Khasanova2017, Hashimoto2017, Devries2017}. The main difference between this line of work and \core is that we do not require to know the style feature explicitly, the set of possible style features is not restricted to a particular class of transformations and we do not aim to create augmented observations in a generative framework.

Recently, various approaches have been proposed that leverage causal motivations for deep learning or use deep learning for causal inference, related to e.g.\ the problems of cause-effect inference and generative adversarial networks \citep{Chalupka2014, Lopez-Paz2017, Lopez-Paz2017a, Goudet2017, Bahadori2017, Besserve2017, Kocaoglu2017}.

\cite{Kilbertus2017} exploit causal reasoning to characterize fairness considerations in machine learning. Distinguishing between the protected attribute and its proxies, they derive causal non-discrimination criteria. The resulting algorithms avoiding proxy discrimination require classifiers to be constant as a function of the proxy variables in the causal graph, thereby bearing some structural similarity to our style features.

Distinguishing between core and style features can be seen as some form of disentangling factors of variation. Estimating disentangled factors of variation has gathered a lot of interested in the context of generative modeling. 
As in \core, \cite{Bouchacourt2017} exploit grouped observations. In a variational autoencoder framework, they aim to separate style and content---they assume that samples within a group share a common but unknown value for one of the factors of variation while the style can differ. \cite{Denton2017} propose an autoencoder framework to disentangle style and content in videos using an adversarial loss term where the grouping structure induced by clip identity is exploited. Here we try to solve a classification task directly without estimating the latent factors explicitly as in a generative framework.

In the computer vision literature, various works have used identity information to achieve pose invariance in the context of face recognition \citep{Bartlett1996, Tran2017}. More generally, the idea of exploiting various observations of the same underlying object is related to multi-view learning \citep{Xu2013}.
In the context of adversarial examples, \cite{Kannan2018} recently proposed the defense ``Adversarial logit pairing'' which is methodologically equivalent to the \core penalty ${C}_{f,1,\theta}$ when using the squared error loss. Several empirical studies have shown mixed results regarding the performance on $\ell_\infty$ perturbations \citep{Engstrom18, Mosbach18}, so far this setting has not been analyzed theoretically and hence it is an open question whether a \core-type penalty constitutes an effective defense against adversarial examples.

\section{Conclusion}\label{sec:discussion}

Distinguishing the latent features in an image into \emph{core} and
\emph{style} features, we have proposed conditional variance regularization
(\core) to achieve robustness with respect to arbitrarily large
interventions on the  style or ``orthogonal'' features.
The main idea of the \core estimator is to exploit the fact that we
often have instances of the same object in the training data. By
demanding invariance of the classifier amongst a group of instances
that relate to the same object, we can achieve invariance of the
classification performance with respect to interventions
 on style features such as image quality, fashion type, color, or body posture. The training also works despite sampling
biases in the data.

There are two main application areas: 
\begin{enumerate}
\item  If the style features are known
explicitly, we
can achieve the same classification performance as standard data
augmentation approaches with substantially fewer augmented
samples, as shown for example in \S\ref{subsec:data_aug}.
\item
Perhaps more
interesting are settings in which it is unknown what the style
features are, with examples in \S\ref{subsubsec:small_n}, \S\ref{subsec:gender}, \S\ref{subsec:celeba_conf},
\S\ref{subsec:stick} and \S\ref{subsec:celeb_brightness}.
 \core regularization forces  predictions to be based on features
 that do not vary strongly between instances of the same
 object. We could show in the examples and in Theorems~\ref{thm:logistic}
 and~\ref{th:2} that this regularization achieves
 distributional robustness with respect to changes in the distribution
 of the (unknown) style variables.
\end{enumerate}
An interesting line of work would be to use larger models
such as Inception or large ResNet architectures \citep{Szegedy2015, He2016}. These models have been trained to be invariant
to an array of explicitly defined style features. In \S\ref{subsec:gender}
we include results which show that using Inception V3 features does not
guard against interventions on more implicit style features. We would thus like to
assess what benefits \core can bring for training Inception-style models end-to-end,
both in terms of sample efficiency and in terms of generalization performance.

\clearpage
\newpage
\section*{Acknowledgments} We thank Brian McWilliams, Jonas Peters, and Martin
Arjovsky for helpful comments and discussions and CSCS for provision
of computational resources.
A preliminary version of this work was presented at the NIPS 2017
Interpretable ML Symposium
and we thank participants of the symposium for very helpful discussions.
We would also like to thank three anonymous referees and the action
editor Edo  Airoli for detailed and very  helpful feedback on an
earlier version of the manuscript.
\bibliography{bib_new}
\bibliographystyle{iclr2018_conference}

\clearpage
\newpage
\appendix
\begin{center}
{\Large \sc Appendix}
\end{center}
\renewcommand\thefigure{\thesection.\arabic{figure}}
\renewcommand\thetable{\thesection.\arabic{table}}
\setcounter{figure}{0}
\setcounter{table}{0}
\FloatBarrier
\section{Proof of Theorem~\ref{thm:logistic}}\label{sec:log_reg_supp}

{\bf First part.} To show the first part, namely that with probability 1,
\[ L_\infty(\hat{\theta}^{pool}) =\infty  ,\]
we need to show that $\Wmat^t \hat{\theta}^{pool}
\neq {0}$ with probability 1.
The reason this is sufficient is as follows: if $\Wmat^t \theta \neq {0}$, then
$L_\infty(\theta)=\infty$ as we can then find a $v\in
\mathbb{R}^q$ such that $\gamma:= \theta^t \Wmat v \neq 0$.
Assume without limitation of generality that $v$ is normed such that
$E(E(v^t \Sigma_{y,\i}^{-1} v|Y=y,\I=\i))=1$.
Setting $\Delta_\xi =
\xi v$ for $\xi\in \mathbb{R}$, we have that $(\I,Y,\style +
\Delta_\xi)$ is in the class $F_{|\xi|}$ if the distribution of $(\I,Y,\style)$
is equal to $F_0$. Furthermore,
$x(\Delta_\xi)^t \theta = x(\Delta=0)^t \theta + \xi
\gamma$. Hence
$\log(1+\exp(-y\cdot x(\Delta_\xi)^t \theta))\to\infty $ for either $\xi \to \infty$
or $\xi\to -\infty$.

To show that $\Wmat^t \hat{\theta}^{pool}
\neq {0}$ with probability 1, let $\hat{\theta}^*$ be the oracle estimator that is  constrained to be
orthogonal to the column space of $\Wmat$:
\begin{equation} \label{eq:oracle} \hat{\theta}^* = \mbox{argmin}_{\theta: \Wmat^t \theta = {0}} \;
L_\ntot(\theta)  \quad \mbox{  with  } \quad  L_\ntot(\theta) := \frac{1}{\ntot}\sum_{i=1}^\ntot \ell (y_i, f_\theta(x_i)) .\end{equation}
We show $\Wmat^t \hat{\theta}^{pool}
\neq {0}$ by contradiction. Assume hence that  $\Wmat^t \hat{\theta}^{pool}
= {0}$. If this is indeed the case, then the constraint $\Wmat^t
\theta = {0}$ in~\eqref{eq:oracle} becomes non-active and we have
$\hat{\theta}^{pool} = \hat{\theta}^*$. This would imply that
taking the directional derivative of the training loss  with respect to any
$\delta \in \mathbb{R}^p$ in the column space of $\Wmat$ should vanish at the solution
$\hat{\theta}^*$. In other words, define the gradient as
$g(\theta)=\nabla_\theta L_\ntot(\theta) \in \mathbb{R}^p$. The implication is then that for all  $\delta$ in the
column-space of $W$,
\begin{equation}\label{eq:tsw} \delta^t g(\hat{\theta}^*)=0\end{equation}
and we will show the latter condition is violated almost surely.

As we work with the logistic loss and $\mathcal{Y}\in\{-1,1\}$, the
loss is given by
$ \ell (y_i, f_\theta(x_i)) =
\log(1+\exp(- y_i x_i^t\theta)) .$
 Define $r_i(\theta):=y_i /(1+\exp( y_i x_i^t \theta)) $. For all
$i=1,\ldots,\ntot$ we have $r_i\neq 0$.  Then
\begin{equation}\label{eq:dir}  g(\hat{\theta}^*)= \frac{1}{\ntot}\sum_{i=1}^\ntot
  r_i(\hat{\theta}^*) x_{i}.  \end{equation}
The training images can be written according to the model as $x_i =
x^0_i + W \stylesm_i$, where $X^0:= k_x(\corefeat, \varepsilon_X)$ are
the images in absence of any style variation.
 Since the style features only have an effect on the
column space of $\Wmat$ in $X$, the oracle estimator $\hat{\theta}^*$ is
identical under the true training data and the (hypothetical) training
data $x^0_i$, $i=1,\ldots,n$ in absence of style variation. As $X -X^0= \Wmat
\style$, equation~\eqref{eq:dir} can also be written as
\begin{equation}\label{eq:dir2}  \delta^t g(\hat{\theta}^*) = \frac{1}{\ntot}\sum_{i=1}^\ntot
   r_i(\hat{\theta}^*)  (x^0_{i})^t\delta +  \frac{1}{\ntot}\sum_{i=1}^\ntot
   r_i(\hat{\theta}^*) (\stylesm_{i})^t W^t \delta
   .  \end{equation}
Since $\delta$ is in the
column-space of $\Wmat$, there exists $u\in \mathbb{R}^q$  such that
$\delta=\Wmat u$ and we can write~\eqref{eq:dir2} as
\begin{equation}\label{eq:dir3}  \delta^t g(\hat{\theta}^*) = \frac{1}{\ntot}\sum_{i=1}^\ntot
   r_i(\hat{\theta}^*) (x^0_{i})^t W u  +  \frac{1}{\ntot}\sum_{i=1}^\ntot
   r_i(\hat{\theta}^*)  (\stylesm_{i})^t W^t   Wu
   .  \end{equation}
From (A2) we have that the eigenvalues of $\Wmat^t \Wmat$ are all
positive. Also $r_i(\hat{\theta}^*)$ is not a function of the
interventions $\stylesm_{i}$, $i=1,\ldots,n$  since, as
above, the estimator $\hat{\theta}^*$ is identical whether trained on
the original data $x_{i}$ or on the intervention-free
data $x^0_{i}$, $i=1,\ldots, n$. If we condition on everything except for the random
interventions by conditioning on $(x^0_i,y_i)$ for $i=1,\ldots,\ntot$,
then the rhs of~\eqref{eq:dir3}
can be written as
\[ a^t u + B^t u ,\]
where $a\in \mathbb{R}^q$ is fixed (conditionally) and $B=\frac{1}{\ntot}\sum_{i=1}^\ntot
   r_i(\hat{\theta}^*) (\stylesm_{i})^t W^t W\in \mathbb{R}^q$ is a random vector and $B\neq
-a\in\mathbb{R}^q$ with probability~1 by (A1) and (A2)
Hence the left hand side of~\eqref{eq:dir3}  is not
identically 0  with probability 1 for any given $\delta$ in the column-space of
$W$.
 This shows that the implication~\eqref{eq:tsw} is incorrect with
probability 1 and hence completes the
proof of the first part by contradiction.
\smallskip

{\bf Invariant parameter space.}
Before continuing with the second part of the proof,
some definitions. Let $I$ be the \emph{invariant parameter space}
\begin{align*} I &:= \{\theta: f_\theta(x(\Delta)) \mbox{ is constant as function
  of } \Delta \in \mathbb{R}^q \mbox{ for all } x \in \mathbb{R}^p\} .
\end{align*}
 For all $\theta \in I$, the loss~\eqref{eq:adv_loss} for any $F\in {F}_\xi$  is
 identical to the loss under $F_0$. That is for all $\xi\ge 0$,
\[  \text{ if } \theta \in I \text{, then }  \sup_{F\in \mathcal{F}_\xi} E_{F}\Big[
\ell\big(Y, f_\theta\big(X\big)\big)\Big]  =  E_{F_0}\Big[
\ell\big(Y, f_\theta\big(X\big)\big)\Big] .\]
The optimal predictor in the
invariant space $I$ is
\begin{equation}\label{eq:star} \theta^* \;=\; \mbox{argmin}_\theta \;   E_{F_0}\Big[  \ell( Y, f_\theta(X))  \Big]
\mbox{ such that } \theta \in I.\end{equation}
If $f_\theta$ is only a function of the core features $\corefeat$,
then $\theta \in I$. The challenge is that the core features are not
directly observable and we have to infer the invariant space $I$ from
data.

{\bf Second part.} For the second part, we first show that with
probability at least $p_n$, as defined in (A3), $\hat{\theta}^{core}=\hat{\theta}^*$ with
$\hat{\theta}^*$ defined as in~\eqref{eq:oracle}. The
invariant space for this model is the linear subspace
$I=\{\theta: \Wmat^t\theta=0\}$ and by their respective definitions,
\begin{align*}
\hat{\theta}^* &= \mbox{argmin}_{\theta}\;
\frac{1}{\ntot}\sum_{i=1}^\ntot  \ell (y_i, f_\theta(x_{i})) \mbox{   such that }
                 \theta \in I, \\
\hat{\theta}^{core} &= \mbox{argmin}_{\theta}\;
\frac{1}{\ntot}\sum_{i=1}^\ntot \ell (y_i, f_\theta(x_{i})) \mbox{   such that }
                 \theta \in \In.
\end{align*}
Since we  use
$\In=\In(\tau)$ with $\tau=0$,
\[ \In = \big\{\theta: \hat{E}(\hat{\text{Var}}( f_\theta(X)|Y,\I))   =0
\big\}  .\]
This implies that for $\theta\in \In$,  $f_\theta(x_{i})=f_\theta(x_{i'})$ if $i,i'\in
S_j$ for some $j\in\{1,\ldots,\nid\}$\footnote{recall that
$(y_i,\i_i)=(y_{i'},\i_{i'})$ if $i,i'\in S_j$ as the subsets $S_j$,
$j=1,\ldots,m$,
collect all observations that have a unique realization of $(Y,\I)$}.  Since
$f_\theta(x)=f_\theta(x')$ implies $(x-x')^t \theta=0$, it follows that
$ (x_{i} - x_{i'})^t \theta =0 $ if $i,i'\in S_j$ for some $j\in\{1,\ldots,\nid\}$ and
hence
\[ \In \subseteq \big\{\theta: (x_{i} - x_{i'})^t \theta   =0 \text{  if }
i,i'\in S_j \text{ for some } j\in\{1,\ldots,\nid\}
\big\} . \]
 Since $\style$ has a
linear influence on $X$ in~\eqref{eq:modelI},  $x_{i} -
x_{i'}= \Wmat (\Delta_{i}-\Delta_{i'})$ if $i,i'$ are in the same group
$S_j$ of observations for some $j\in\{1,\ldots,\nid\}$.
Note that the number of
grouped examples $\ntot-\nid$ is equal to or exceeds the rank $q$ of
$\Wmat$ with probability $p_n$, using (A3), and $p_n\to 1$ for $n\to\infty$.
By (A2), it follows then with probability at least $p_n$ that $\In \subseteq
\{\theta: \Wmat^t\theta=0\}=I$.
As, by definition,  $I\subseteq \In$ is always true, we have with
probability $p_n$ that $I=\In$. Hence, with probability $p_n$ (and
$p_n\to 1$ for $n\to\infty$),
$\hat{\theta}^{core}=\hat{\theta}^*$.  It thus remains to be shown that
\begin{equation} \label{eq:ts1} L_\infty(\hat{\theta}^*)   \to_p
  \inf_{\theta} L_{\infty}(\theta) .\end{equation}
Since $\hat{\theta}^*$ is in $I$, we have $\ell(y, x(\Delta)) =
\ell(y, x^0)$, where $x^0$ are the
previously defined data in absence of any style variance.  Hence
\begin{equation}\label{eq:tmp1}
\hat{\theta}^* = \mbox{argmin}_{\theta}\;\frac{1}{\ntot}\sum_{i=1}^\ntot
 \ell (y_i, f_\theta(x^0_{i}))   \mbox{   such that }
                 \theta \in I,
\end{equation}
that is the estimator is unchanged if we use the (hypothetical) data $x^0_i$,
$i=1,\ldots,n$ as training data.
The population optimal parameter vector defined in~\eqref{eq:star} as
\begin{equation}\label{eq:tmp2}
{\theta}^* = \mbox{argmin}_\theta \;   E_{F_0}\Big[  \ell( Y, f_\theta(X))  \Big]
\mbox{ such that } \theta \in I. \end{equation}
is for all $\xi\ge 0$ identical to
\[
\mbox{argmin}_\theta \;   \sup_{F\in \mathcal{F}_\xi} E_{F}\Big[  \ell( Y, f_\theta(X))  \Big]
\mbox{ such that } \theta \in I. \]
Hence~\eqref{eq:tmp1} and~\eqref{eq:tmp2} can be written as
\begin{align*}
\hat{\theta}^* &= \mbox{argmin}_{\theta:\theta\in I}\; L^{(0)}_\ntot
(\theta) \mbox{   with } L^{(0)}_\ntot
(\theta) := \frac 1 \ntot \sum_{i=1}^\ntot \ell(y_i, f_\theta(x^0_{i})) \\
{\theta}^* &= \mbox{argmin}_{\theta:\theta\in I}\; L^{(0)}
(\theta) \mbox{   with } L^{(0)}
(\theta) := E[  \ell(Y, f_\theta(X^0))].
\end{align*}
 By uniform convergence
 of $L_\ntot^{(0)}$ to the population loss $L^{(0)}$, we have
 $L^{(0)}(\hat{\theta}^*) \to_p L^{(0)}(\theta^*)$.
By definition of $I$ and $\theta^*$, we have $L_\infty^* = L_\infty(\theta^*)=L^{(0)}(\theta^*)$.  As
$\hat{\theta}^*$ is in $I$, we also have $L_\infty(\hat{\theta}^*) =
L^{(0)}(\hat{\theta}^*)$.
Since, from above, $L^{(0)}(\hat{\theta}^*) \to_p L^{(0)}(\theta^*)$, this also
implies $L_\infty(\hat{\theta}^*) \to_p L_\infty(\theta^*)=L_\infty^*$. Using the previously established result that
$\hat{\theta}^{core}= \hat{\theta}^*$ with probability at least~$p_n$ and $p_n
\to 1$ for $\ntot\to\infty$, this completes the proof.
\section{Proof of Theorem~\ref{th:2}}\label{sec:proof2}

Let $F_0$ be the training distribution of $(\I,Y,\style)$ and $F$ a
distribution for $(\I,Y,\tstyle)$ in
$\mathcal{F}_\xi$. By definition of $\mathcal{F}_\xi$, we can write
$\tstyle=\style+\Delta$ for a suitable random variable $\Delta\in \mathbb{R}^q$  with
\[ \Delta \in \mathcal{U}_\xi,\qquad \text{where   } \mathcal{U}_\xi=\{\Delta:
 E( E( \Delta^t \Sigma_{Y,\I}^{-1} \Delta |Y,\I)) \le \xi \}.\] Vice versa:
if we can write $\tstyle=\style+ \Delta$ with $\Delta\in \mathcal{U}_\xi$,  then the distribution is in $F_\xi$.
While $X$ under $F_0$ can be written as $X(\Delta=0)$, the distribution
of $X$ under $F$ is of the form $X(\Delta)$ or, alternatively, $X(\sqrt{\xi}
U)$ with $U\in \mathcal{U}_1$. Adopting from now on the latter
constraint that $U\in \mathcal{U}_1$, and using
(B2),
\[ E_F\Big[ \ell\big(Y,f_\theta(X) \Big]=  E_{F_0} \Big[ h_\theta( 0)
\Big] +  \sqrt{\xi}\; E_{F_0} \Big[  (\nabla h_\theta)^t U \Big] + o(\xi),\]
where $\nabla h_\theta$ is the gradient of $h_\theta(\delta)$ with
respect to $\delta$, evaluated at $\delta\equiv 0$.
Hence
\[ \sup_{F\in F_\xi} E_F\Big[ h_\theta( \Delta ) \Big] =  E_{F_0} \Big[ h_\theta( 0)
\Big] + \sqrt{\xi} \sup_{U\in \mathcal{U}_1} E_{F_0} \Big[  (\nabla
h_\theta)^t U \Big] + o(\xi).\]
The proof is complete if we can show that
\[ C_{\ell,1/2,\theta} = \sup_{U\in \mathcal{U}_1} E_{F_0} \Big[  (\nabla
h_\theta)^t U \Big] +O(\zeta).\]
On the one hand,
\[ \sup_{U\in \mathcal{U}_1} E_{F_0} \Big[  (\nabla h_\theta)^t U
\Big]  =  E_{F_0}  \Big[  \sqrt{ (\nabla h_\theta)^t \Sigma_{Y,\I} (\nabla
h_\theta)} \Big] .\]
This follows for a matrix $\Sigma$ with Cholesky decomposition
$\Sigma=C^t C$,
\begin{align*} \max_{u: u^t \Sigma^{-1} u\le 1} (\nabla h_\theta)^t u
  & = \max_{w:
  \|w\|_2^2\le 1} (\nabla h_\theta)^t C^t w \\ &= \|C (\nabla h)\|_2 =\sqrt{ (\nabla h)^t \Sigma (\nabla h) } .\end{align*}
On the other hand, the conditional-variance-of-loss can be expanded as
\[ C_{\ell,1/2,\theta} = E_{F_0} \big[ \sqrt{ \text{Var} (\ell(Y,f_\theta(X)    ) |Y,\I
)} \big] =E_{F_0}  \Big[ \sqrt{ (\nabla h_\theta)^t \Sigma_{Y,\I} (\nabla
h_\theta)} \Big] + O(\zeta)  ,\]
which completes the proof.

\input{experiments_supp}

\end{document}

%% file: definitions.tex
\newcommand{\nid}{m}
\newcommand{\ntot}{n}
\newcommand{\In}{I_n}

\newcommand{\core}{{\sc CoRe}\xspace}
\newcommand{\ncf}{c}

\newcommand{\phc}{\hat{P}^{\small\textit{core}}}
\newcommand{\php}{\hat{P}^{\small\textit{pool}}}
\newcommand{\corefeat}{X^\text{core}}
\newcommand{\corefeato}{\text{core}}

\newcommand{\orth}{X^{\text{style}}}
\newcommand{\orthsm}{x^{\text{style}}}
\newcommand{\torth}{\tilde{X}^{\text{style}}}
\newcommand{\style}{\orth}
\newcommand{\tstyle}{\torth}
\newcommand{\stylesm}{\orthsm}
\newcommand{\styleo}{\text{style}}

\newcommand{\domain}{D}
\newcommand{\meanexp}{\beta}

\newcommand{\I}{\mathrm{ID}}
\renewcommand{\i}{\mathrm{id}}
\newcommand{\Wmat}{{W}}

\newcommand{\propmales}{\kappa}

\newtheorem{assn}{Assumption}

%% file: experiments_supp.tex
\section{Network architectures}\label{subsec:architec}
We implemented the considered models in TensorFlow \citep{Abadi2015}. The model architectures used are detailed in Table~\ref{tab:architecture}. \core and the pooled estimator use the same network architecture and training procedure; merely the loss function differs by the  \core regularization term. In all experiments we use the Adam optimizer \citep{Kingma2015}. All experimental results are based on training the respective model five times (using the same data) to assess the variance due to the randomness in the training procedure.
In each epoch of the training, the training data $x_{i}, i = 1,
\ldots, \ntot$ are randomly shuffled, keeping the grouped observations
$(x_{i})_{i\in I_j}$ for $j\in \{1,\ldots,m\}$ together to ensure that mini batches will contain grouped observations. In all experiments the mini batch size is set to 120. For small $\ncf$ this implies that not all mini batches contain grouped observations, making the optimization more challenging.

\begin{table}
\begin{center}
\begin{small}
\begin{tabular}{llll}
\hline
\abovespace\belowspace
Dataset & Optimizer  & & Architecture \\
\hline
\abovespace
MNIST & Adam & Input & $28\times28\times1$\\
 & & CNN & Conv $5 \times 5 \times 16$,  $5 \times 5 \times 32$\\
 & & & (same padding, strides $=2$, ReLu activation),   \\
 \belowspace
 & & & fully connected, softmax layer \\

Stickmen & Adam & Input & $64\times64\times1$\\
 & & CNN & Conv $5 \times 5 \times 16$,  $5 \times 5 \times 32$,  $5 \times 5 \times 64$,  $5 \times 5 \times 128$\\
 & & & (same padding, strides $=2$, leaky ReLu activation),   \\
 \belowspace
 & & & fully connected, softmax layer \\

CelebA & Adam &  Input & $64\times48\times3$\\
(all experiments &  &  CNN & Conv $5 \times 5 \times 16$,  $5 \times 5 \times 32$,  $5 \times 5 \times 64$,  $5 \times 5 \times 128$\\
using CelebA) & & & (same padding, strides $=2$, leaky ReLu activation),   \\
\belowspace
 & & & fully connected, softmax layer \\

AwA2 & Adam & Input & $32\times32\times3$\\
 & &  CNN & Conv $5 \times 5 \times 16$,  $5 \times 5 \times 32$,  $5 \times 5 \times 64$,  $5 \times 5 \times 128$\\
 & & & (same padding, strides $=2$, leaky ReLu activation),   \\
 & & & fully connected, softmax layer \\
\hline
\end{tabular}
\end{small}
\end{center}
\caption{\small Details of the model architectures used.}\label{tab:architecture}
\end{table}

\section{Additional experiments}\label{supp:sec:add_exp}

\subsection{Eyeglasses detection with small sample size}
\begin{figure}
\centering
\subfloat[]{
     \includegraphics[width=.48\textwidth, keepaspectratio=true]{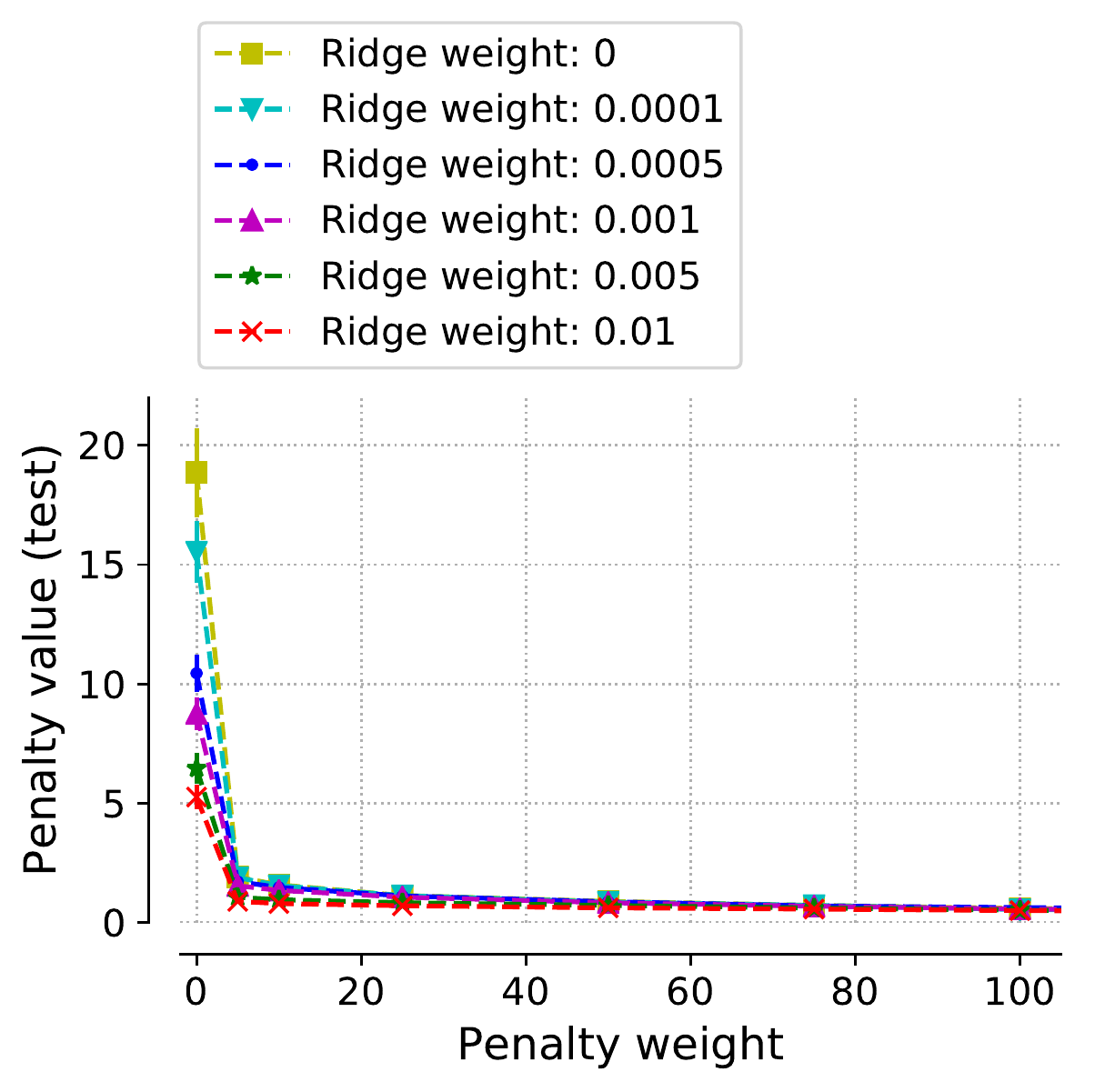}\label{fig:smalln_lambda_core_pen}
}
\subfloat[]{
     \includegraphics[width=.48\textwidth, keepaspectratio=true]{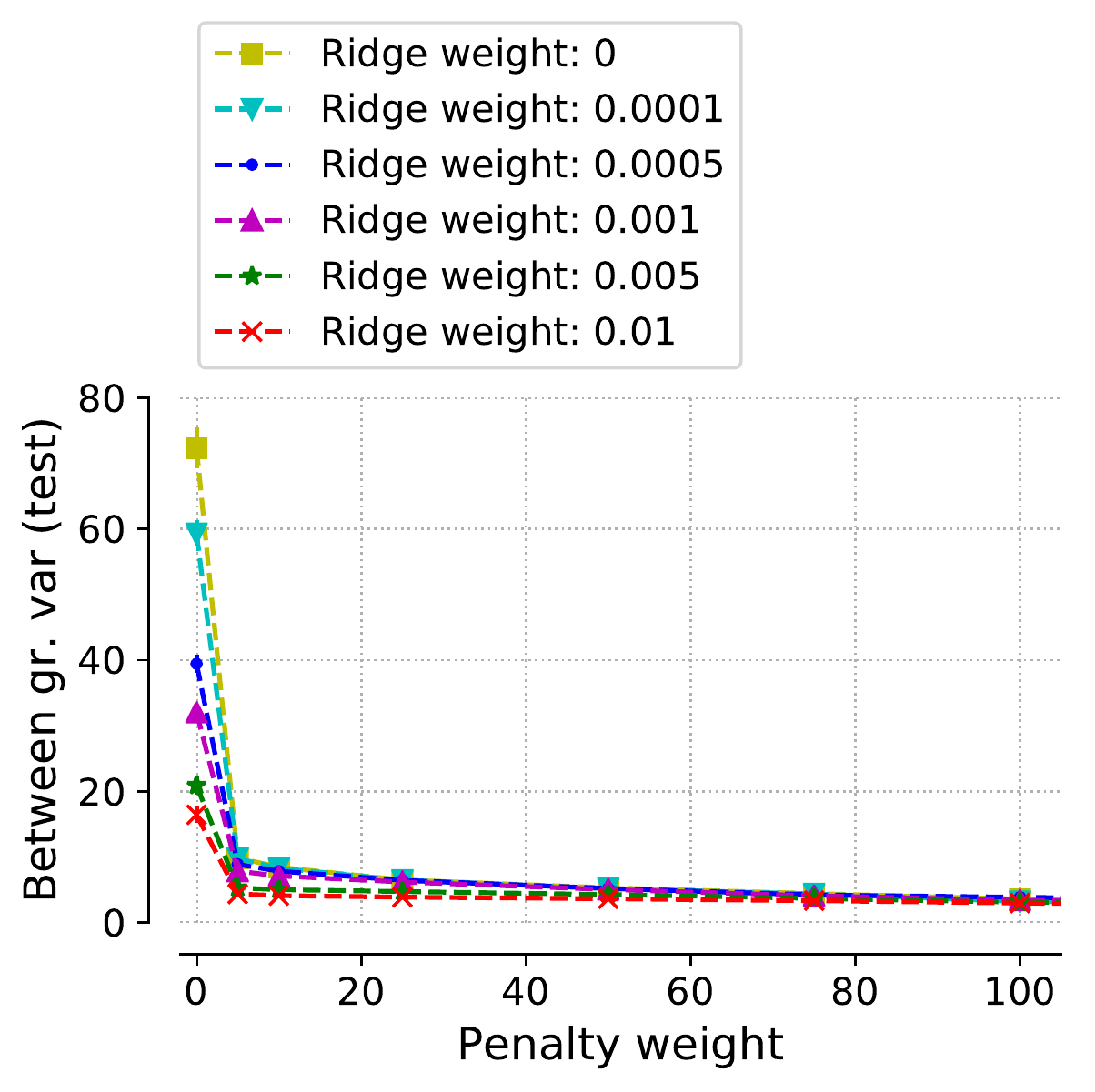}\label{fig:smalln_lambda_between_gr}
}
\captionof{figure}{{\small Eyeglass detection, trained on a small subset (DS1) of the CelebA dataset with disjoint identities. Panel~\protect\subref{fig:smalln_lambda_core_pen} shows the numerator of the variance ratio defined in Eq.~\eqref{eq:varratio} on test data as a function of both the \core  and ridge penalty weights. Panel~\protect\subref{fig:smalln_lambda_between_gr} shows the equivalent plot for the denominator. A ridge penalty decreases
  both the within- and between-group variance while the \core penalty penalizes the
  within-group variance selectively (the latter can be seen more clearly in Figure~\ref{fig:smalln_lambda}\protect\subref{fig:smalln_lambda_var_ratio}).}} \label{fig:smalln_lambda_supp}
\end{figure}

Figure~\ref{fig:smalln_lambda_supp} shows the numerator
  and the denominator of the variance ratio defined in Eq.~\eqref{eq:varratio} separately as a function of the
  \core penalty weight. In conjunction with Figure~\ref{fig:smalln_lambda}\subref{fig:smalln_lambda_var_ratio}, we observe that a ridge penalty decreases
  both the within- and between-group variance while the \core penalty penalizes the
  within-group variance selectively.

\begin{figure*}
\begin{minipage}[t]{0.5\hsize}
\centering

\subfloat[Examples of misclassified observations.]{
\parbox{.33\linewidth}{%
{\tiny
$y \equiv \textit{glasses}$ \\
$\phc(\textit{gl.}) = 1.00$ \\
$\php(\textit{gl.}) = 0.21$ \\  }

\centering
      \includegraphics[width=.75\linewidth, keepaspectratio=true, trim={10.79cm 0 .88cm 0}, clip]{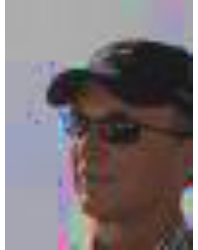} }
\parbox{.33\linewidth}{%
    {\tiny
$y \equiv \textit{no glasses}$ \\
$\phc(\textit{no gl.}) = 0.84$ \\
$\php(\textit{no gl.}) = 0.13$ \\}

\centering
       \includegraphics[width=.75\linewidth, keepaspectratio=true, trim={4.88cm 0 6.81cm 0}, clip]{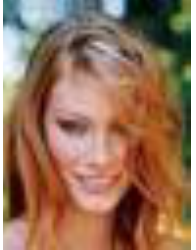} }
\parbox{.33\linewidth}{%
 { \tiny
$y \equiv \textit{glasses}$ \\
$\phc(\textit{gl.}) = 0.90$ \\
$\php(\textit{gl.}) = 0.14$ \\}

\centering
       \includegraphics[width=.75\linewidth, keepaspectratio=true, trim={10.79cm 0 .88cm 0}, clip]{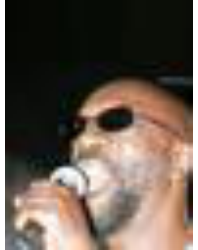} }
    }
\end{minipage}
\begin{minipage}[t]{0.5\hsize}
\subfloat[Misclassification rates.]{
\centering
\includegraphics[width=1\textwidth, keepaspectratio=true]{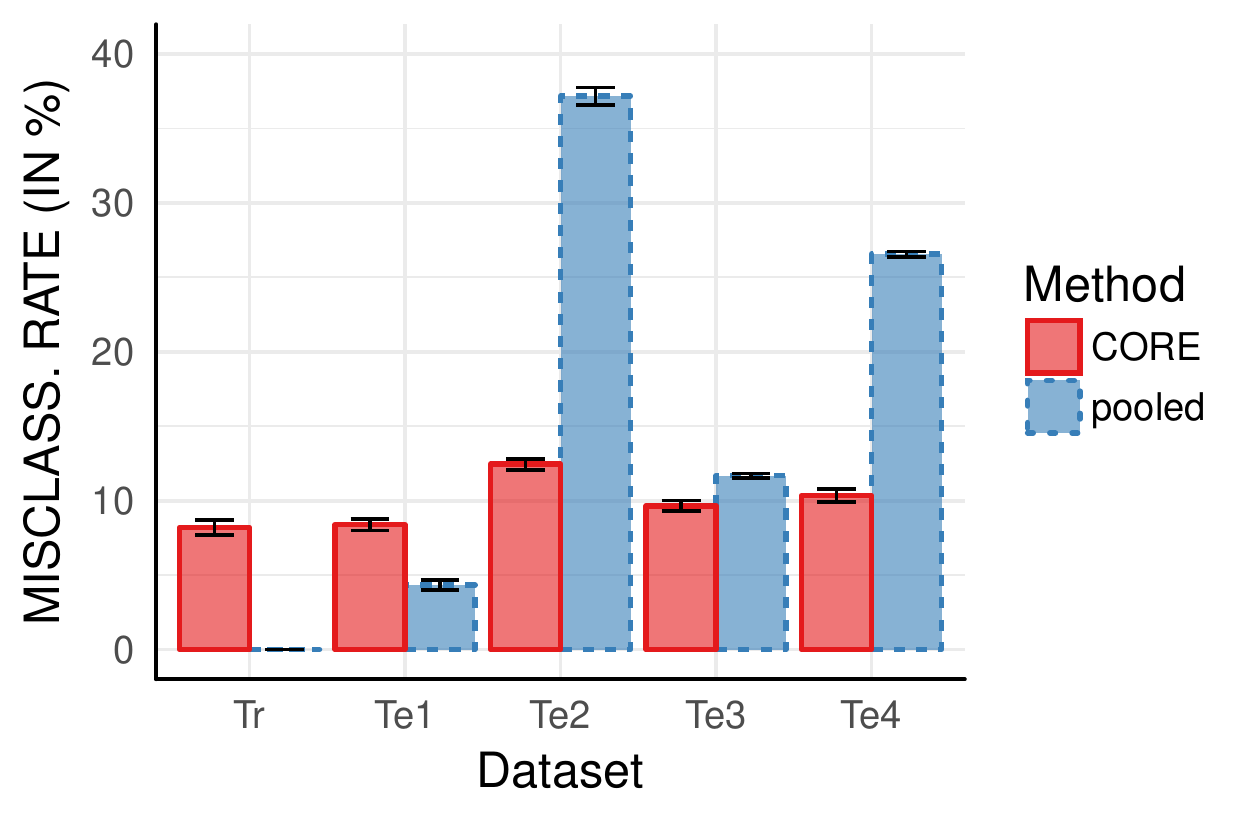}\label{fig:celeb_bri_res}}
\end{minipage}
\vspace{-.3cm}
\captionof{figure}{\small CelebA eyeglasses detection with brightness interventions, grouping setting~1. (a) Misclassified examples from the test
  sets. (b) Misclassification rates for $\meanexp =
  20$ and $\ncf = 2000$. Results for different test sets,
  grouping settings, $\meanexp \in \{ 5, 10, 20\}$ and $\ncf \in \{200, 5000\}$ can be found in Figure~\ref{fig:res_celeb1_supp}.}
\label{fig:data_set_celeb_bri}
\end{figure*}

\subsection{Eyeglasses detection: known and unknown brightness interventions}\label{subsec:celeb_brightness_supp}
Here, we show additional results for the experiment discussed in  \S\ref{subsec:celeb_brightness}. Recall that we work with the CelebA dataset and consider the problem of classifying whether the person in the image is wearing eyeglasses. 
We discuss two alternatives for constructing different test sets and we vary the number of grouped observations in
$\ncf\in\{200,2000,5000\}$ as well as the strength of the brightness
interventions in $\meanexp \in \{ 5, 10, 20\}$, all with sample size $\ntot=20000$.
Generation of training and test sets~1 and~2 were already described in
\S\ref{subsec:celeb_brightness}. Here, we consider additionally test set 3
where all images are left unchanged (no brightness interventions at
all) and in test set 4 the brightness of all images is increased.

\begin{figure*}[!t]
\centering
\subfloat[Grouping setting 1, $\meanexp=5$]{
     \includegraphics[width=.3\textwidth, keepaspectratio=true]{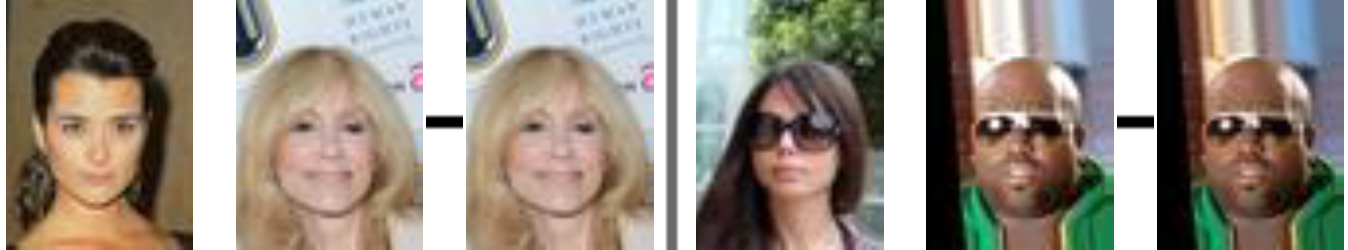}
}
\hspace{.2cm}
\subfloat[Grouping setting 1, $\meanexp=10$]{
     \includegraphics[width=.3\textwidth, keepaspectratio=true]{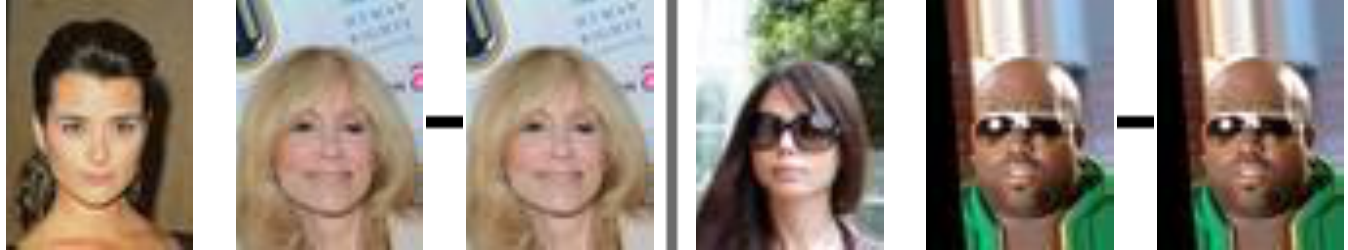}
}
\hspace{.2cm}
\subfloat[Grouping setting 1, $\meanexp=20$]{
     \includegraphics[width=.3\textwidth, keepaspectratio=true]{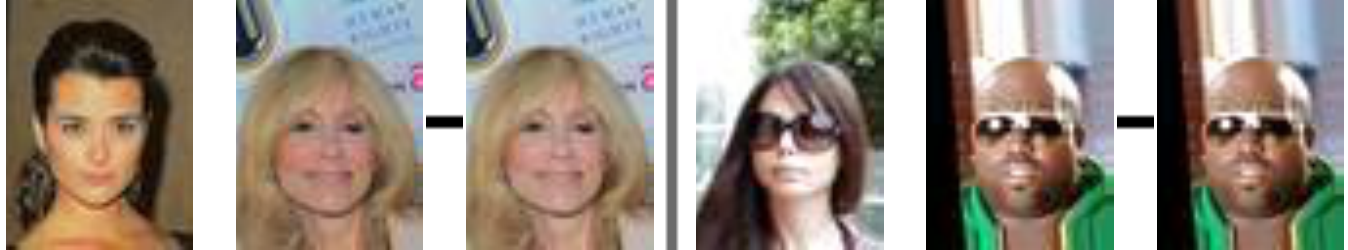}
}

\vspace{-.1cm}
\subfloat{
     \includegraphics[width=.3\textwidth, keepaspectratio=true]{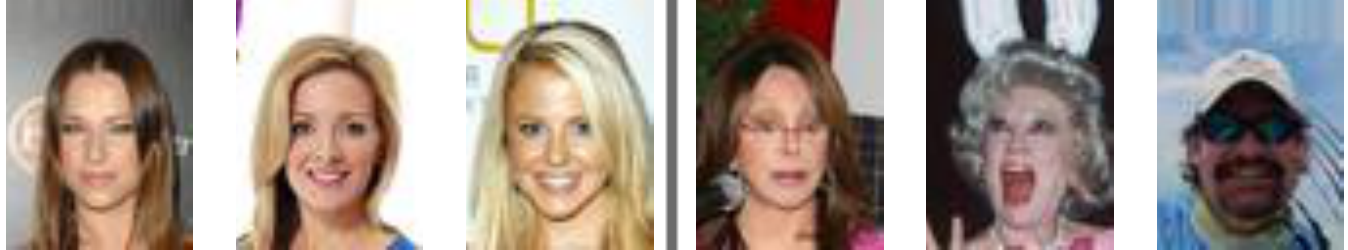}
}
\hspace{.2cm}
\subfloat{
     \includegraphics[width=.3\textwidth, keepaspectratio=true]{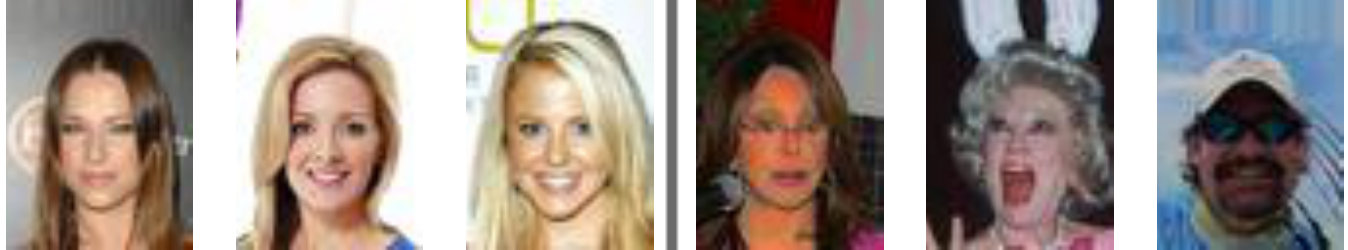}
}
\hspace{.2cm}
\subfloat{
     \includegraphics[width=.3\textwidth, keepaspectratio=true]{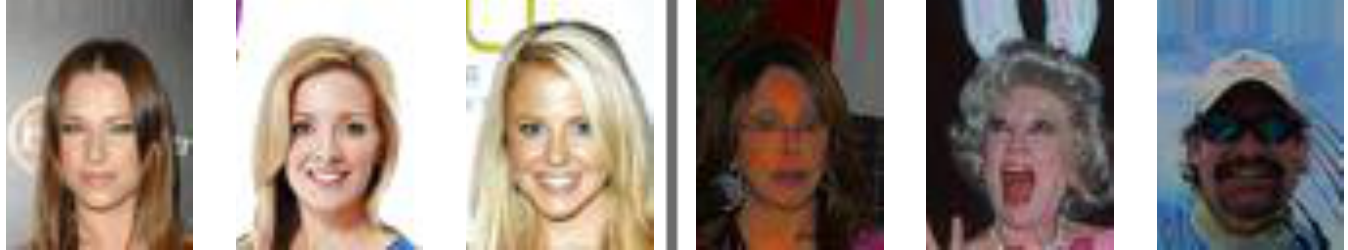}
}

\vspace{-.1cm}
\subfloat{
     \includegraphics[width=.3\textwidth, keepaspectratio=true]{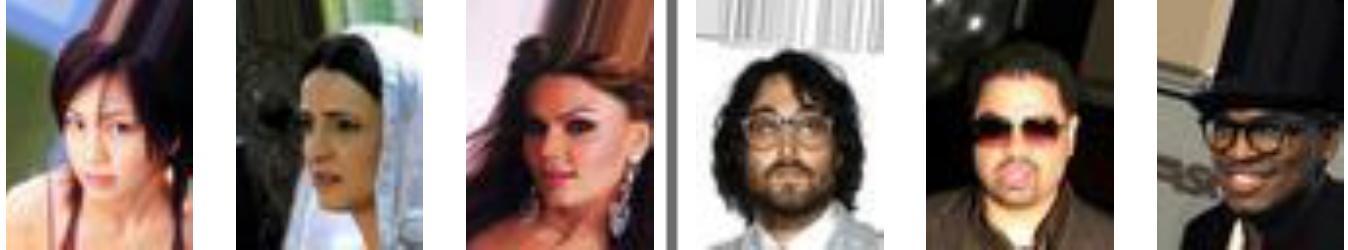}
}
\hspace{.2cm}
\subfloat{
     \includegraphics[width=.3\textwidth, keepaspectratio=true]{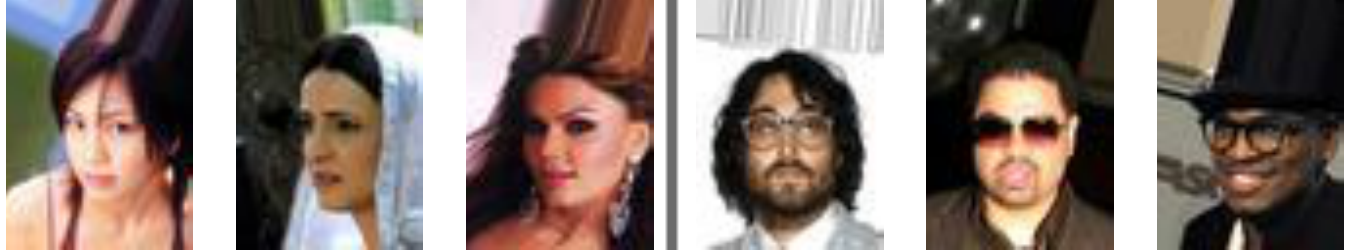}
}
\hspace{.2cm}
\subfloat{
     \includegraphics[width=.3\textwidth, keepaspectratio=true]{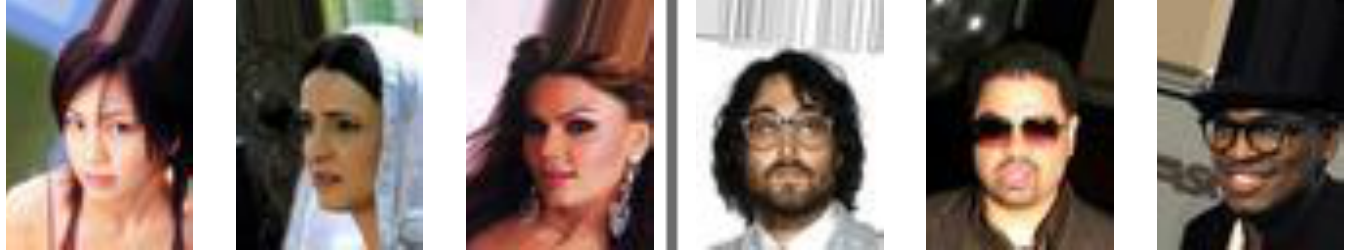}
}

\vspace{-.1cm}
\subfloat{
     \includegraphics[width=.3\textwidth, keepaspectratio=true]{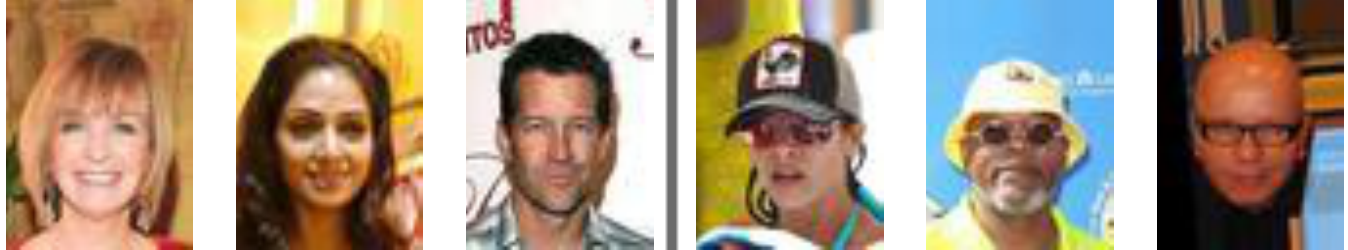}
}
\hspace{.2cm}
\subfloat{
     \includegraphics[width=.3\textwidth, keepaspectratio=true]{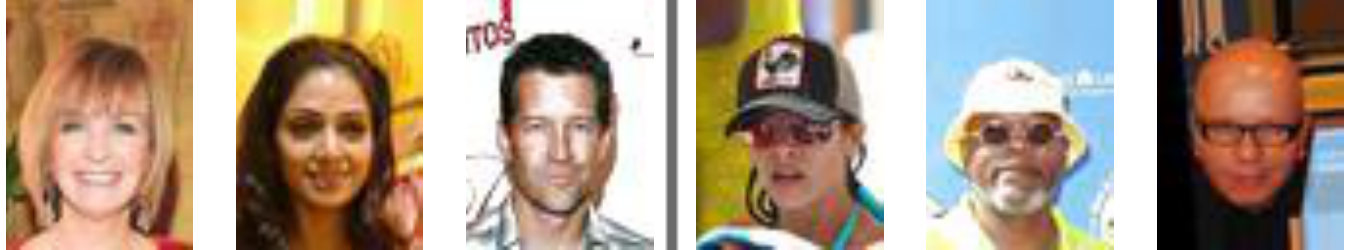}
}
\hspace{.2cm}
\subfloat{
     \includegraphics[width=.3\textwidth, keepaspectratio=true]{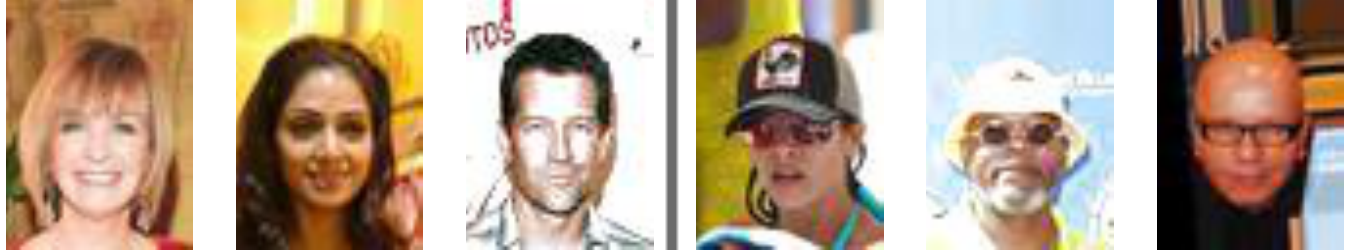}
}

\addtocounter{subfigure}{-9}
\subfloat[Grouping setting 2, $\meanexp=5$]{
     \includegraphics[width=.3\textwidth, keepaspectratio=true]{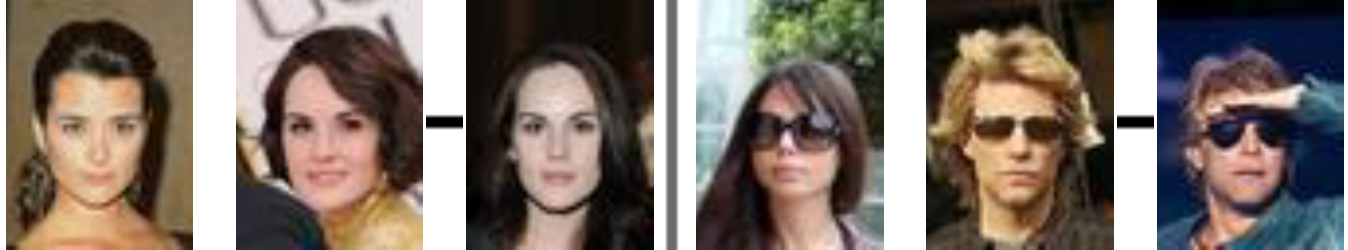}
}
\hspace{.2cm}
\subfloat[Grouping setting 2, $\meanexp=10$]{
     \includegraphics[width=.3\textwidth, keepaspectratio=true]{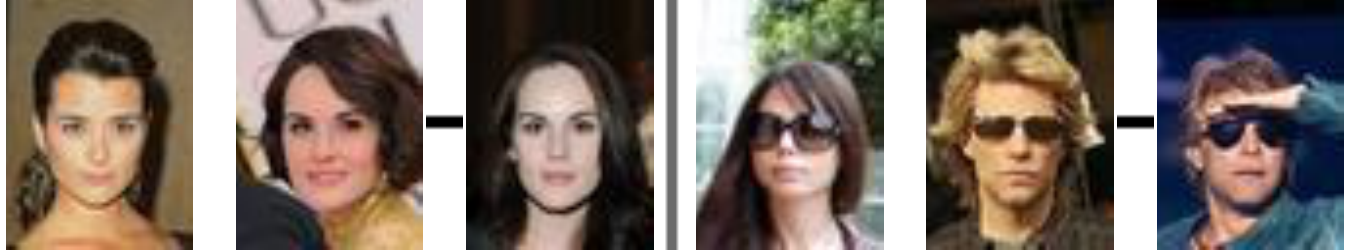}
}
\hspace{.2cm}
\subfloat[Grouping setting 2, $\meanexp=20$]{
     \includegraphics[width=.3\textwidth, keepaspectratio=true]{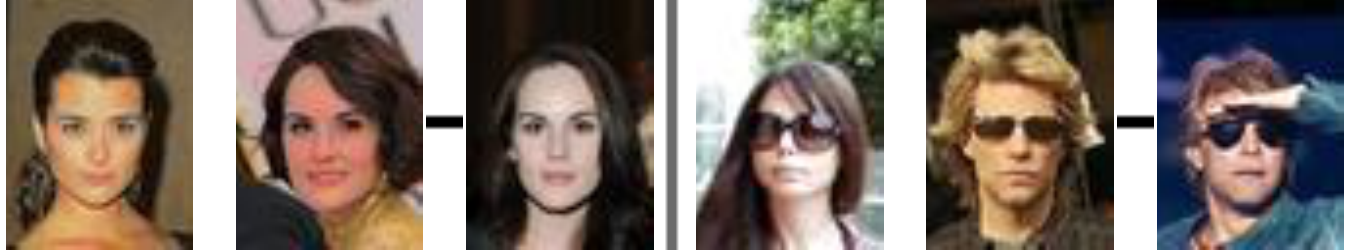}
}

\subfloat[Grouping setting 3, $\meanexp=5$]{
     \includegraphics[width=.3\textwidth, keepaspectratio=true]{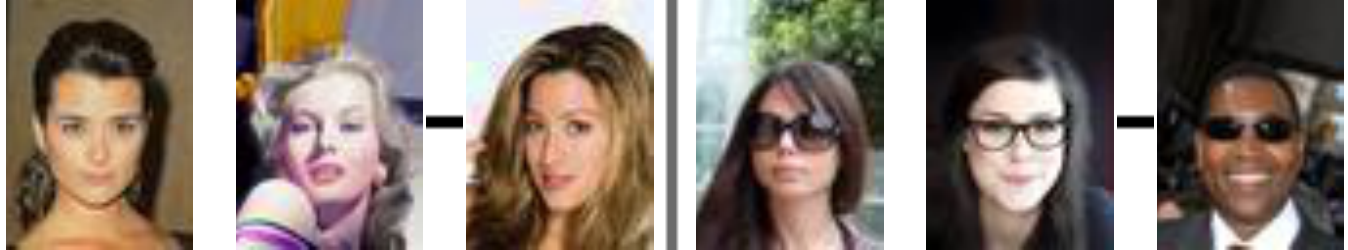}
}
\hspace{.2cm}
\subfloat[Grouping setting 3, $\meanexp=10$]{
     \includegraphics[width=.3\textwidth, keepaspectratio=true]{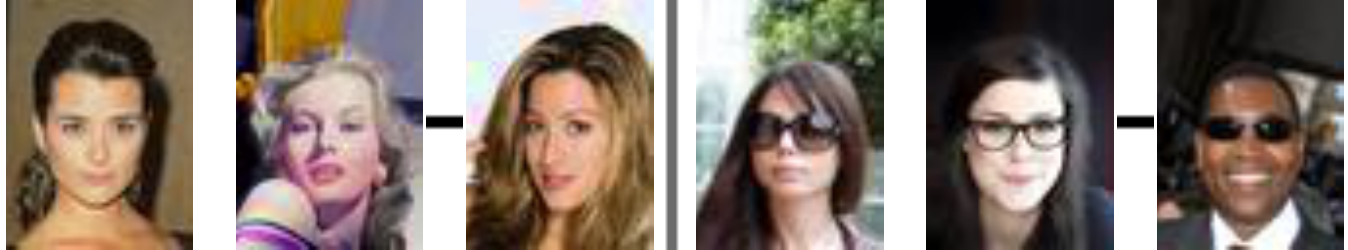}
}
\hspace{.2cm}
\subfloat[Grouping setting 3, $\meanexp=20$]{
     \includegraphics[width=.3\textwidth, keepaspectratio=true]{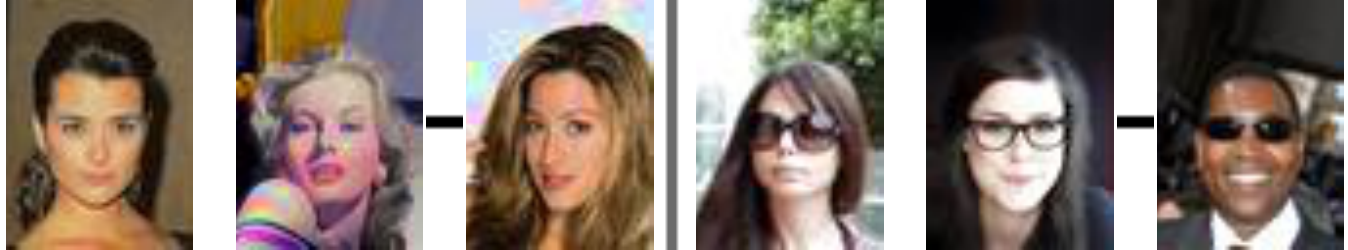}
}
\captionof{figure}{\small Examples from the CelebA eyeglasses detection with brightness interventions, grouping settings 1--3 with $\meanexp \in \{5, 10, 20 \}$. In all rows, the first three images from the left have $y\equiv\textit{no glasses}$; the remaining three images have $y\equiv\textit{glasses}$. Connected images are grouped examples. In panels (a)--(c), row 1 shows examples from the training set, rows 2--4 contain examples from test sets 2--4, respectively. Panels (d)--(i) show examples from the respective training sets.}\label{fig:data_set_celeb_bri_supp}
\end{figure*}

Furthermore, we consider three different ways of grouping images. 
In \S\ref{subsec:celeb_brightness} we used images of the same person  to
create a  grouped observation by sampling a different value for the
brightness intervention. We refer to this as `Grouping setting 2'
here. An alternative is to use the same image of the same person in
different brightnesses (drawn from the same distribution) as a group over which the conditional variance
is calculated. We call this `Grouping setting 1' and it can be useful if we
know that we want to protect against brightness interventions in the
future. For comparison, we also evaluate grouping with an image of a different
person (but sharing the same class label) as a baseline (`Grouping setting 3'). Examples from
the training sets using grouping settings 1, 2 and 3 can be found in Figure~\ref{fig:data_set_celeb_bri_supp}.

\begin{figure}
\centering
\subfloat[Grouping setting 1, $\ncf = 200$]{
     \includegraphics[width=.48\textwidth, keepaspectratio=true]{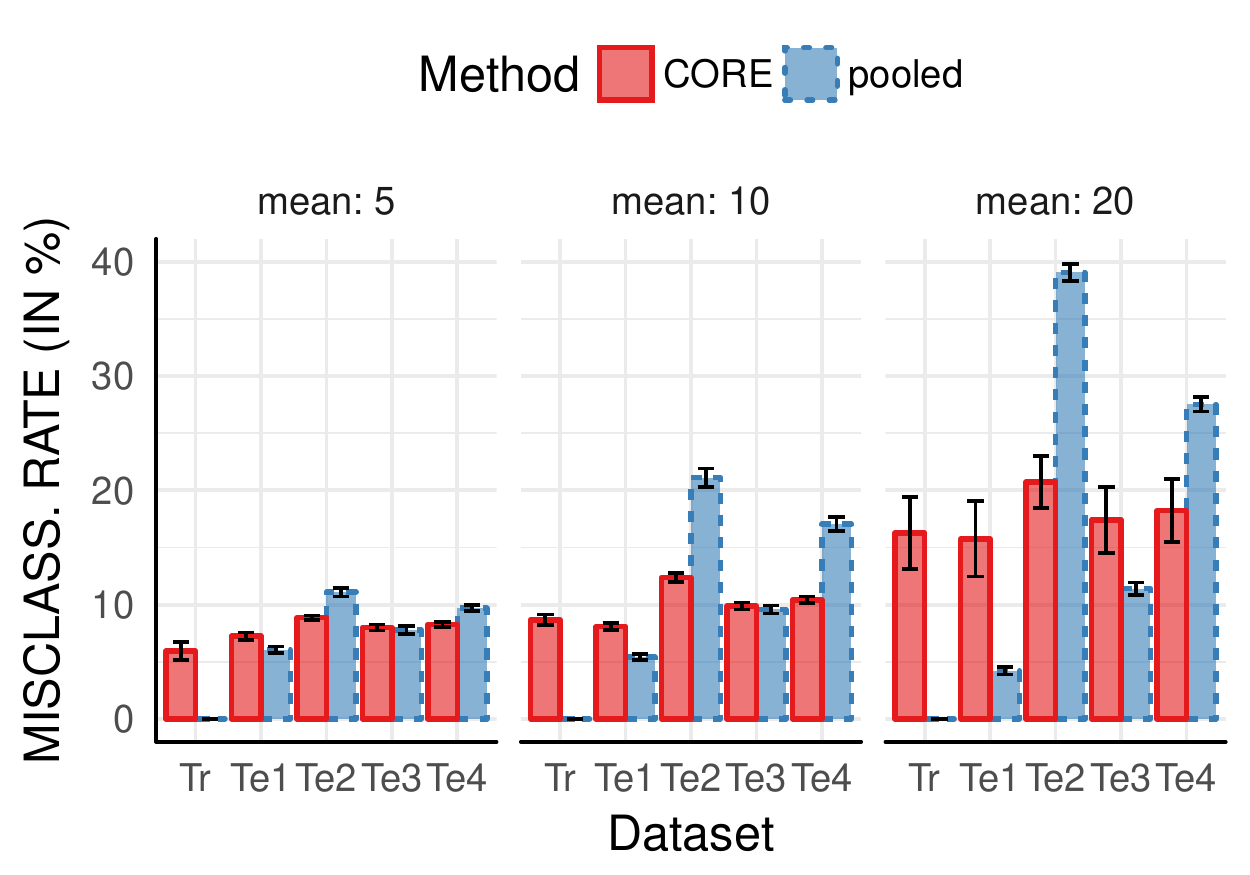}
}
\subfloat[Grouping setting 1, $\ncf = 2000$]{
     \includegraphics[width=.48\textwidth, keepaspectratio=true]{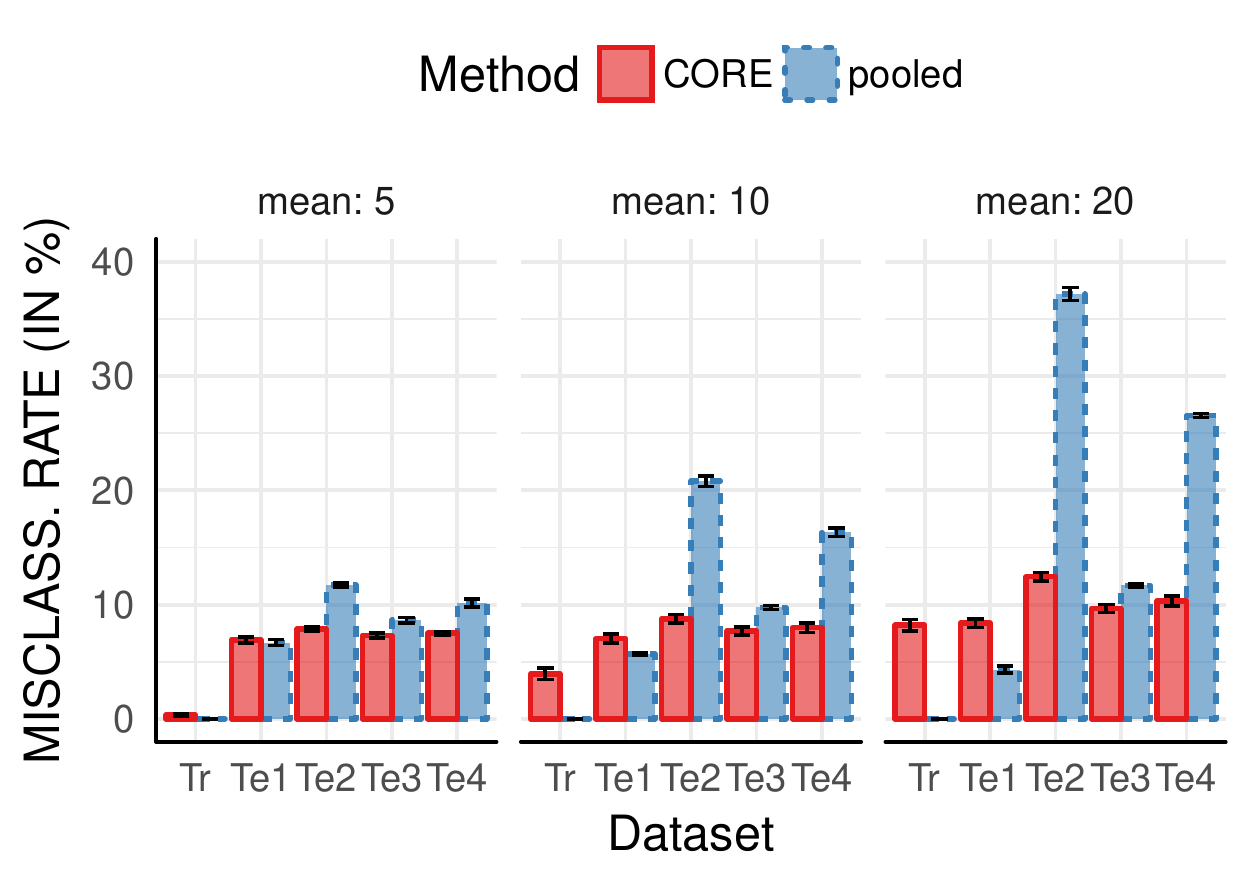}
}

\subfloat[Grouping setting 2, $\ncf = 2000$]{
     \includegraphics[width=.48\textwidth, keepaspectratio=true, trim={0 0 0 1.3cm}, clip]{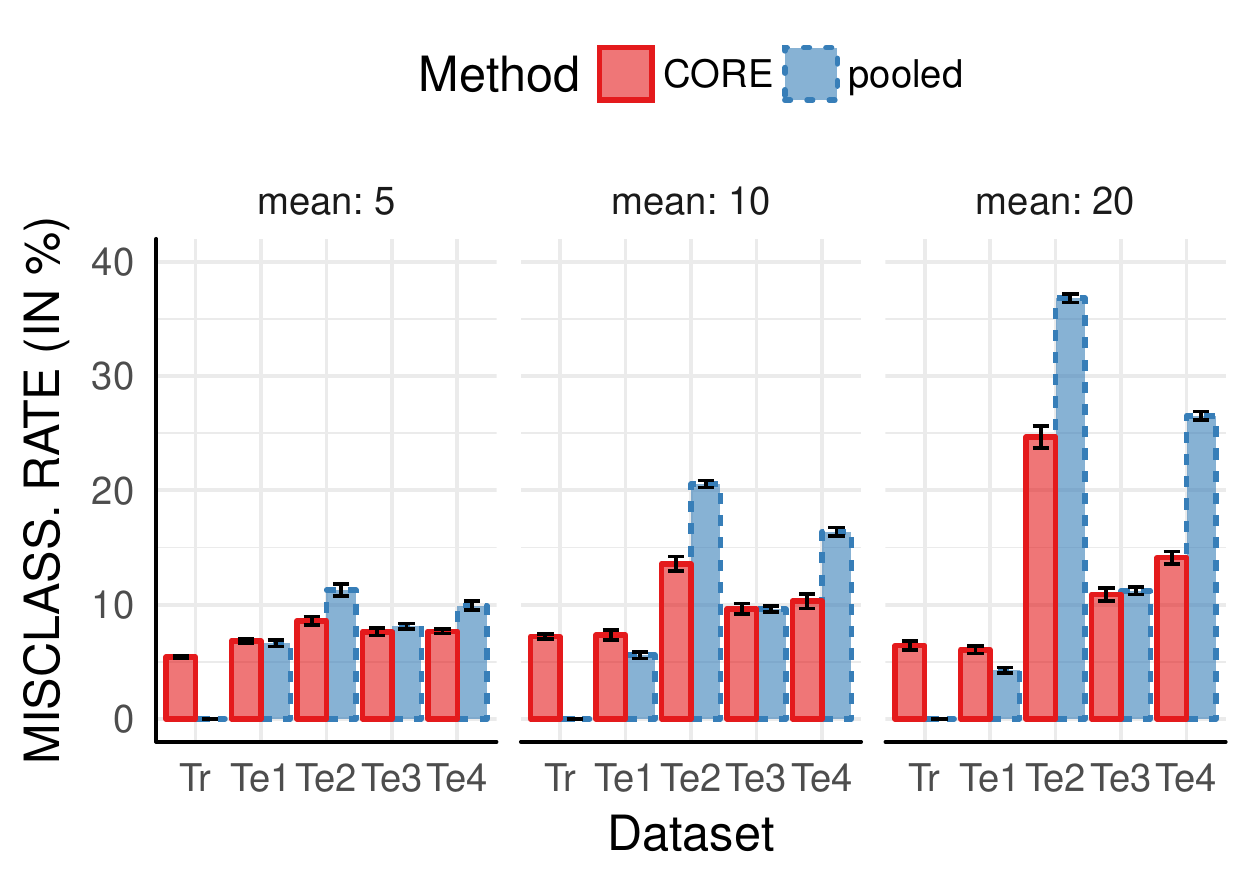}
}
\subfloat[Grouping setting 2, $\ncf = 5000$]{
     \includegraphics[width=.48\textwidth, keepaspectratio=true, trim={0 0 0 1.3cm}, clip]{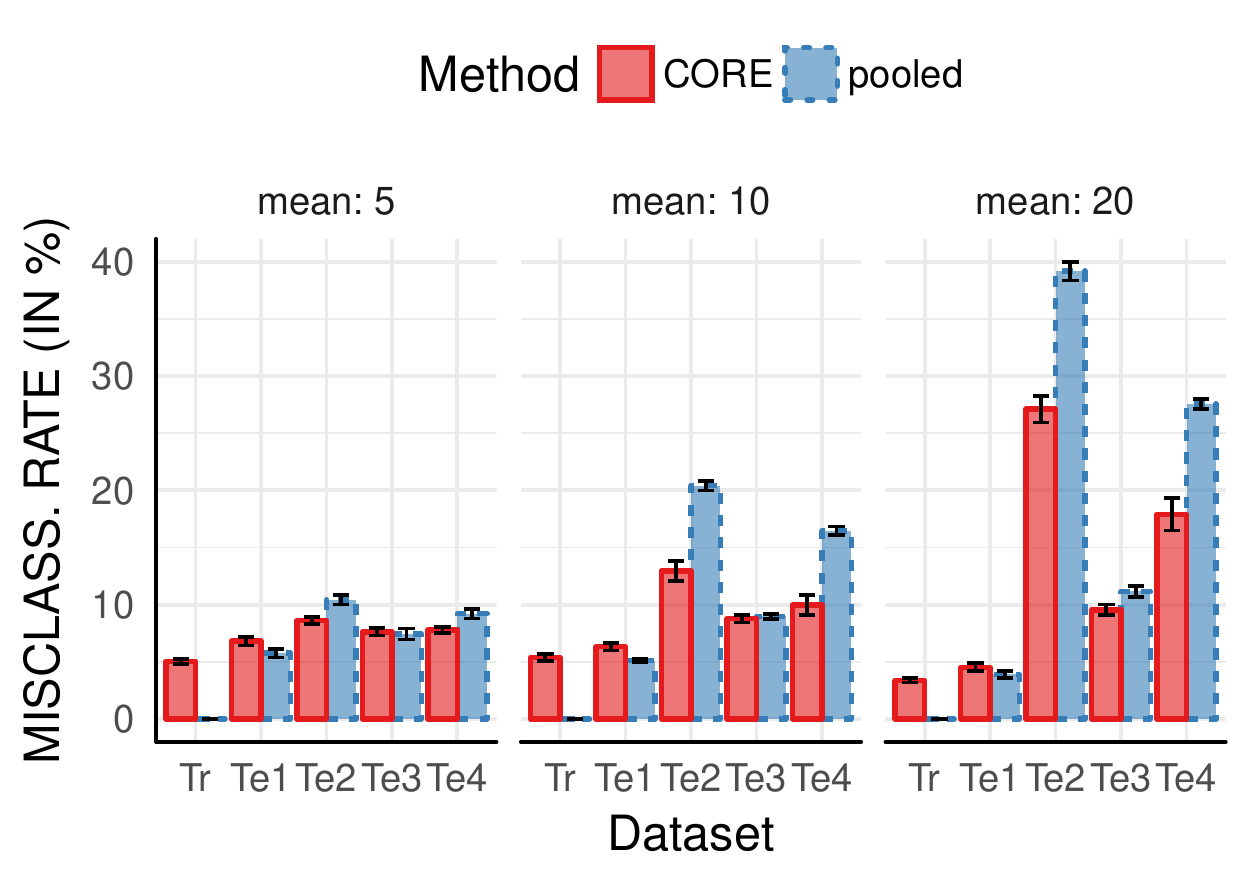}
}

\subfloat[Grouping setting 3, $\ncf = 2000$]{
     \includegraphics[width=.48\textwidth, keepaspectratio=true, trim={0 0 0 1.3cm}, clip]{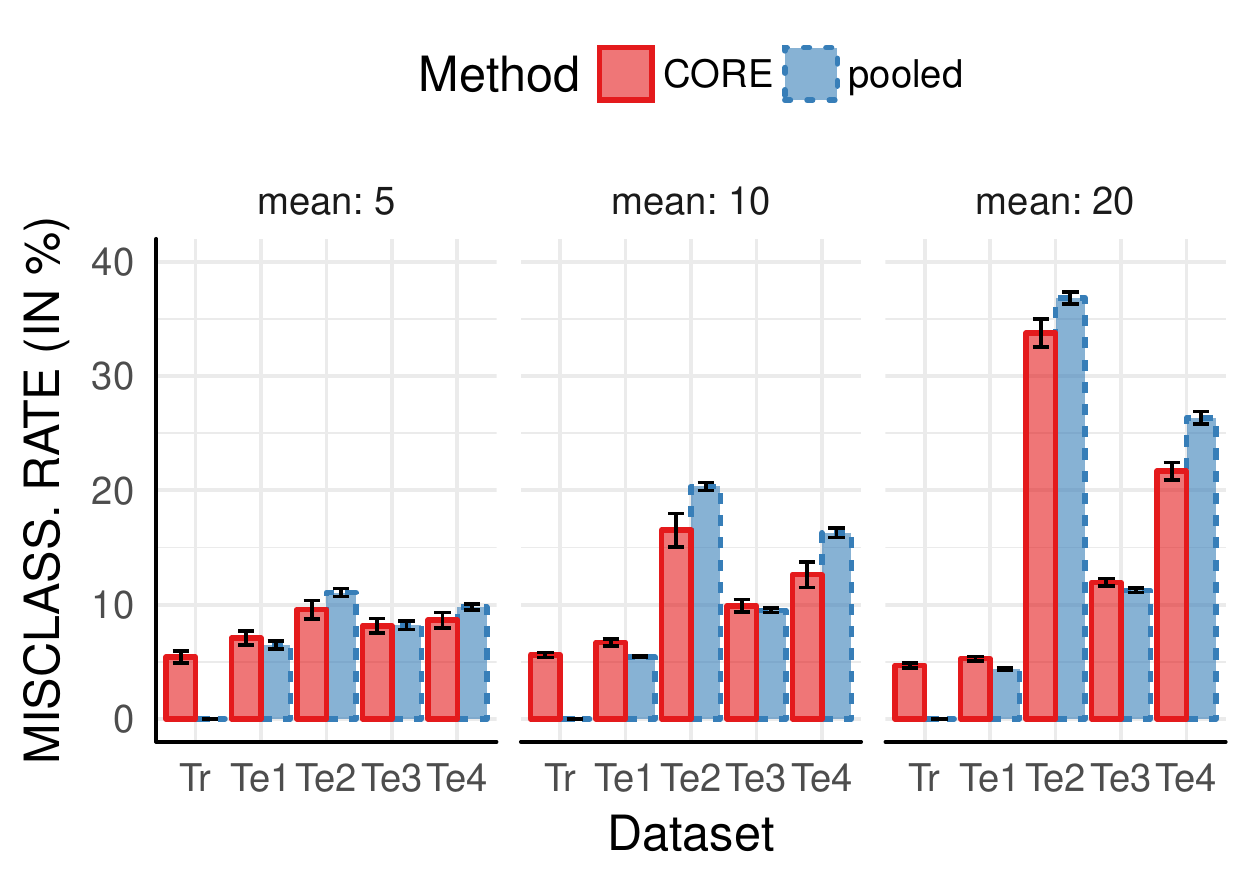}
}
\subfloat[Grouping setting 3, $\ncf = 5000$]{
     \includegraphics[width=.48\textwidth, keepaspectratio=true, trim={0 0 0 1.3cm}, clip]{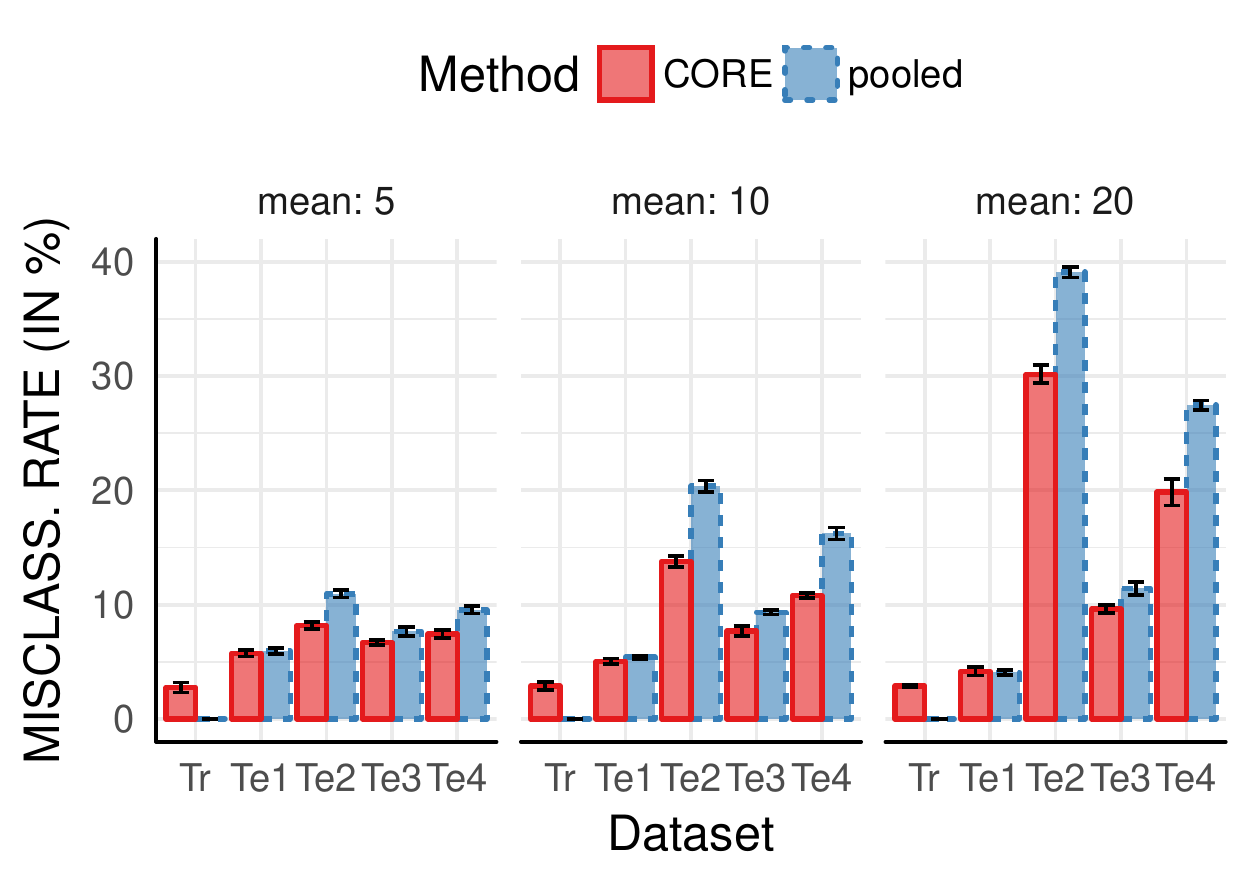}
}
\captionof{figure}{{\small Misclassification rates for the CelebA eyeglasses detection with brightness interventions, grouping settings 1--3 with $\ncf \in \{200, 2000, 5000\}$ and the mean of the exponential distribution $\meanexp \in \{5, 10, 20 \}$.}}\label{fig:res_celeb1_supp}
\end{figure}

Results for all grouping settings, $\meanexp \in \{ 5, 10, 20\}$
and $\ncf \in \{200, 5000\}$ can be found in
Figure~\ref{fig:res_celeb1_supp}. We see that using grouping setting 1
works best since we could explicitly control that only
$\style\equiv\textit{brightness}$ varies between grouping examples. In
grouping setting 2, different images of the same person can vary in
many factors, making it more challenging to isolate brightness as the
factor to be invariant against.  Lastly, we see that if we group images
of different persons (`Grouping setting 3'), the difference between
\core estimator and the pooled estimator  becomes much smaller than in
the previous settings.

Regarding the results for grouping setting~1 in Figure~\ref{fig:data_set_celeb_bri}, we notice that the pooled estimator performs better than \core on test set 1. This can be explained by the fact that it can exploit the predictive information contained in the brightness of an image while \core is restricted not to do so.
Second, we observe that the pooled estimator does not perform well on test sets 2 and 4 as its learned representation seems to use the image's brightness as a predictor for the response which fails when the brightness distribution in the test set differs significantly from the training set. In contrast, the predictive performance of \core is hardly affected by the changing brightness distributions.

\subsection{Gender classification}\label{subsec:celeb_gender_supp} 
Table~\ref{tab:gender_males_prop_supp} additionally reports the standard errors for the results discussed in \S\ref{subsec:gender}.

{\small
\begin{sidewaystable}
\begin{center}
{\renewcommand{\arraystretch}{1.4}
\begin{tabular}{cl|ccc|ccc}
& & \multicolumn{3}{c|}{Error} &  \multicolumn{3}{c}{Penalty value} \\
& Method & Train & Test 1 & Test 2 & Train & Test: Females & Test: Males \\ \hline
\multirow{2}{*}{\rotatebox[origin=c]{90}{$\propmales = .5$}} 
& 5-layer CNN & 0.00\% (0.00\%)  &  2.00\% (0.17\%) & 38.54\% (0.81\%) & 22.77 (0.27) & 74.05 (2.17) & 30.67 (0.88)\\
& + \core & 6.43\% (0.25\%) & 5.85\% (0.17\%) & 24.07\% (0.80\%) & 0.01 (0.01) & 1.61 (0.14) & 0.93 (0.09)\\ \hline 
\multirow{2}{*}{\rotatebox[origin=c]{90}{$\propmales = .75$}} 
& 5-layer CNN & 0.00\% (0.00\%) & 1.98\% (0.09\%) & 43.41\% (0.44\%) & 8.23 (0.33) & 32.98 (1.44) & 11.76 (0.59) \\
& + \core & 7.61\% (0.44\%) & 6.99\% (0.42\%) & 27.05\% (1.32\%) & 0.00 (0.00) & 1.44 (0.20) & 0.62 (0.03) \\ \hline 
\multirow{2}{*}{\rotatebox[origin=c]{90}{$\propmales = .9$}} 
& 5-layer CNN & 0.00\% (0.00\%) & 2.00\% (0.11\%) & 47.64\% (1.11\%) & 9.47 (0.75) & 40.51 (1.62) & 14.37 (1.12) \\
& + \core & 8.76\% (0.59\%) & 7.74\% (0.68\%) & 30.63\% (1.73\%) & 0.00 (0.01) & 1.26 (0.31) & 0.42 (0.08) \\ \hline 
\multirow{2}{*}{\rotatebox[origin=c]{90}{$\propmales = .95$}} 
& 5-layer CNN & 0.00\% (0.00\%) & 1.89\% (0.16\%) & 48.96\% (1.02\%) & 13.62 (1.64) & 61.01 (1.22) & 21.26 (1.66)\\
&  + \core & 10.45\% (1.42\%) & 9.35\% (1.79\%) & 29.57\% (2.43\%) & 0.00 (0.00) & 0.42 (0.30) & 0.16 (0.11)\\ \hline 
\multirow{2}{*}{\rotatebox[origin=c]{90}{$\propmales = .99$}} 
& 5-layer CNN & 0.00\% (0.00\%) & 1.70\% (0.10\%) & 50.11\% (0.65\%) & 20.66 (1.25) & 70.80 (1.81) & 27.80 (1.12)\\
& + \core & 11.10\% (1.17\%) & 10.51\% (1.26\%) & 32.91\% (0.81\%) &  0.00 (0.00) &  0.00 (0.00) &  0.00 (0.00) \\ \hline 
\multirow{2}{*}{\rotatebox[origin=c]{90}{$\propmales = 1$}} 
& 5-layer CNN &  0.00\% (0.00\%)  & 1.93\% (0.05\%) & 49.41\% (1.50\%) & 821.32 (144.68) & 2524.77 (219.96) & 1253.21 (184.84) \\
&  + \core & 11.12\% (0.34\%) & 10.11\% (0.32\%) & 35.68\% (2.03\%) & 0.00 (0.00) & 0.02 (0.02) & 0.01 (0.01)  \\ \hline 
\end{tabular}
}
\end{center}
\caption{{\small  Classification for 
  $Y \in \{\text{woman},\text{man}\}$. We compare six different datasets that vary with respect to the
distribution of $Y$ {\it in the grouped observations}. Specifically, we vary the proportion of images showing men between $\propmales=0.5$ and $\propmales=1$. In all training 
datasets, the total number of observations is 16982 and the total number of grouped observations is 500.  Both the pooled
estimator as well as the \core estimator perform better if the
distribution of $Y$ in the grouped observations is more balanced. The \core estimator improves the
error rate of the pooled estimator by $\approx 28-39\%$ on a relative scale. }}\label{tab:gender_males_prop_supp}
\end{sidewaystable}}

\newpage
\subsection{MNIST: more sample efficient data
  augmentation }\label{subsec:mnist_supp}
\begin{figure}
\centering
\subfloat[$\nid = 1000$]{
     \includegraphics[width=.4\textwidth, keepaspectratio=true]{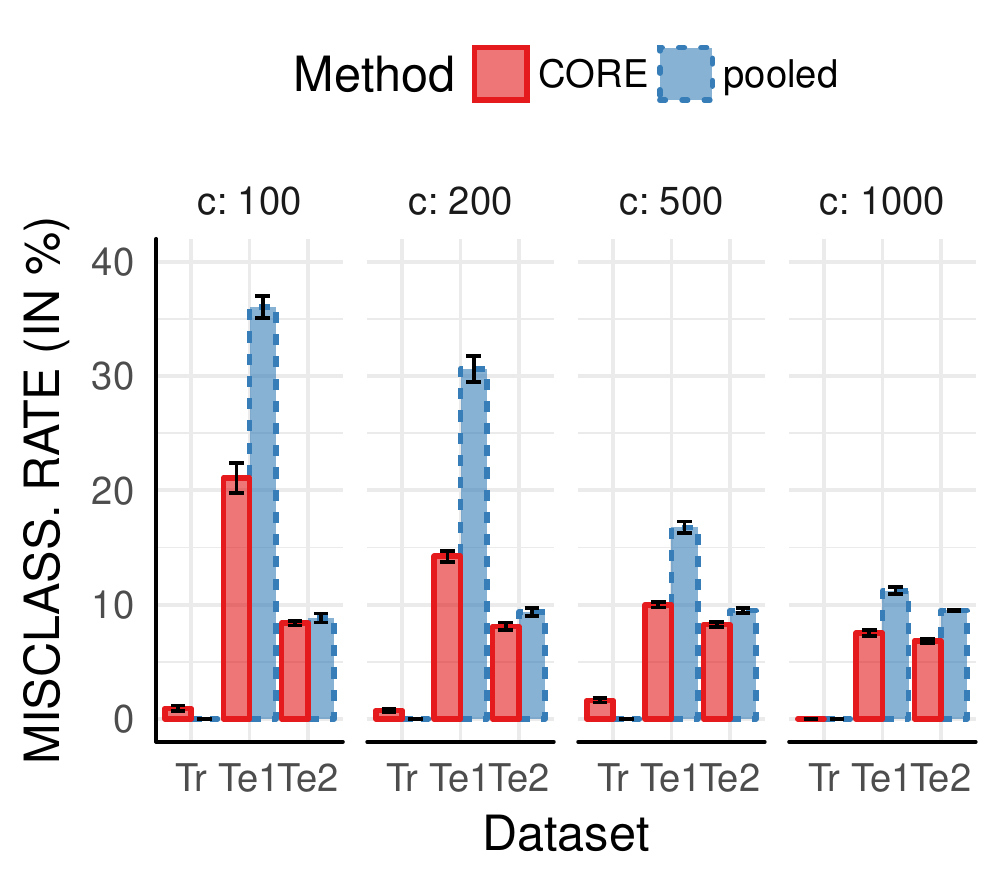}
}
\subfloat[$\nid = 10000$]{
     \includegraphics[width=.6\textwidth, keepaspectratio=true]{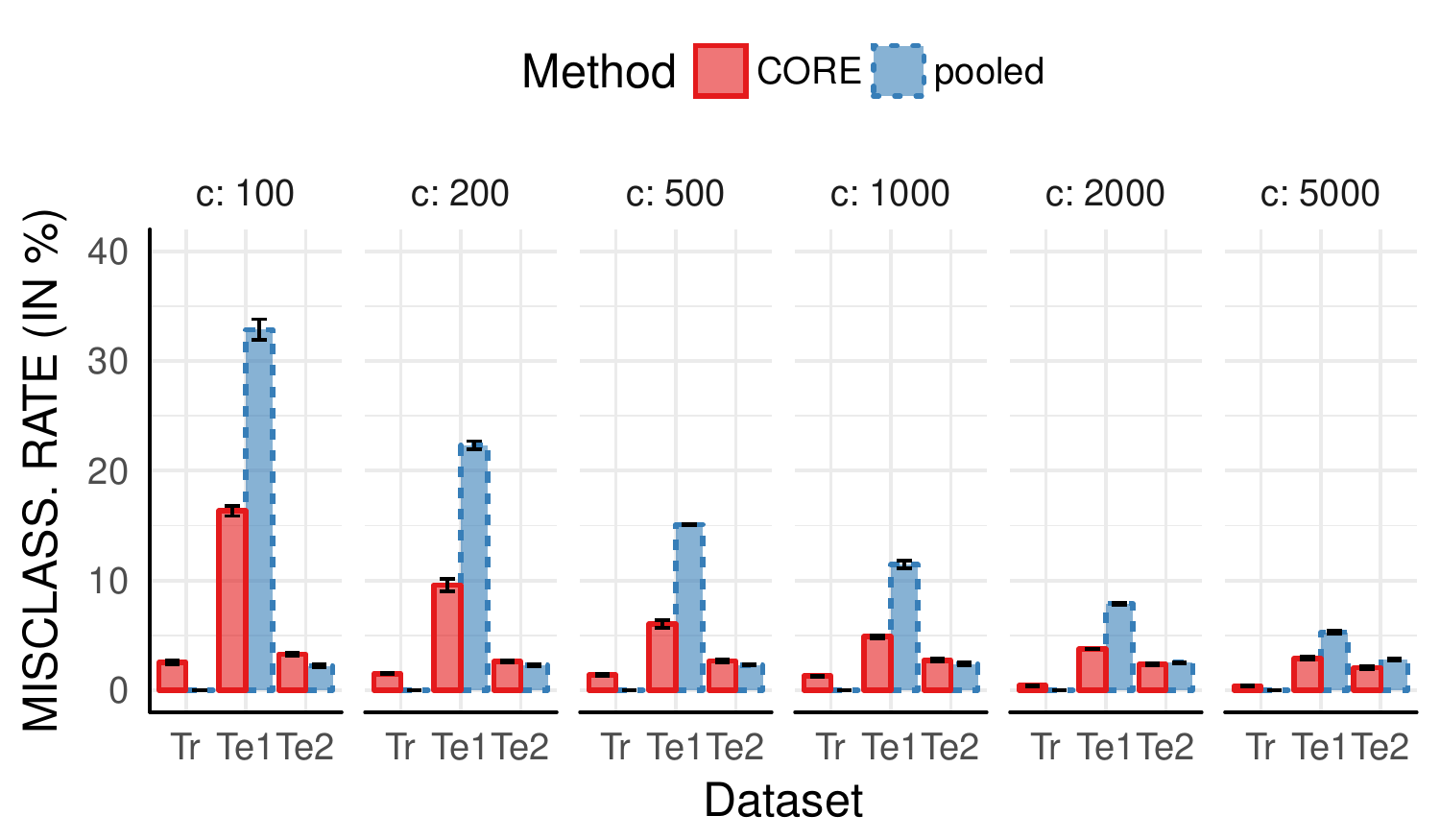}
}
\vspace{-.3cm}
\captionof{figure}{{\small Data augmentation setting: Misclassification rates for MNIST and $\orth \equiv \textit{rotation}$. In test set 1 all digits are rotated by a degree randomly sampled from $[35, 70]$. Test set 2 is the usual MNIST test set.}}\label{fig:res_mnist_da_supp}
\end{figure}

Here, we show further results for the experiment introduced in \S\ref{subsec:data_aug}. We vary the number of
augmented training examples $\ncf$ from $100$ to $5000$ for $\nid=10000$ and $\ncf \in \{100, 200, 500, 1000\}$ for $\nid=1000$. 
  The degree of the rotations is sampled uniformly at random from $[35,70]$. Figure~\ref{fig:res_mnist_da_supp} shows the misclassification rates. Test set 1 contains rotated digits only, test set 2 is the usual MNIST test set. We see that the misclassification rates of \core are always lower on test set 1, showing that it makes data augmentation more efficient. For $\nid=1000$, it even turns out to be beneficial for performance on test set 2. 

\subsection{Stickmen image-based age classification}
\begin{figure}
\begin{center} 
\begin{small}
\begin{tikzpicture}[>=stealth',shorten >=1pt,auto,node distance=3.2cm,
                    semithick]
  \tikzstyle{every state}=[fill=none,shape=rectangle,rounded corners=2mm,text=black]

  \draw (0,4.5) node[state] (H) {place of observation
    $\domain$};
  \draw (0,3) node[state, fill=pastelgray] (Y) {adult/child $Y$};
  \draw (1.5,1.5) node[state, dashed] (S) {movement $\orth(\Delta)$ };
  \draw (2.5,3) node[state,dashed] (D) {$\Delta$};
  \draw (-3,3) node[state, fill=pastelgray] (I) { person $\I$};
  \draw (-1.5,1.5) node[state] (W) {height $\corefeat$};
  \draw (0,0) node[state, fill=pastelgray, dashed] (X) {image $X(\Delta)$};
  \draw (2.5,0) node[state, fill=pastelgray, dashed] (Yhat) {$\hat{Y}(X(\Delta))$};

 \draw [-arcsq] (H) -- (Y);
 \draw [-arcsq]  (Y) -- (I);
 \draw [-arcsq]  (I) -- (W);
 \draw [-arcsq] (H) -- (D);
 \draw [-arcsq] (Y) -- (W);
 \draw [-arcsq] (Y) -- (S);
 \draw [-arcsq] (D) -- (S);
 \draw [-arcsq] (W) -- (X);
 \draw [-arcsq] (S) -- (X);
 \draw [-arcsq, dashed] (X) -- node {$f_\theta$} (Yhat);
\end{tikzpicture}
\captionof{figure}{\small Data generating process for the stickmen example.}
\label{fig:DAG_stick}
\end{small}
\end{center}
\end{figure}
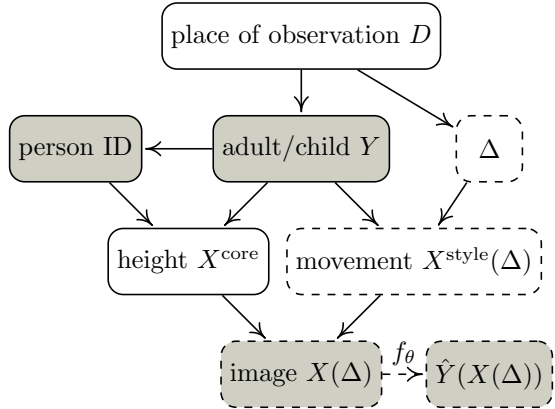

\begin{figure*}
\begin{minipage}[t]{0.45\hsize}
\centering
\subfloat[Examples from test sets 1--3.]{
    \includegraphics[width=.85\textwidth, keepaspectratio=true]{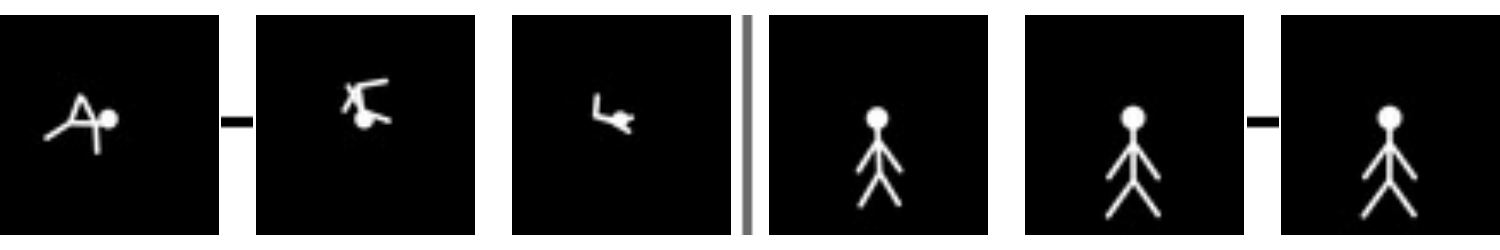}
}

\subfloat{
    \includegraphics[width=.85\textwidth, keepaspectratio=true]{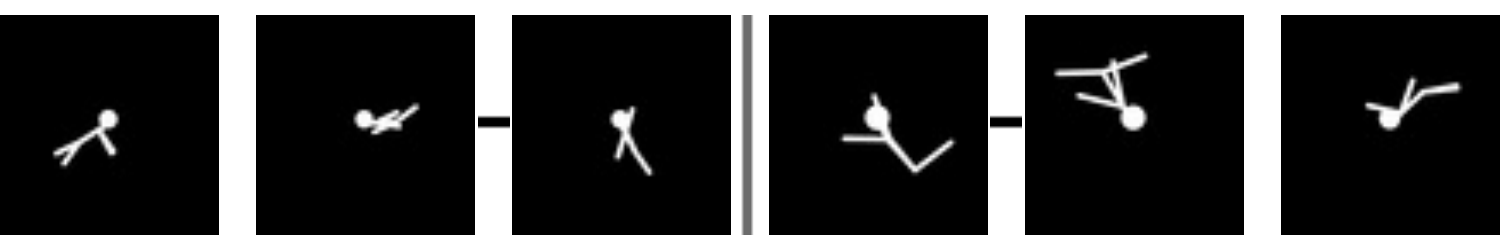}
}

\subfloat{
    \includegraphics[width=.85\textwidth, keepaspectratio=true]{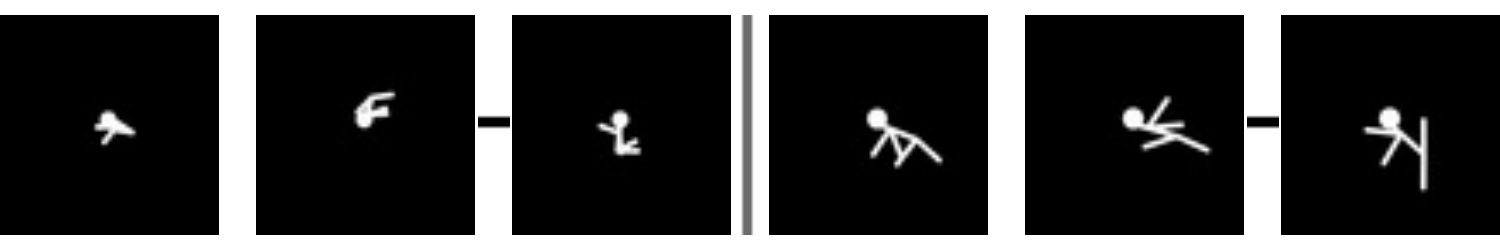}
}
\end{minipage}
\begin{minipage}[t]{0.55\hsize}
\centering
\addtocounter{subfigure}{-2}
\subfloat[Misclassification rates.]{
\includegraphics[width=1\textwidth, keepaspectratio=true]{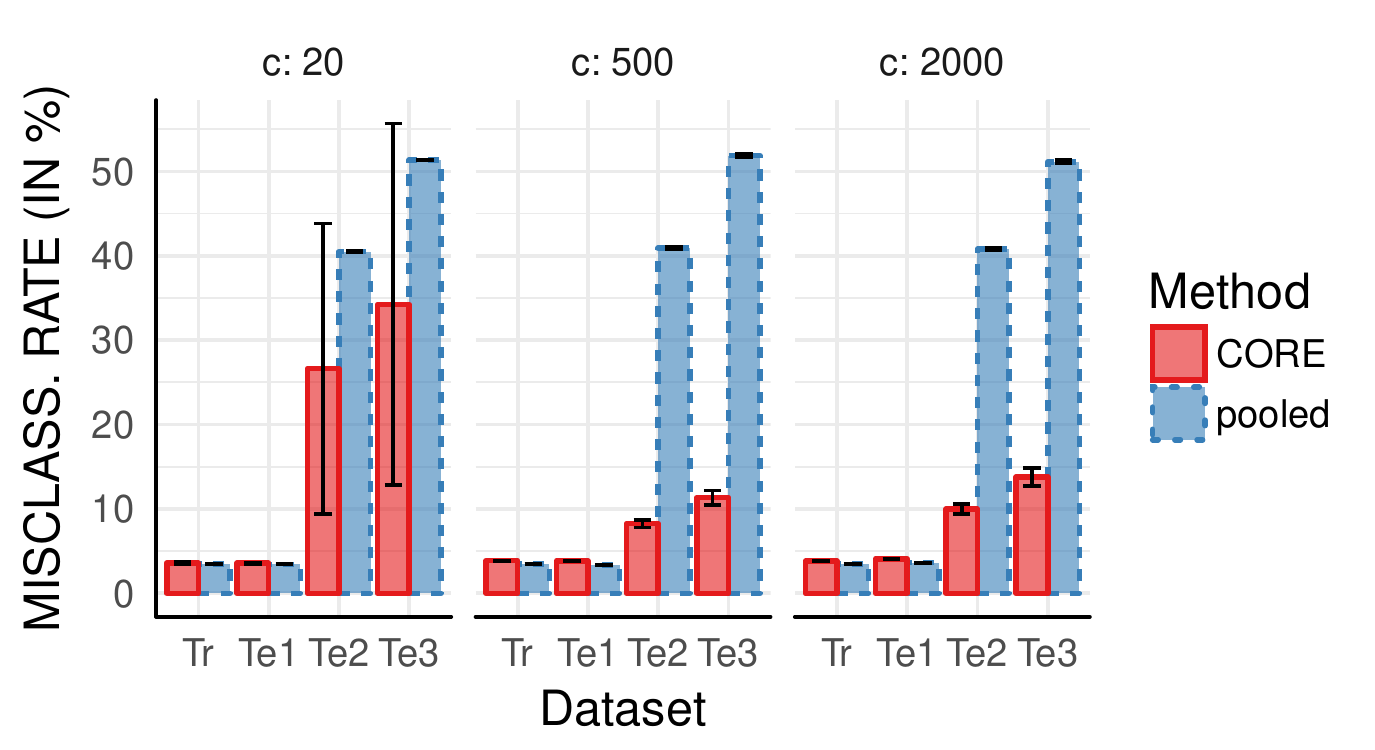}\label{fig:res_stick_supp_res}
}
\end{minipage}
\vspace{-.3cm}
\captionof{figure}{{\small a) Examples from the stickmen test set 1 (row 1), test set 2 (row 2) and test sets 3 (row 3). In each row, the first three images from the left have $y\equiv\textit{child}$; the remaining three images have $y\equiv\textit{adult}$. Connected images are grouped examples. b) Misclassification rates for different numbers of grouped examples.}}
\label{fig:res_stick_supp}
\end{figure*}

Here, we show further results for the experiment introduced in
\S\ref{subsec:stick}. Figure~\ref{fig:DAG_stick} illustrates the data
generating process. Recall that test set 1 follows the same distribution as the training set. In test sets 2 and 3 large
movements are associated with both children and adults, while the movements are heavier in test set 3 than in test set 2.
Figure~\ref{fig:res_stick_supp_res} shows results for different numbers of grouping examples. For $\ncf=20$ the misclassification rate of \core estimator has a large variance. For $\ncf \in \{ 50, 500, 2000 \}$, the \core estimator shows similar results. Its performance is thus not sensitive to the number of grouped examples, once there are sufficiently many grouped observations in the training set. The pooled estimator fails to achieve good predictive performance on test sets 2 and 3 as it seems to use ``movement'' as a predictor for ``age''. 

\subsection{Eyeglasses detection: image quality intervention}\label{subsec:celeba_conf_supp}
Here, we show further results for the experiments introduced in
\S\ref{subsec:celeba_conf}. Specifically, we consider interventions of different strengths by varying the mean of the quality intervention in $\mu \in \{30, 40, 50\}$. 
Recall that we use ImageMagick to modify the image quality.  
In the training set and in test set 1, we sample the image quality value as 
$q_{i,j} \sim \mathcal{N}(\mu, \sigma = 10)$ and apply the command \texttt{convert -quality q\_ij input.jpg output.jpg} if $y_i \equiv \textit{glasses}$. If $y_i \equiv \textit{no glasses}$, the image is not modified. 
In test set 2, the above command is applied if $y_i \equiv \textit{no glasses}$ while images with $y_i \equiv \textit{glasses}$ are not changed. In test set 3 all images are left unchanged and in test set 4 the command is applied to all images, i.e.\ the quality of all images is reduced.

\begin{figure*}[!t]
\centering
\subfloat[Grouping setting 1, $\mu = 50$]{
     \includegraphics[width=.3\textwidth, keepaspectratio=true]{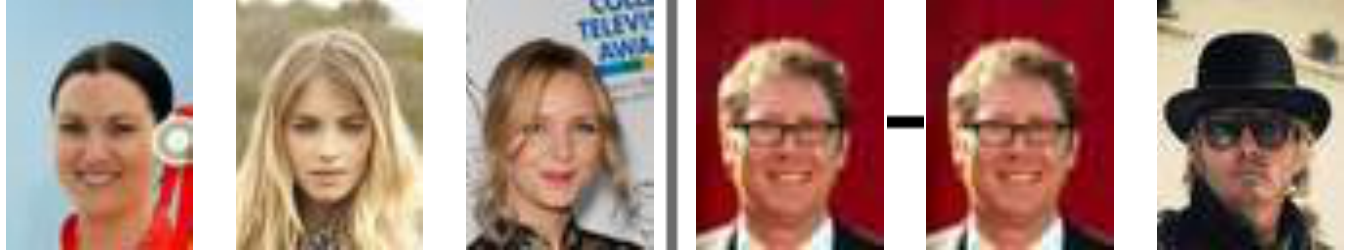}
}
\hspace{.2cm}
\subfloat[Grouping setting 1, $\mu = 40$]{
     \includegraphics[width=.3\textwidth, keepaspectratio=true]{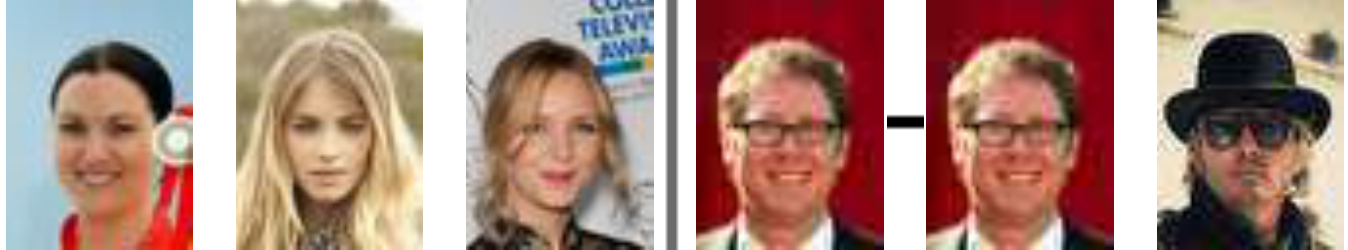}
}
\hspace{.2cm}
\subfloat[Grouping setting 1, $\mu = 30$]{
     \includegraphics[width=.3\textwidth, keepaspectratio=true]{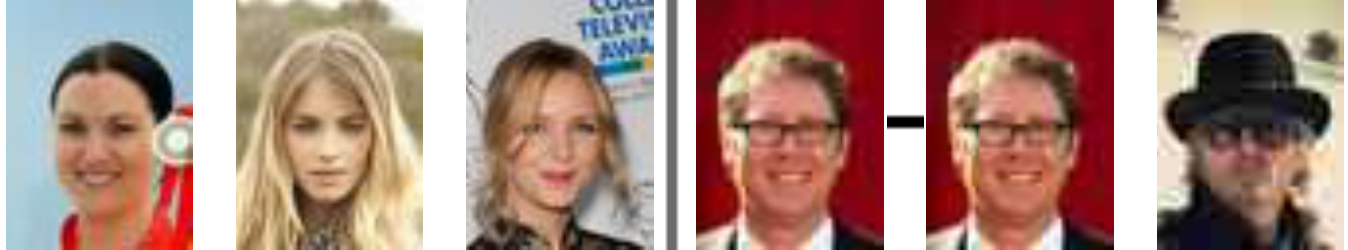}
}

\vspace{-.1cm}
\subfloat{
     \includegraphics[width=.3\textwidth, keepaspectratio=true]{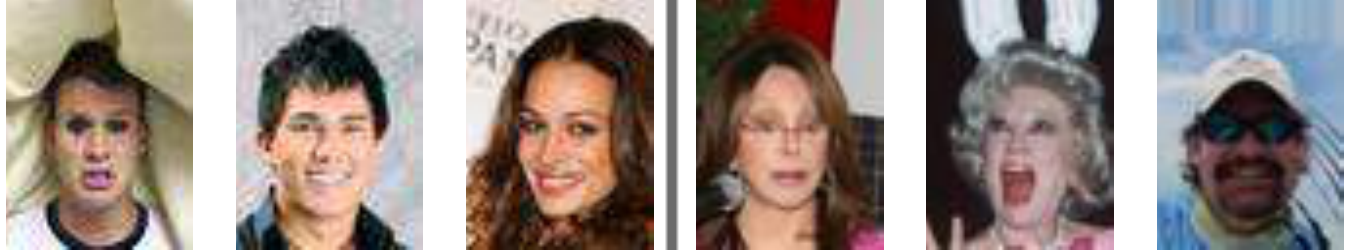}
}
\hspace{.2cm}
\subfloat{
     \includegraphics[width=.3\textwidth, keepaspectratio=true]{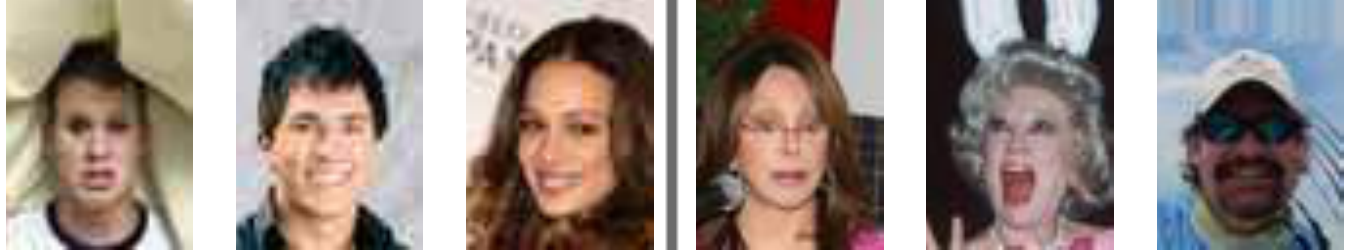}
}
\hspace{.2cm}
\subfloat{
     \includegraphics[width=.3\textwidth, keepaspectratio=true]{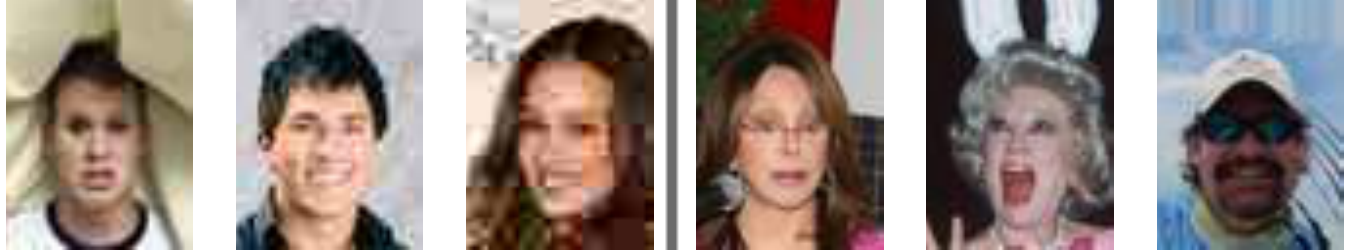}
}

\vspace{-.1cm}
\subfloat{
     \includegraphics[width=.3\textwidth, keepaspectratio=true]{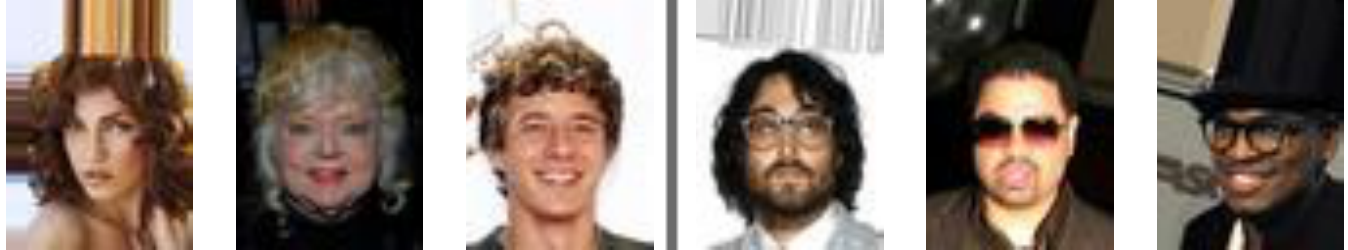}
}
\hspace{.2cm}
\subfloat{
     \includegraphics[width=.3\textwidth, keepaspectratio=true]{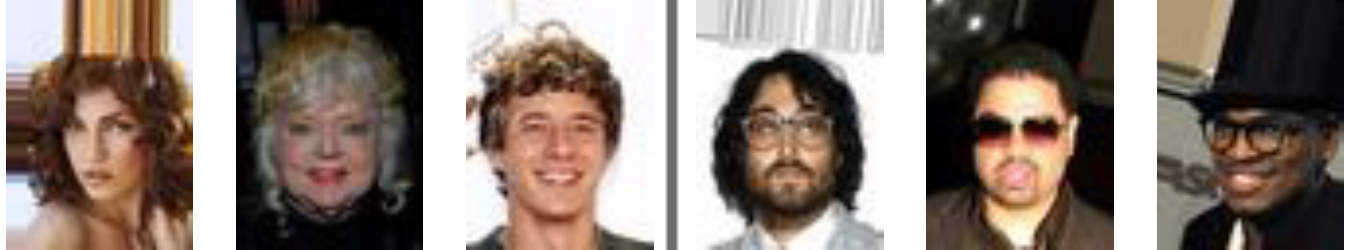}
}
\hspace{.2cm}
\subfloat{
     \includegraphics[width=.3\textwidth, keepaspectratio=true]{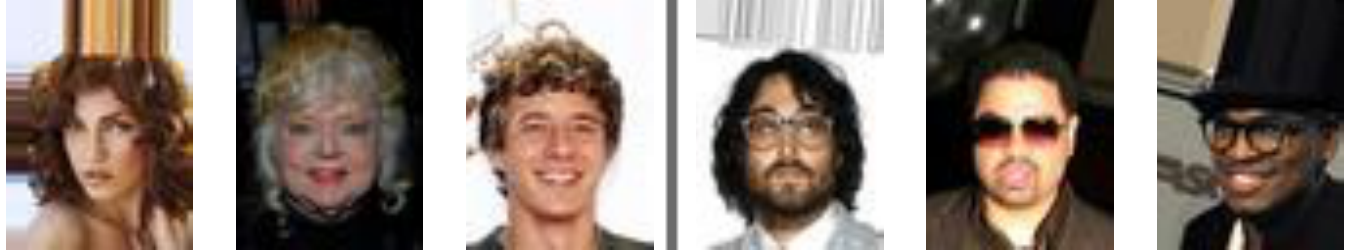}
}

\vspace{-.1cm}
\subfloat{
     \includegraphics[width=.3\textwidth, keepaspectratio=true]{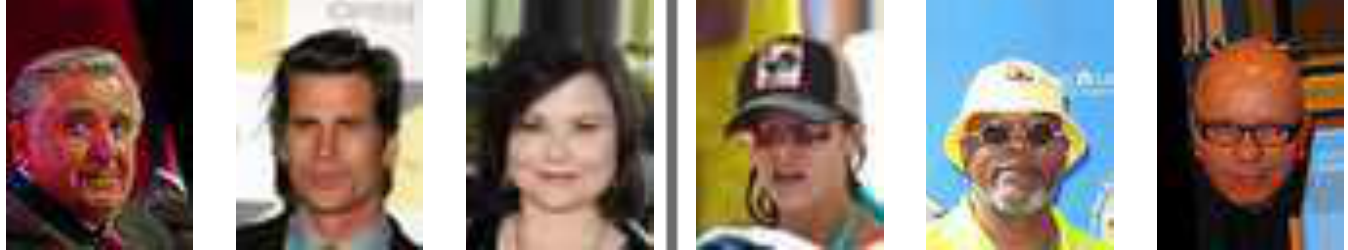}
}
\hspace{.2cm}
\subfloat{
     \includegraphics[width=.3\textwidth, keepaspectratio=true]{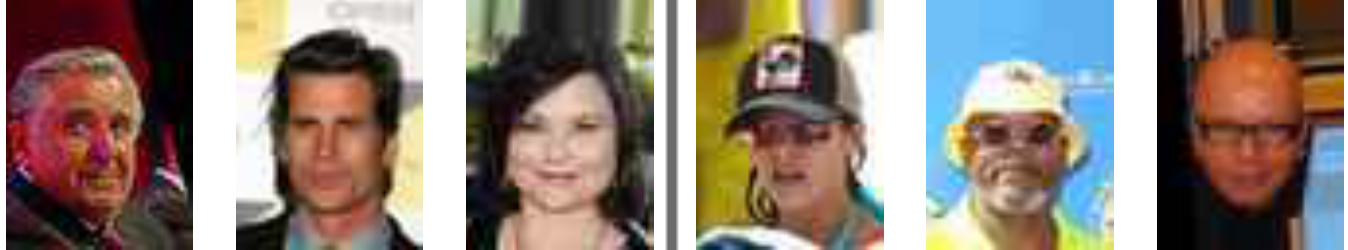}
}
\hspace{.2cm}
\subfloat{
     \includegraphics[width=.3\textwidth, keepaspectratio=true]{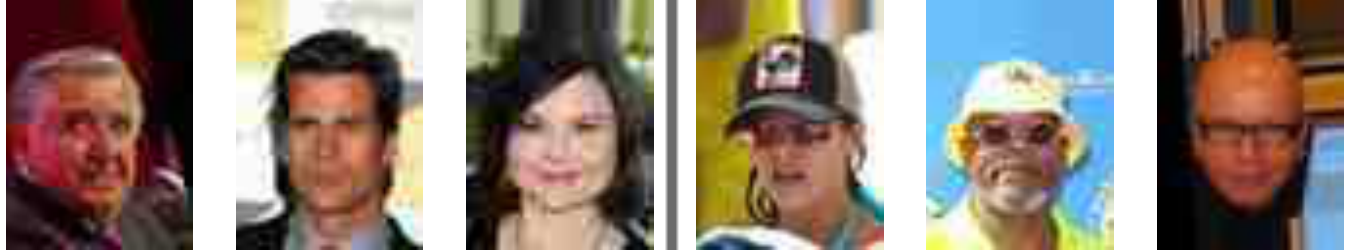}
}

\addtocounter{subfigure}{-9}
\subfloat[Grouping setting 2, $\mu=50$]{
     \includegraphics[width=.3\textwidth, keepaspectratio=true]{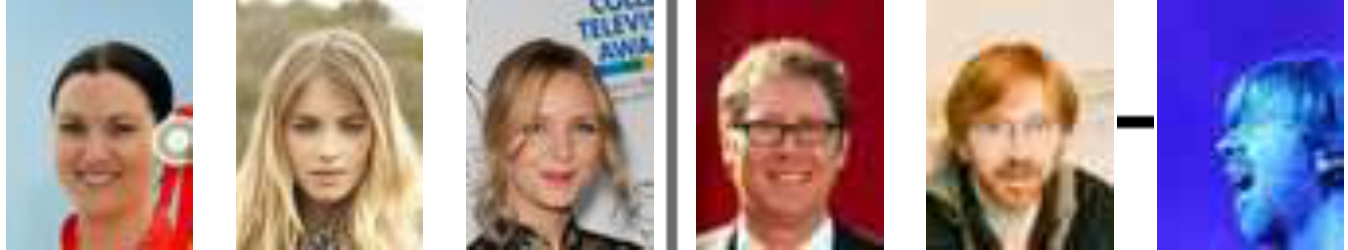}
}
\hspace{.2cm}
\subfloat[Grouping setting 2, $\mu=40$]{
     \includegraphics[width=.3\textwidth, keepaspectratio=true]{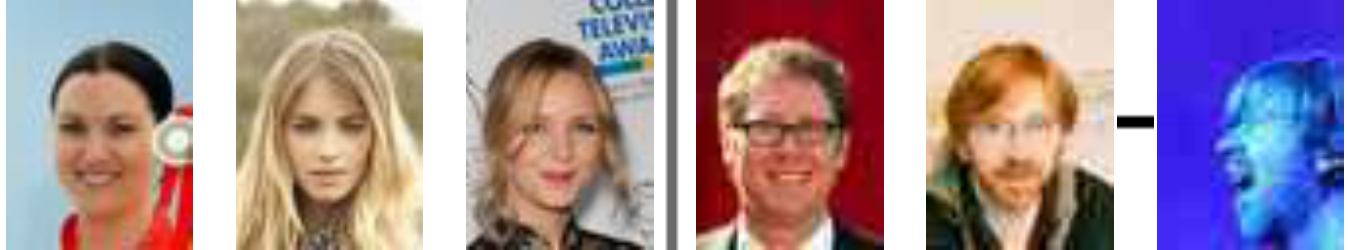}
}
\hspace{.2cm}
\subfloat[Grouping setting 2, $\mu=30$]{
     \includegraphics[width=.3\textwidth, keepaspectratio=true]{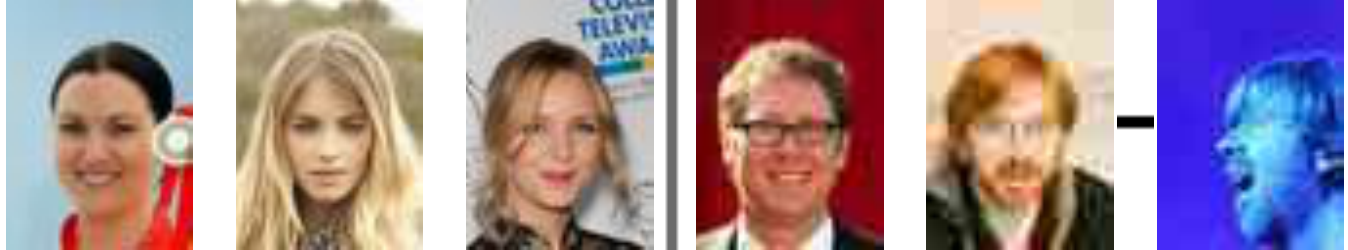}
}

\subfloat[Grouping setting 3, $\mu=50$]{
     \includegraphics[width=.3\textwidth, keepaspectratio=true]{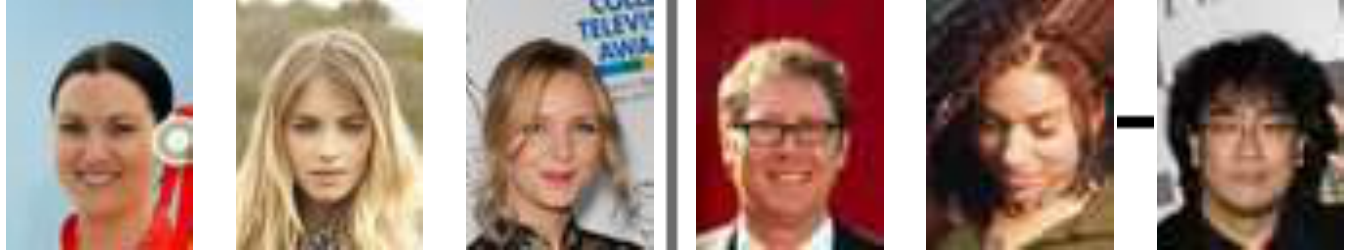}
}
\hspace{.2cm}
\subfloat[Grouping setting 3, $\mu=40$]{
     \includegraphics[width=.3\textwidth, keepaspectratio=true]{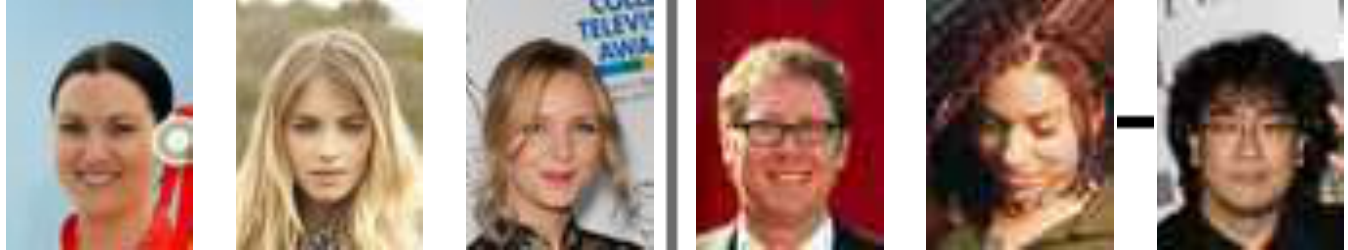}
}
\hspace{.2cm}
\subfloat[Grouping setting 3, $\mu=30$]{
     \includegraphics[width=.3\textwidth, keepaspectratio=true]{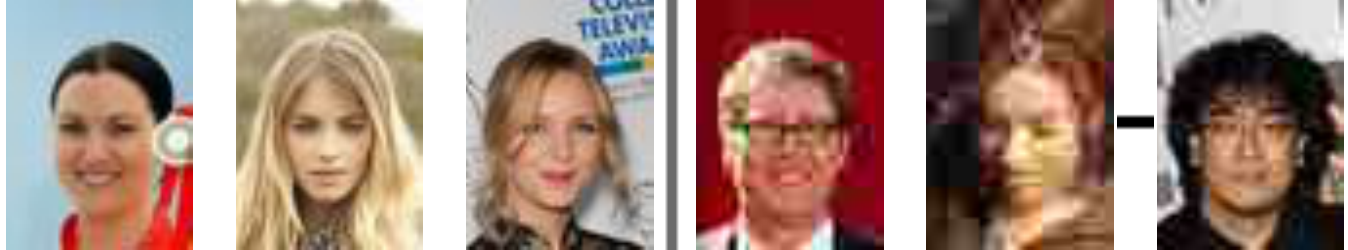}
}
\captionof{figure}{{\small  Examples from the CelebA image quality
    datasets, grouping settings 1--3 with $\mu \in \{30, 40, 50\}$. In
    all rows, the first three images from the left have
    $y\equiv\textit{no glasses}$; the remaining three images have
    $y\equiv\textit{glasses}$. Connected images are grouped
    observations over which we calculate the conditional variance. In panels (a)--(c), row 1 shows examples from the training set, rows 2--4 contain examples from test sets 2--4, respectively. Panels (d)--(i) show examples from the respective training sets.}}\label{fig:data_set_celeb1_qual_supp}
\end{figure*}

\begin{figure}
\centering
\subfloat[Grouping setting 1]{
     \includegraphics[width=.33\textwidth, keepaspectratio=true]{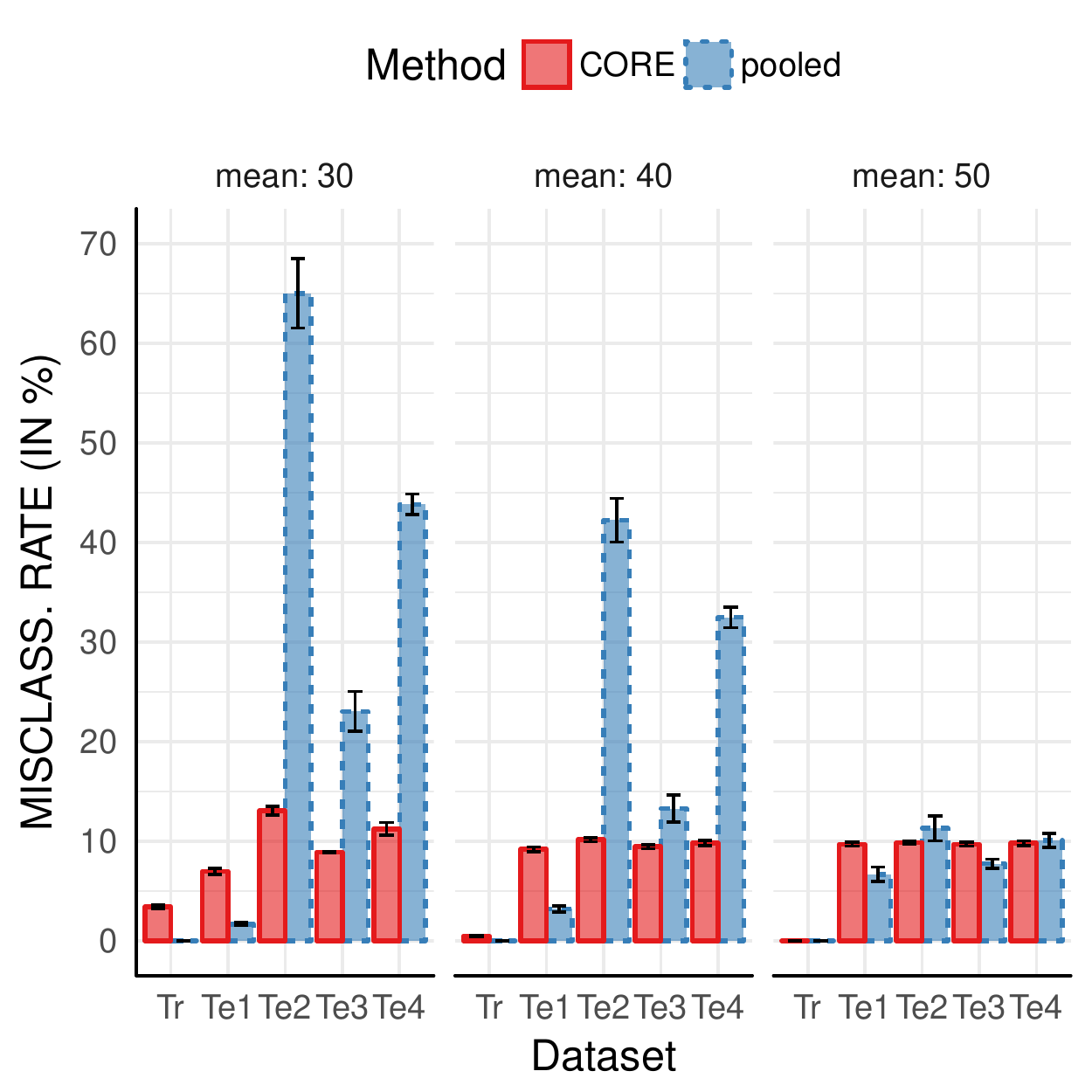}
}
\subfloat[Grouping setting 2]{
     \includegraphics[width=.33\textwidth, keepaspectratio=true]{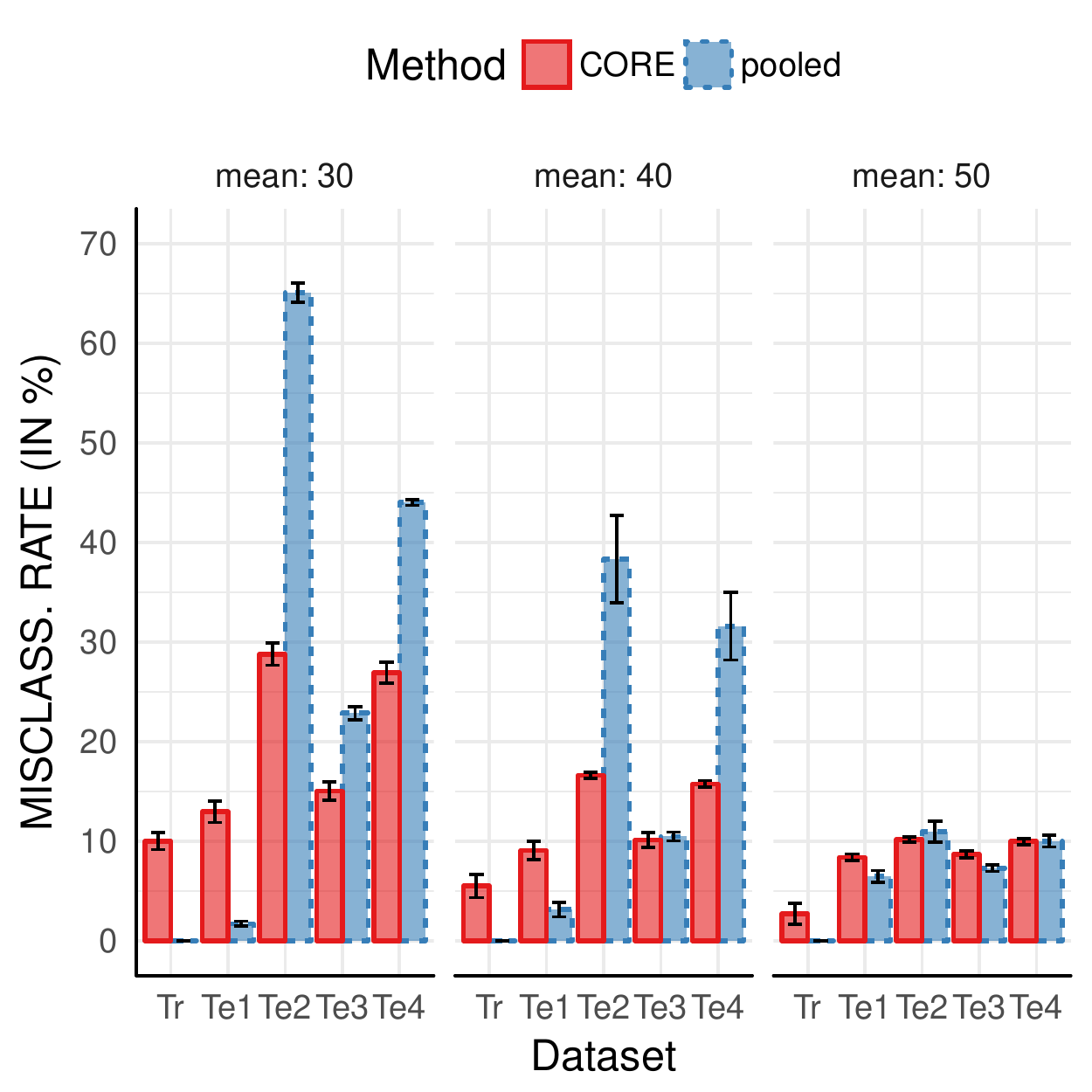}
}
\subfloat[Grouping setting 3]{
     \includegraphics[width=.33\textwidth, keepaspectratio=true]{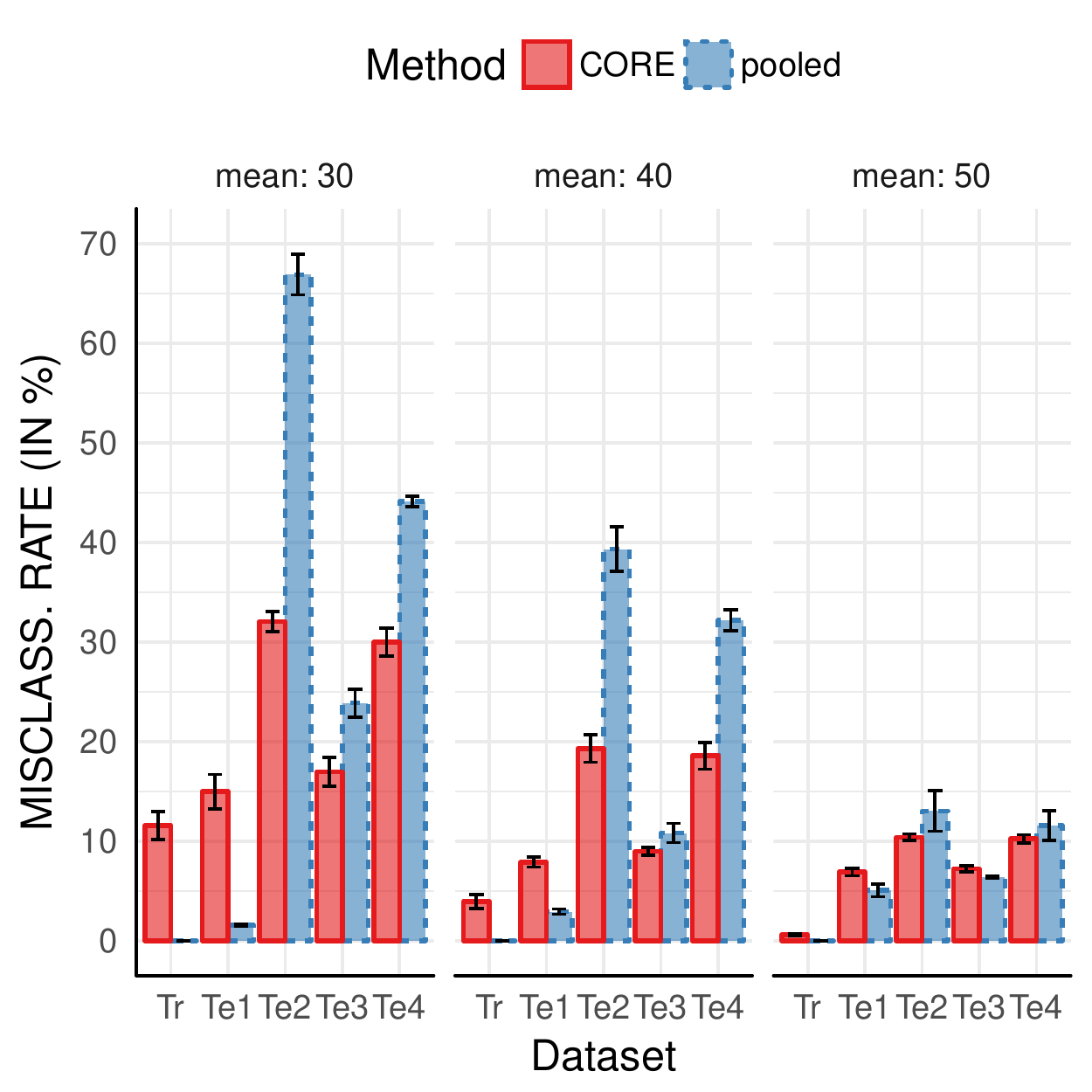}
}
\captionof{figure}{{\small Misclassification rates for the CelebA  eyeglasses detection with image quality interventions, grouping settings 1--3 with $\ncf= 5000$ and the mean of the Gaussian distribution $\mu \in \{30, 40, 50\}$.}}\label{fig:res_celeb_qual_supp}
\end{figure}

We run experiments for grouping settings 1--3 and for $\ncf=5000$,
where the definition of the grouping settings 1--3 is identical to \S\ref{subsec:celeb_brightness_supp}. Figure~\ref{fig:data_set_celeb1_qual_supp} shows examples from the respective training and test sets and Figure~\ref{fig:res_celeb_qual_supp} shows the corresponding misclassification rates. Again, we observe that grouping setting 1 works best, followed by grouping setting~2. Interestingly, there is a large performance difference between $\mu=40$ and $\mu=50$ for the pooled estimator. Possibly, with $\mu=50$ the image quality is not sufficiently predictive for the target. 

\clearpage
\subsection{Elmer the Elephant}\label{subsec:elmer_supp}

\begin{figure*}[!t]
\centering
\subfloat{
     \includegraphics[width=.4\textwidth, keepaspectratio=true]{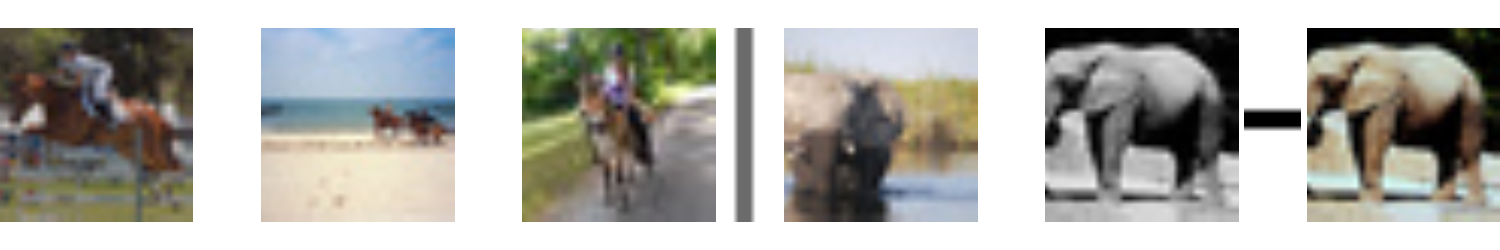}
}

\vspace{-.4cm}
\subfloat{
     \includegraphics[width=.4\textwidth, keepaspectratio=true]{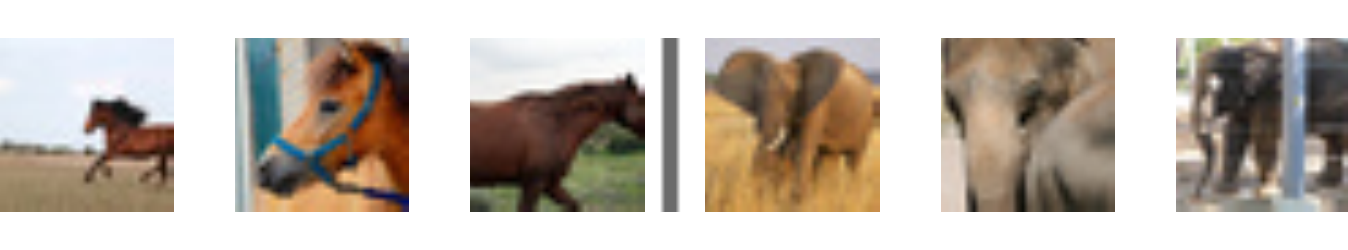}
}

\vspace{-.4cm}
\subfloat{
     \includegraphics[width=.4\textwidth, keepaspectratio=true]{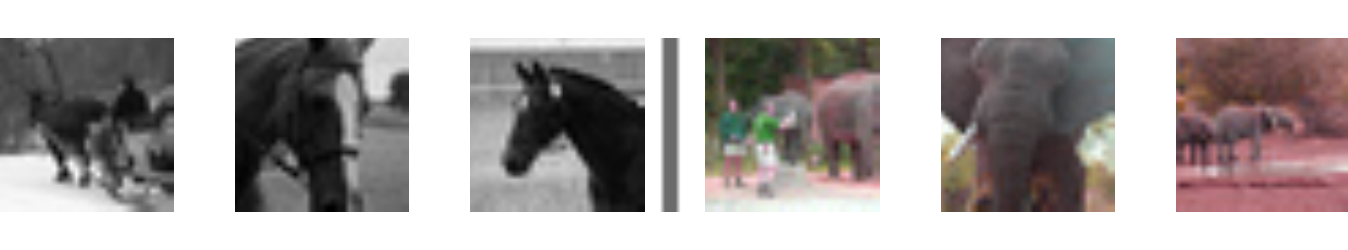}
}

\vspace{-.4cm}
\subfloat{
     \includegraphics[width=.4\textwidth, keepaspectratio=true]{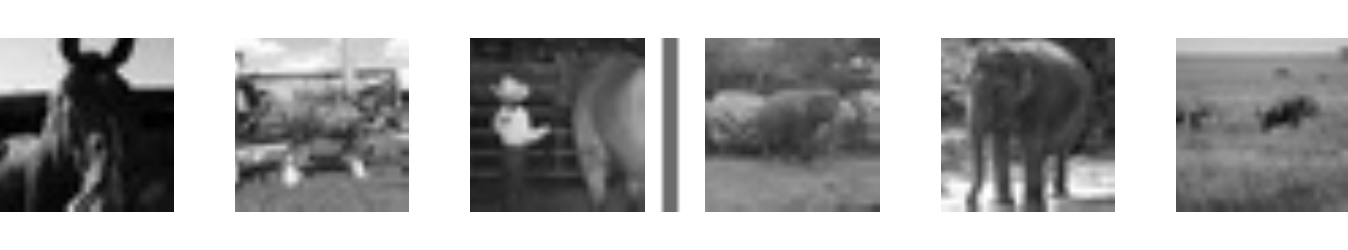}
}

\vspace{-.4cm}
\subfloat{
     \includegraphics[width=.4\textwidth, keepaspectratio=true]{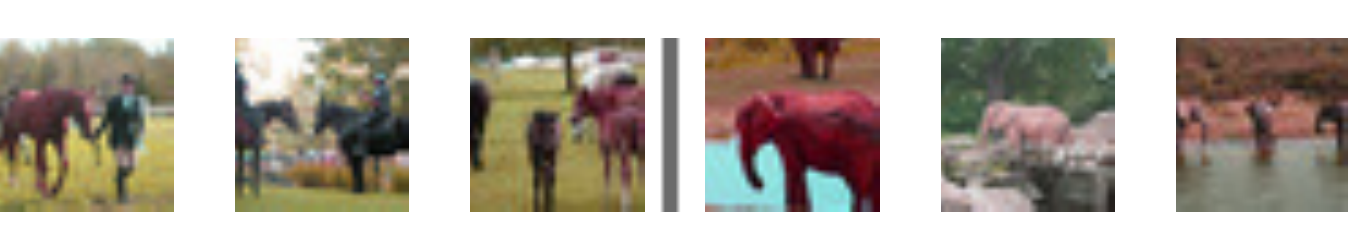}
}
\captionof{figure}{{\small Examples from the subsampled and augmented AwA2 dataset (Elmer-the-Elephant dataset). Row 1 shows examples from the training set, rows 2--5 show examples from test sets 1--4, respectively.}}\label{fig:data_set_ani_supp}
\end{figure*}

\begin{figure*}
\begin{minipage}[t]{0.5\hsize}
\centering

\vspace{-.1cm}
\subfloat[Examples of misclassified observations.]{
\parbox{.33\linewidth}{%
{\tiny
$y \equiv \textit{horse}$ \\
$\phc(\textit{horse}) = 0.72$ \\
$\php(\textit{horse}) = 0.01$ \\  }
\centering
      \includegraphics[width=.75\linewidth, keepaspectratio=true, trim={4.5cm 0 6.8cm 0}, clip]{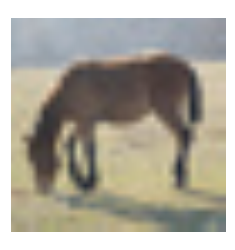} }
\parbox{.33\linewidth}{%
    {\tiny
$y \equiv \textit{horse}$ \\
$\phc(\textit{horse}) = 1.00$ \\
$\php(\textit{horse}) = 0.01$ \\}
\centering
       \includegraphics[width=.75\linewidth, keepaspectratio=true, trim={0cm 0 11.4cm 0}, clip]{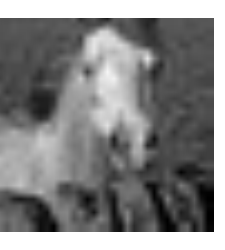} }
\parbox{.33\linewidth}{%
 {\tiny
$y \equiv \textit{elephant}$ \\
$\phc(\textit{ele.}) = 0.95$ \\
$\php(\textit{ele.}) = 0.00$ \\}
\centering
       \includegraphics[width=.75\linewidth, keepaspectratio=true, trim={6.9cm 0 4.5cm 0}, clip]{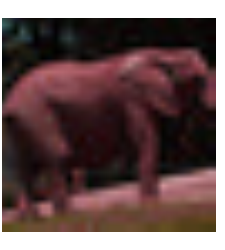} }
    }
\end{minipage}
\begin{minipage}[t]{0.5\hsize}
\subfloat[Misclassification rates.]{
\centering
\includegraphics[width=1\textwidth, keepaspectratio=true]{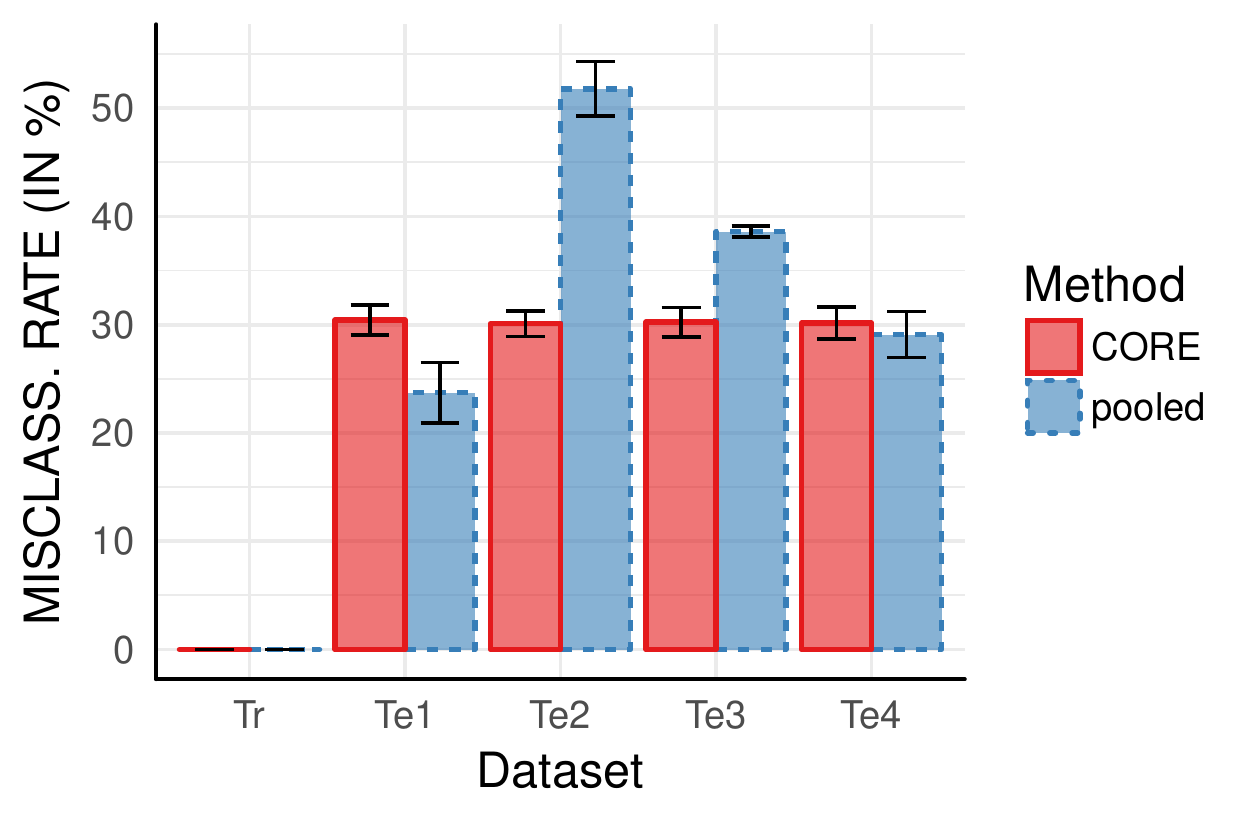} \label{fig:elmer_res}}
\end{minipage}
\vspace{-.4cm}
\captionof{figure}{\small Elmer-the-Elephant dataset. (a) Misclassified examples from the test
  sets. (b) Misclassification rates on test sets~1 to~4.}
\label{fig:data_set_elmer}
\vspace{-.3cm}
\end{figure*}

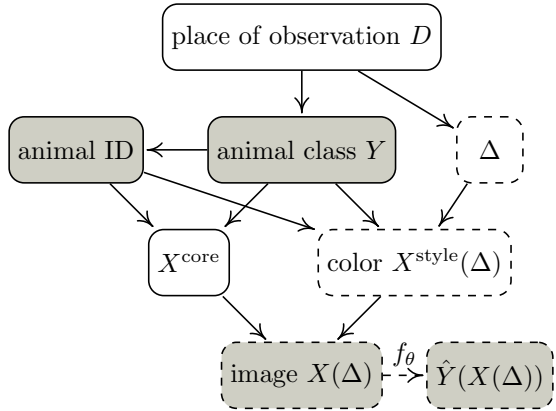
\begin{figure}
\begin{center} 
\begin{small}
\begin{tikzpicture}[>=stealth',shorten >=1pt,auto,node distance=3.2cm,
                    semithick]
  \tikzstyle{every state}=[fill=none,shape=rectangle,rounded corners=2mm,text=black]

  \draw (0,4.5) node[state] (H) {place of observation
    $\domain$};
  \draw (0,3) node[state, fill=pastelgray] (Y) {animal class $Y$};
  \draw (1.5,1.5) node[state, dashed] (S) {color $\orth(\Delta)$ };
  \draw (2.5,3) node[state, dashed] (D) {$\Delta$};
  \draw (-3,3) node[state, fill=pastelgray] (I) { animal  $\I$};
  \draw (-1.5,1.5) node[state] (W) { $\corefeat$};
  \draw (0,0) node[state, fill=pastelgray, dashed] (X) {image $X(\Delta)$};
  \draw (2.5,0) node[state, fill=pastelgray, dashed] (Yhat) {$\hat{Y}(X(\Delta))$};

 \draw [-arcsq] (H) -- (Y);
 \draw [-arcsq]  (Y) -- (I);
 \draw [-arcsq]  (I) -- (W);
 \draw [-arcsq]  (I) -- (S);
 \draw [-arcsq]  (Y) -- (S);
 \draw [-arcsq] (H) -- (D);
 \draw [-arcsq] (Y) -- (W);
 \draw [-arcsq] (D) -- (S);
 \draw [-arcsq] (W) -- (X);
 \draw [-arcsq] (S) -- (X);
 \draw [-arcsq, dashed] (X) -- node {$f_\theta$} (Yhat);
\end{tikzpicture}
\captionof{figure}{\small Data generating process for the Elmer-the-Elephant example.}
\label{fig:DAG_ani}
\end{small}
\end{center}
\end{figure}

The color interventions for the experiment introduced in
\S\ref{subsec:elmer} were created as follows. In the training set, if
$y_i \equiv \textit{elephant}$ we apply the following ImageMagick
command  for the grouped examples \texttt{convert -modulate
  100,0,100 input.jpg output.jpg}.
Test sets~1 and~2 were already discussed in \S\ref{subsec:elmer}:
in test set~1, all images are left unchanged. In test set~2, the above
command is applied if $y_i \equiv \text{horse}$. If $y_i \equiv
\text{elephant}$, we sample $c_{i,j} \sim \mathcal{N}(\mu = 20,
\sigma = 1)$ and apply \texttt{convert -modulate 100,100,100-c\_ij
  input.jpg output.jpg} to the image. 
Here, we consider again some more test sets than in \S\ref{subsec:elmer}. 
In test set~4, the latter command is applied to all images. It rotates the colors of the image, in a cyclic manner\footnote{For more details, see 
\url{http://www.imagemagick.org/Usage/color_mods/\#color_mods}.}. In
test set~3, all images are changed to grayscale. 
The causal graph for the data generating process is shown in Figure~\ref{fig:DAG_ani}. 
Examples from all four test sets are shown in
Figure~\ref{fig:data_set_ani_supp} and classification results are
shown in Figure~\ref{fig:data_set_elmer}.

